\documentclass[final]{article}

\usepackage[main]{nips26/neurips_2026}

\usepackage{nips26/nips_style}
\usepackage[utf8]{inputenc} %
\usepackage[T1]{fontenc}    %
\usepackage{url}            %
\usepackage{booktabs}       %
\usepackage{amsfonts}       %
\usepackage{nicefrac}       %
\usepackage{microtype}      %
\usepackage{xcolor}         %
\usepackage{longtable}

\usepackage{colortbl}
\usepackage{float}%
\usepackage{titletoc}
\usepackage{standalone} %
\usepackage{adjustbox} %
\usepackage{overpic}    %

\usepackage[colorlinks=true,
    linktoc=all]{hyperref}       %
\hypersetup{
    urlcolor=[rgb]{0,0,0.5},
    citecolor=[rgb]{0,0,0.5},
    linkcolor=[rgb]{0,0,0.5}
}

\newcommand{\renyi}{Rényi}
\newcommand{\byzalg}{{\tt ByzULDP}}
\newcommand{\byzobj}{F_{\calG}}
\newcommand{\orange}{\color{orange!70}}
\newcommand{\ua}{{\blue \uparrow}}
\newcommand{\da}{{\orange \downarrow}}

\title{Optimal Byzantine-resilient Federated Learning with User-level Differential Privacy}

\author{%
Ming Xiang$^1$ \quad
Stratis Ioannidis$^1$ \quad 
Edmund Yeh$^1$ \quad 
Carlee Joe-Wong$^2$ \quad 
Lili Su$^1$ \\
$^1$Northeastern University, Boston, MA, USA; \quad 
$^2$Carnegie Mellon University, Pittsburgh, PA, USA.\\
\texttt{\{xiang.mi,e.ioannidis,e.yeh,l.su\}@northeastern.edu},
\texttt{cjoewong@andrew.cmu.edu}
}

\begin{document}

\maketitle

\begin{abstract}
Real-world deployment of federated learning (FL) requires resilience against adversarial clients and data privacy. Byzantine fault captures the worst-case behavior of adversarial clients, while differential privacy (DP) provides statistical guarantees against privacy loss. In particular, user-level DP guards a client's entire contribution and provides more realistic protection against information leakage. Despite intensive efforts, the minimax-optimal convergence bound of Byzantine-resilient FL with user-level DP remains open. We establish an information-theoretic lower bound for any Byzantine-resilient empirical risk minimization problem under user-level DP and heterogeneous clients. We propose~\byzalg, a central DP algorithm featuring a two-level momentum across the client and server. Client-side momentum mitigates stochastic gradient variance during robust aggregation. Unlike the fault-free case, naively combining non-linear Byzantine-resilient aggregation with central DP noise can be insufficient to guarantee convergence to a bounded neighborhood of a stationary point. Our key contribution is a server-side momentum that leverages a property specific to user-level DP, where privacy sensitivity depends on how client updates are constructed, to empower convergence.~\byzalg~achieves the order-optimal convergence bound for strongly-convex objectives, with vanishing error terms decaying at the optimal ${\mathcal{O}}(1/T)$ and ${\mathcal{O}}(1/\sqrt{T})$ rates for strongly-convex and non-convex objectives, respectively. We corroborate our analysis with numerical experiments on real-world datasets across diverse privacy budgets.

\end{abstract}

\section{Introduction}
\label{sec: introduction}
Federated learning (FL) enables large-scale AI/ML training without transmitting raw local data through collaboration between a parameter server (PS) and clients.
Its success has been evidenced by widespread commercial deployments in diverse fields such as 
autonomous vehicles 
\cite{chen2021bdfl,peng2023privacy,zeng2022federated},
the Internet of things \cite{ghimire2022recent,nguyen2019diot,zhang2022federated},
and natural language processing \cite{lin2022fednlp,liu2023communication,moskvoretskii2024low}. 

While FL keeps raw data locally, shared model parameters can still leak sensitive information about clients' training data~\cite{bai2024membership,huang2021evaluating}. 
Differential privacy (DP) provides statistical guarantees against such leakage~\cite{dwork2014algorithmic}, with the client as the natural unit of protection.
In regulated domains such as healthcare and finance, the aggregating server operates as a trusted intermediary under legal accountability.
The privacy threat comes from external auditors or curious clients,
who attempt to infer sensitive information about participating clients.
Meanwhile, FL operates over a complex and heterogeneous client population \cite{kairouz2021advances},
where 
faulty or abnormal clients arise
due to 
software or hardware bugs \cite{rathbun2024},
poisoned data \cite{song2020poisonrec},
or adversarial manipulations~\cite{yang2024robust}.
Such unstructured failures are well captured by the Byzantine failure model \cite{lynch1996distributed}. 
Byzantine-resilience~\cite{blanchard2017machine,chen2017distributed,xie2019zeno}
and user-level DP~\cite{hu2020personalized,levy2021learning,wei2020federated}
have each received extensive attention in isolation.
However, real-world deployments, such as healthcare, mobile personalization, and financial collaborations, require addressing both of them simultaneously.
Their integration is fundamentally challenging:
Byzantine-resilience aims to preserve the utility
by discriminating adversarial updates, 
whereas user-level DP aims to obscure clients' contributions, potentially drowning useful signals. 
We therefore ask:
\begin{center}
    {\bf \em Question:} 
    {\em 
    Under a trusted server, what is the minimax-optimal convergence bound of user-level DP empirical risk minimization (ERM) under Byzantine adversaries and heterogeneous clients?
    }
\end{center}

Combining DP and Byzantine-resilience remains largely underexplored.
One line of work provides per-step privacy guarantees
\cite{guerraoui2021differential,ma2022differentially,xiang2022distributed,zhubridging}, yielding a loose cumulative privacy cost that grows with the number of rounds $T$.
Only a handful of works study the synthesis of DP and Byzantine-resilience with multi-round guarantees.
Allouah et al.~\cite{allouah2023privacy} and Gu et al. \cite{gu2025dp} both consider item-level DP with Byzantine resilience, with the analysis of the latter requiring~\iid~clients.
Yet, FL is by definition heterogeneous, and item-level DP often provides insufficient protection against user-level leakage \cite{levy2021learning}.
For example, in a specialized hospital,
client records often exhibit correlated structures.
Per-item perturbations cannot fully mask cross-record correlations,
leaving the cumulative client signature recoverable.
Closer to our setting, Allouah et al.~\cite{allouah2025towards} achieve user-level DP without a trusted server via locally correlated noise injection at each client, seeking to attain the utility of central DP algorithms, where the server is trusted.
In the absence of Byzantine adversaries, central DP is known to achieve a favorable privacy-utility trade-off~\cite{duchi2013local}.
However, the minimax-optimal convergence bound for Byzantine-resilient ERM with user-level DP remains open.

\noindent {\bf Contributions.} 
In this work, we establish order-matching lower and upper bounds for Byzantine-resilient ERM with user-level DP.
The lower bound is privacy-regime agnostic, while the upper bound is attained by a central DP algorithm, showing that central DP suffices to achieve the optimum.
Our contributions are fourfold:
\begin{itemize}[leftmargin=*]
    \item 
    In~\prettyref{sec: lower bound}, we establish an information-theoretic lower bound for any Byzantine-resilient ERM problem under user-level DP constraints and heterogeneous clients.
    \item 
    In~\prettyref{sec: algorithm}, we propose the~\byzalg, which features two-level momentum across the client and the server. 
    Client-side momentum mitigates stochastic gradient variance during robust aggregation, which ensures Byzantine resilience. 
    Server-side momentum reshapes each client's contribution so that the privacy sensitivity achieves a reduction specific to user-level DP,
    enabling convergence.
    \item 
    In~\prettyref{sec: convergence}, we provide two convergence results for~\byzalg.
    For strongly convex objectives, the non-vanishing errors match our lower bound, and the vanishing error terms decay at the optimal $\calO(1/T)$ rate for stochastic first-order methods.
    For non-convex objectives, we obtain the optimal $\calO (1/\sqrt{T})$ rate on the vanishing error terms.
    \item 
    In~\prettyref{sec: numerical experiment}, we corroborate our analysis with numerical experiments on real-world datasets across a range of privacy budgets.
\end{itemize}

\section{Related Work}
\label{sec: related work}

\paragraph{User-level Private Learning.}
Without Byzantine resilience, user-level private learning has received growing attention.
For example,
Badih et al.~\cite{badih2021user} study user-level DP for PAC learning and characterize the number of clients required to learn a concept class. 
The seminal work~\cite{levy2021learning} by Levy et al. studies user-level DP for mean estimation, ERM, and stochastic convex optimization (SCO) under homogeneous client distributions or mildly heterogeneous data.
Bassily et al.~\cite{bassily2023user} and Liu et al.~\cite{liu2024user} both achieve the optimal rates for user-level DP-SCO, with the latter adopting more relaxed assumptions on how the number of clients scales with the problem dimension.
Another line of work explores intermediate threat models bridging local and central DP by using cryptographic techniques or anonymization via shuffling ~\cite{cheu2019distributed,jayaraman2018distributed}.
Yet, Byzantine clients fundamentally alter the problem.
Most prior work relies on a simple mean aggregation, which is highly vulnerable to adversarial updates.
Naively combining robust aggregators with DP noise is insufficient to guarantee convergence to a bounded neighborhood of a stationary point, as these nonlinear aggregators often exhibit high privacy sensitivity; see~\prettyref{rmk: nonconvex convergence} for details.

\paragraph{Byzantine-resilient Learning.}
Tolerating Byzantine attacks in distributed learning has received intensive attention \cite{allouah2023fixing,blanchard2017machine,chen2017distributed,farhadkhani2022byzantine,feng2014distributed,ghosh2020communication,karimireddybyzantine,karimireddy2021learning,su2016fault,sundaram2015consensus,xie2019zeno,yin2018byzantine}.
A robust aggregator is the most common way to mitigate the negative impacts of adversarial updates from Byzantine clients.
Numerous robust aggregators have been developed over the recent years, including Geometric Median~\cite{chen2017distributed}, 
Krum~\cite{blanchard2017machine}, 
Coordinate-wise Trimmed Mean~\cite{yin2018byzantine}, 
Coordinate-wise Median~\cite{yin2018byzantine}, 
Bulyan~\cite{guerraoui2018hidden}, 
and centered clipping~\cite{karimireddy2021learning}.
Yet, their original analysis is under a simplified~\iid setting.
Karimireddy et al.~\cite{karimireddybyzantine} and Allouah et al. \cite{allouah2023fixing} have shown that any distributed algorithm can only reach a neighborhood of a stationary point in the presence of coupled non-\iid data and Byzantine faults.
Karimireddy et al.~\cite{karimireddybyzantine} propose a simple bucketing mechanism
to reduce the likelihood that any subset of clients is consistently compromised.
Allouah et al.~\cite{allouah2023fixing} propose nearest-neighbor mixing, which remixes the clients' updates with their closest neighbors
to obtain a better representation of the global gradient.

\paragraph{Robust learning and estimation with DP.}
Recent advances focus on the user-level private-mean estimation problem 
with \cite{acharya2021robust,cheu2021manipulation,kasiviswanathan2011can}
or without Byzantine adversaries~\cite{agarwal2025private,girgis2022distributed,narayanan2022tight}, where the clients send the perturbed local data to the server only once.
However, FL prohibits direct data transmission and requires multi-round interactions.
Another line of work leverages cryptographic techniques, \eg, 
homomorphic encryption and secure multi-party computation to reconcile robustness and privacy \cite{thevenon2024b,corrigan2017prio}, but these approaches incur prohibitive computation overhead when combined with robust aggregations.
Recently, Allouah et al.~\cite{allouah2023privacy} establish a privacy-utility-robustness trilemma for item-level DP in the presence of Byzantine adversaries.
Gu et al.~\cite{gu2025dp} also study item-level DP and propose algorithms in both local and central DP modes, yet their analysis is only applicable to~\iid clients.
Allouah et al.~\cite{allouah2025towards} consider user-level DP with an untrusted server and propose an algorithm that uses correlated local noise to achieve utility comparable to
central DP algorithms. 
We establish an improved convergence bound in the central DP setting; see~\prettyref{rmk: sc optimal} for details.

\section{Problem Formulation}
\label{sec: problem formulation}

\noindent{\bf Federated Learning System.}
We consider the classic master-slave architecture with a collection of $n$ clients and a PS.
The clients hold local data sets $D_{1}, \ldots, D_{n}$.
The local loss function at client $i$ is defined as
$F_i (\bfthe) \triangleq ~ \Expect_{(x, y)\sim D_i} [\ell (\bfthe, (x,y))]$,  
where, with a little abuse of notation, $\Expect_{(x, y)\sim D_i}$ can be understood as taking expectation over the empirical data distribution.
The global objective is the average of clients' local loss functions: 
\[
F(\bfthe) \triangleq \frac{1}{n} \sum_{i=1}^n F_i (\bfthe).
\]

\paragraph{Threat model.}
We consider an FL environment with two simultaneous threats: {\em Byzantine faults} and {\em client privacy leakage}.
The PS aims to robustly learn a model despite malicious updates from clients under Byzantine faults.
At the same time, it also faces privacy adversaries,
including external auditors who analyze the deployed model and participating clients who observe per-round updates.

{\em Byzantine fault.}
We follow the classic notion of Byzantine fault in distributed systems, where up to $f$ clients can be compromised
to send malicious messages. 
These clients, collectively denoted by $\calB$, can collude and adaptively send malicious messages based on previously observed information exchange, but cannot impersonate peers or observe or interfere with the communications of honest clients~\cite{lynch1996distributed}.
Let $\calG \triangleq [n] \setminus \calB$ denote the set of honest clients.
Due to the existence of Byzantine clients, it is impossible to minimize the fault-free objective $F$ \cite{su2020byzantine}. 
Therefore, we follow the mainstream in the literature of Byzantine-resilient distributed learning \cite{allouah2023fixing,allouah2023privacy,allouah2025towards,karimireddybyzantine}, and henceforth consider~\eqref{eq: Byzantine global obj}. 
\begin{align}
    \label{eq: Byzantine global obj}
    \byzobj (\bfthe)
    &\triangleq
    \frac{1}{|\calG|}\sum_{i\in \calG} F_i(\bfthe)
\end{align}

{\em User-level differential privacy.}
We consider central DP \cite{bassily2023user,levy2021learning}, where the PS is assumed to be trustworthy. 
To protect honest clients from having their participation inferred by privacy adversaries, we adopt Byzantine user-level DP in~\prettyref{def: dp federated learning}, with adjacency defined exclusively over the set of honest clients to account for Byzantine behavior.

\begin{definition}[$(\epsilon,\delta)$-Byzantine user-level DP]
    \label{def: dp federated learning}
    Let $\calX$ be the space of possible client datasets,
    $[n]$ be the set of client indices, which includes a set of honest clients $\calG$ and a set of Byzantine clients $\calB$
    such that $\calG \cap \calB = \emptyset$.
    \begin{itemize}[leftmargin=*]
       \item {\bf Byzantine strategy.}
        A Byzantine strategy $\mathfrak{B}$ is an arbitrary, possibly randomized and
        colluding rule whose round-$t$ actions depend on the honest datasets
        $\mathcal{D}_{\calG}$ only through what the server publishes in rounds $s < t$.
        \item {\bf Output.}
        For an algorithm $\calA$, let
        $\calA_{\mathfrak{B}}(\mathcal{D}_{\mathcal{G}})$
        denote its output, \ie, everything the server publishes over the course of its execution, when the honest clients hold the dataset collection
        $\mathcal{D}_{\mathcal{G}}
        = \{D_i\}_{i\in\mathcal{G}}
        \in \calX^{|\mathcal{G}|}$,
        and the Byzantine clients follow strategy $\mathfrak{B}$.
        \item {\bf Neighboring.}
        Two honest dataset collections
        $\mathcal{D}_{\mathcal{G}}$ and
        $\mathcal{D}_{\mathcal{G}}'$
        are said to be \emph{neighboring}, denoted by
        $\mathcal{D}_{\mathcal{G}}
        \sim
        \mathcal{D}_{\mathcal{G}}'$,
        if they differ in the dataset of exactly one honest client.
    \end{itemize}
    We say that $\calA$ satisfies
    $(\epsilon,\delta)$-Byzantine user-level DP if,
    for every Byzantine strategy $\mathfrak{B}$, every pair of neighboring collections
    $\mathcal{D}_{\mathcal{G}}
    \sim
    \mathcal{D}_{\mathcal{G}}'$,
    it holds for every measurable set $Q$,
    \begin{align*}
        \prob{\calA_{\mathfrak{B}}(\mathcal{D}_{\mathcal{G}}) \in Q}
        \le
        e^{\epsilon}
        \prob{\calA_{\mathfrak{B}}(\mathcal{D}_{\mathcal{G}}') \in Q}
        + \delta.
    \end{align*}
\end{definition}

\section{Lower Bound}
\label{sec: lower bound}
In this section, we introduce an information-theoretic lower bound in~\prettyref{thm: dp final lower bound cl} for any Byzantine-resilient ERM that ensures $(\epsilon, \delta)$ user-level DP under heterogeneous clients.
The full proofs are deferred to~\prettyref{app: lower bound}.

\begin{assumption}
\label{ass: bounded similarity}
There exist $B \ge 0$ and $G \ge 0$
such that $\forall \bfthe \in \reals^{d}$
\begin{align}
\label{eq: BG condition}
    \frac{1}{\abth{\calG}}\sum_{i \in \calG} \norm{\nabla F_i(\bfthe)- \nabla \byzobj (\bfthe)}^2 &\le B^2 \norm{\nabla \byzobj (\bfthe)}^2+ G^2.
\end{align} 
\end{assumption}

\prettyref{ass: bounded similarity} captures the heterogeneity across honest clients \cite{allouah2023robust,karimireddybyzantine,karimireddy2021learning}, which is what makes both Byzantine robustness and user-level DP fundamentally hard.
When $G=0$, all honest clients share common stationary points and become almost statistically indistinguishable.
Specifically, clients' full-batch gradients become identical when $B=G=0$, making the DP guarantees trivial.
Even when $B>0$ and $G=0$, the same degeneracy may arise, \eg, if each $F_i$ is strongly-convex or admits a unique stationary point.
In any of these regimes, Byzantine clients cannot deviate without being identified as outliers and are readily excluded.
Henceforth, we focus on the regime where $G > 0$.

\begin{theorem}
    \label{thm: dp final lower bound cl}
    For any given $L, B\ge 0$ and $G>0$, there exists a collection of $n$ functions $\{F_i\}_{i \in [n]}$,
    each defined over a dataset of $m$ samples, that are all
    i) $L$-smooth,  
    ii) $\mu$-strongly convex with $\mu = \Theta(L)$, 
    and iii) satisfy~\prettyref{ass: bounded similarity}.
    Let $m, n, d\ge 1$ and $ 1 \le f < n/2$.  
    Suppose that 
    $\epsilon, \delta \in (0,1)$ 
    and 
    $ 1/ 2^{(n - f)^{1 - \gamma}} \le 
    \delta 
    \le 1/ (n - f)^{1 + \gamma} 
    $,
    where $\gamma \in (0, 1)$.
    If any distributed algorithm $\calA$ is $(\epsilon, \delta)$-Byzantine user-level DP, then its output $\hat{\bfthe}$ satisfies
        \begin{align*}
        \Expect 
        \|\hat{\bfthe} - \bfthe^{\ast} \|_2^2
        &=
        \Omega
        \big(\frac{f}{n}
        \frac{G^2}{\mu^2}
        +
        \big(1 + \frac{f}{n}\big)
        \frac{d G^2 \log (1 / \delta)}{\mu^2 \epsilon^2 (n - f)^2
        } \big)
        , \\
        \Expect 
        \|\nabla \byzobj (\hat{\bfthe})\|_2^2
        &= 
        \Omega \big(
        \frac{f}{n}
        G^2
        +
        \big(1 + \frac{f}{n}\big)
        \frac{d G^2 \log (1 / \delta)}{\epsilon^2 (n - f)^2}
        \big)
        ,
        \end{align*}
   where $\hat{\bfthe}$ is the output of $\calA$, the expectation is taken over the randomness in $\calA$ and the data generation, and $\bfthe^{\ast} = \argmin_{\bfthe} \frac{1}{\abth{\calG}}\sum_{i \in \calG} F_i(\bfthe)$. 
\end{theorem}
The number of samples per client $m$ does not appear in the bound 
because we focus on the ERM problem with {\em arbitrary} data.
In the worst case, each client $i \in [n]$ may hold $m$ identical copies of a single point, with the point differing across clients.
User-level DP reduces to item-level DP over $n$ data points, and no improvement in $m$ is possible.
Note that the construction of~\prettyref{thm: dp final lower bound cl} is based on $m$ data samples.
The absence of $m$ is inherent to the worst-case ERM setting, rather than an artifact of our proof.
When no Byzantine clients are present ($f=0$), 
the bound reduces to $\Omega (d G^2 \log (1/\delta)/(\epsilon^2 n^2))$, matching the centralized item-level ERM lower bound in \cite{bassily2014private} up to the dependence on $G$, which is normalized to $G=\calO(1)$ in their construction.
The constraints on $\epsilon$ and $\delta$ are necessary only for
the lower bound \cite{steinke2016between}, which can be relaxed at the cost of a looser bound 
via techniques 
from \cite{bassily2014private}.
The proof of~\prettyref{thm: dp final lower bound cl} includes three subcases, each captured by a separate lemma. 
The theorem follows by taking a maximum. 
\prettyref{lmm: dp lower bound non-adversarial cl} recovers the DP-only cost 
in the absence of 
Byzantine clients.

\begin{lemma}[Non-Byzantine]
\label{lmm: dp lower bound non-adversarial cl}
    Suppose that the conditions in~\prettyref{thm: dp final lower bound cl} hold with $f = 0$,
    the output $\hat{\bfthe}$ of a $(\epsilon, \delta)$ user-level DP learning algorithm $\calA$ has an error at least
        \begin{align}
        \label{eq: private dp lower bound sc main text}
        \Expect 
        \|\hat{\bfthe} - \bfthe^{\ast} \|_2^2
        &=
        \Omega
        \big(
        \frac{d
         G^2 \log (1 / \delta)
        }{\mu^2 \epsilon^2 n^2}
        \big)
        ,~
        \Expect 
        \|\nabla F (\hat{\bfthe})\|_2^2
        =
        \Omega
        \big(
        \frac{d
        G^2 \log (1 / \delta)
        }{\epsilon^2 n^2}
        \big)
        .
        \end{align}
\end{lemma}
To show~\prettyref{lmm: dp lower bound non-adversarial cl}, we reduce the distributed learning problem to a centralized DP mean estimation of one-way marginals.
The key technique 
is to adapt~\cite[Proposition 2]{levy2021learning} to our setting: user-level DP mean estimation is equivalent to the item-level mean estimation problem, where each item is the mean of a client's local dataset, thereby reducing a 
multi-point estimation problem to a single-point one.
This reduction clarifies the earlier intuition: the $m$ data points held by each client contain no more private information than their mean, which explains why $m$ does not appear in the bound.
\prettyref{lmm: non private lower bound cl} characterizes the sole cost of Byzantine resilience in the absence of DP constraints.
Its proof follows from the standard roadmap in the Byzantine-resilient learning literature~\cite{allouah2023fixing,allouah2023privacy,karimireddy2021learning}.
\begin{lemma}[Non-private] 
\label{lmm: non private lower bound cl} 
Suppose that the conditions in~\prettyref{thm: dp final lower bound cl} hold with $ 1 \le f < n/2$,
the output $\hat{\bfthe}$ of a $(\epsilon, \delta)$ user-level DP learning algorithm $\calA$ has an error at least
\begin{align}
    \Expect 
    \|\hat{\bfthe} - \bfthe^{\ast} \|_2^2
    & = \Omega 
    \big(
    \frac{f}{n - f}\frac{G^2}{\mu^2}\big),  
    ~\Expect 
    \|\nabla \byzobj (\hat{\bfthe})\|_2^2
     = \Omega\big(\frac{f}{n - f}G^2\big).
\end{align}
\end{lemma}
\begin{lemma}[Coupled]
    \label{lmm: adversarial dp lower bound cl} 
    Suppose that the conditions in~\prettyref{thm: dp final lower bound cl} hold with $ 1 \le f < n/2$,
    the output $\hat{\bfthe}$ of a $(\epsilon, \delta)$ user-level DP learning algorithm $\calA$ has an error at least
        \begin{align}
        \label{eq: adversarial dp lower bound sc main text}
        \Expect 
        \|\hat{\bfthe} - \bfthe^{\ast} \|_2^2
        &=
        \Omega \big(
        \frac{f}{n}
        \frac{d G^2 \log (1 / \delta)}{\mu^2 \epsilon^2 (n - f)^2}
        \big)
        ,~
        \Expect
        \|\nabla \byzobj (\hat{\bfthe})\|_2^2
        =
        \Omega \big(
        \frac{f}{n}
        \frac{d G^2 \log (1 / \delta)}{\epsilon^2 (n - f)^2}
        \big)
        .
        \end{align}
\end{lemma}
Compared with~\eqref{eq: private dp lower bound sc main text},~\eqref{eq: adversarial dp lower bound sc main text} inflates the DP error by a factor of Byzantine fraction $f/n$.
The key step 
is to construct two adversarial configurations that are indistinguishable to any user-level DP algorithm, yet have well-separated optima. 
We partition the $n$ clients into three groups 
whose local dataset means point in opposing directions: $+G_0\Indc$, $-G_0\Indc$, and $\bm{0}$, where $G_0 = \Theta(G)$.
We construct hard instances by combining these groups in different ways. 
By comparing each instance to a common reference, we show that no distributed algorithm can distinguish them, 
which reduces the distributed learning problem to a centralized, item-level DP mean-estimation task on $n - f$ items.

\section{Algorithm:~\byzalg}
\label{sec: algorithm}

\setlength{\columnsep}{1.5cm}
\begin{algorithm}[!t]
    {\bf Input:}
    $\bfthe^0$,
    $\m_i^0 = \bm{0}$, $\forall i \in \calG$,
    $\tilde R^{-1} = \bm{0}$,
    $R$,
    $\sigma_{dp,t}^2$,
    $\eta_t$,
    $\beta_t$, $\alpha_t$,
    and $T$; \\
     {\bf Output:}
    $\bfthe^{T}$.

    \For{$t = 0, \ldots, T-1$}{
    \begin{multicols}{2}
    \tcc{\blue On all clients}
    \For{$i = 1, \cdots, n$}{
    Download $\bfthe^t$ from the PS \;
        $\hat{\g}_i^{t}
        \gets
        {\tt clip} (\nabla \ell_i (\bfthe^t; \xi_i^t), C)$,
        where
        ${\tt clip}(\bm{v}, C) = 
        \bm{v} \cdot \min \big\{
        1, {C}/{\norm{\bm{v}}}
        \big\}$;
        \\
    $\m_i^{t} \gets \alpha_t \m_i^{t - 1} + (1 - \alpha_t) \hat{\g}_i^{t}$\;
      $\hat{\m}_i^{t}
      =
      \begin{cases}
      \m_i^{t},
        ~&\text{if}~ i \in \calG ;\\
        \ast,
        ~&\text{otherwise}.
    \end{cases}$ \;
    Send $\hat{\m}_i^{t}$ to the PS.
    }
    \columnbreak
   \tcc{\blue On the PS}
    \For{$i= 1, \cdots, n$}{
    $\M_i^{t} \gets \beta_t \tilde R^{t - 1} + (1 - \beta_t) \hat{\m}_i^{t}$\;  
    }
    
    $R^t \gets R({\M}_1^t, \ldots, {\M}_n^t)$ 
    
    $\tilde R^t \gets R^t + \X_{dp}^t$, where $\X_{dp}^{t} \sim \calN (0, \sigma^2_{dp, t}\, \identity)$
    
    $\bfthe^{t+1} \gets \bfthe^{t} - \eta_t \tilde R^t$\;
    \end{multicols}
    
    }
    \caption{\byzalg}
    \label{alg: byzantine dp learning base stochastic v2}
\end{algorithm}
In this section, we propose 
{\bf Byz}antine-resilient 
{\bf U}ser-{\bf l}evel 
{\bf D}ifferentially 
{\bf P}rivate 
Learning (\byzalg), formally presented in~\prettyref{alg: byzantine dp learning base stochastic v2}.
\byzalg~departs from FedAvg by tackling the dual challenges of Byzantine attacks and user-level DP together. 
At a high level,~\byzalg~achieves Byzantine resilience through a nonlinear robust aggregator combined with a client-side momentum scheme that mitigates stochastic gradient noise. 
\byzalg~introduces a server-side momentum to regulate the global update.
While the fault-free recipe for user-level central DP relies on injecting central noise directly into a linear simple mean \cite{bassily2023user,levy2021learning}, this approach is insufficient for convergence with nonlinear robust aggregators; see~\prettyref{rmk: nonconvex convergence} for details.

A key innovation in~\byzalg~is the use of server-side momentum to recalibrate each client's message, thereby controlling its global impact and reducing privacy sensitivity without sacrificing robustness.
This recalibration is enabled {\em exclusively} by user-level DP; the gradient is the only viable data representation for item-level DP, lacking such architectural flexibility.

\paragraph{Client-side.}
Each honest client $i \in \calG$ downloads $\bfthe^t$ from the PS, 
computes a clipped stochastic gradient $\hat{\g}_{i}^{t} = {\tt clip}(\nabla \ell_i (\bfthe^t; \xi_i^t), C)$, where ${\tt clip}(\bm{v},C) = \bm{v} \cdot \min \{1, C/\|\bm{v}\|_2\}$ (line 6),
and updates its local momentum $\m_i^{t} = \alpha_t \m_i^{t - 1} + (1 - \alpha_t) \hat{\g}_i^{t}$ (line 7).
Clipping bounds each client's influence on the global model,
which is important for our privacy analysis.
Byzantine clients may report arbitrary values $\ast$ in place of $\m_i^t$.
One-step gradient descent is commonly adopted in Byzantine-resilient FL \cite{allouah2023fixing,allouah2025adaptive,allouah2023robust,allouah2023privacy,karimireddybyzantine,karimireddy2021learning}; extending to multi-step local updates is left to future work.

\paragraph{Server-side.}
The PS computes $\M_{i}^{t} = \beta_t \tilde R^{t-1} + (1 - \beta_t) \hat{\m}_{i}^{t}$ for each client $i \in [n]$ (lines 11-13), where $\tilde R^{t-1}$ is the previous-round privatized aggregate and $\beta_t \in (0,1)$.
Similar server-side momentum has been adopted in \cite{cheng2024momentum,xu2021fedcm} yet
to mitigate the client-drift issue in non-Byzantine and non-private settings.
Such a momentum mechanism enables contraction $\M_i^{t} - \M_{j}^{t} = (1 - \beta_t) (\hat{\m}_{i}^{t} - \hat{\m}_{j}^{t})$, which shrinks the distance between honest clients by a factor $1 - \beta_t$
and reduces the privacy sensitivity of the global aggregation, formally shown in~\prettyref{prop: trimmed mean sensitivity}.
Since $\tilde R^{t-1}$ is already privatized, the DP post-processing property ensures that all subsequent computations remain private.
In line 14, the PS aggregates the server-side momentum via a nonlinear robust aggregator $R(\cdot)$, defined in~\prettyref{def: robust agg}.

\begin{definition}[$(f, \kappa)$-robust~\cite{allouah2023fixing}]
    \label{def: robust agg}
    Let $f < n / 2$ and $\kappa \ge 0$. 
    An aggregation rule $R$ is said to be $(f, \kappa)$-robust if for any vectors $\x_1, \ldots, \x_n \in \reals^d$, and any set $S \subseteq [n]$ of size $n - f$,
    it holds that 
    \[
    \norm{R(\x_1, \ldots, \x_n) - \overline{\x}_S}^2 \le \frac{\kappa}{\abth{S}} \sum_{i \in S} \norm{\x_i - \overline{\x}_S}^2,
    \]
    where $\overline{\x}_S = \frac{1}{\abth{S}} \sum_{i \in S} \x_i$. 
\end{definition}
We refer to $\kappa$ as a robust coefficient.
Robust aggregators aim to filter out the adversarial updates and have been widely adopted in Byzantine-resilient algorithm design \cite{allouah2023fixing,blanchard2017machine,chen2017distributed,karimireddybyzantine,karimireddy2021learning,su2016fault,xie2019zeno,yin2018byzantine}. 
Our analysis can be easily extended to any such $R$.
However, to obtain the optimal
upper bound, we adopt coordinate-wise trimmed mean ({\tt CWTM}) \cite{yin2018byzantine}, defined in~\prettyref{def: cwtm}, 
as the robust aggregator.
\begin{definition}[{\tt CWTM}]
    \label{def: cwtm}
    Let $\x \in \reals^d$, we define $[\x]_r$ the $r$-th coordinate of $\x$.
    Given the input vectors $\x_1, \ldots, \x_n \in \reals^d$,
    we let $\pi_r$ denote a permutation on $[n]$ that sorts the $r$-th coordinate of the input vector in non-decreasing order, \ie, $[\x_{\pi_{r}(1)}]_r \le [\x_{\pi_{r}(2)}]_r  \le \ldots \le [\x_{\pi_{r}(n)}]_r$.
    Then, the coordinate-wise trimmed mean ${\tt CWTM} (\x_1, \ldots, \x_n) \in \reals^d$ is defined coordinate-wise by
    \[
    \qth{
    {\tt CWTM} (\x_1, \ldots, \x_n)
    }_{r}
    \triangleq
    \frac{1}{n - 2 f}
    \sum_{i=f+1}^{n - f}
    \qth{\x_{\pi_{r} (i)}}_r.
    \]
\end{definition}
Finally, the PS forms the privatized aggregate $\tilde R^{t} = R^{t} + \X_{dp}^t$ with $\X_{dp}^{t}\sim \calN (0, \sigma_{dp,t} \identity)$ and updates $\bfthe^{t+1} = \bfthe^{t} - \eta_t \tilde R^t$.
The value of $\sigma_{dp,t}$ will be specified in~\prettyref{sec: privacy results}.

\section{Convergence Analysis of~\byzalg~(Upper Bound)}
\label{sec: convergence}
In this section, we analyze the convergence of~\byzalg.
Missing proofs, intermediate results, and additional privacy analysis are deferred to the Appendix.

\subsection{Additional Assumptions}
\label{sec: convergence assumptions}
Let $\calF^t$ define the natural filtration generated by the randomness up to round $t$.
\begin{assumption}
\label{ass: smmothness}
    Each local objective function
    $\nabla F_{i}(x)$ is $L$-Lipschitz,
    \ie,
    \[
        \norm{\nabla F_{i}(\bfthe_1)-\nabla F_{i}(\bfthe_2)}\le L \norm{\bfthe_1-\bfthe_2},
        ~\forall \bfthe_1,~\bfthe_2\in \reals^{d},~\text{and}~\forall~i\in [n].
    \] 
\end{assumption}
\begin{assumption}
\label{ass: bounded variance client-wise}
Stochastic gradients $\nabla \ell_i(\bfthe^t;\xi_i^t)$ are unbiased with bounded variance, \ie,  
\[  
    \expect{\nabla \ell_i(\bfthe^{t};\xi_i^t) \mid \calF^t}=\nabla F_i(\bfthe^{t})
    ~\text{and}~
    \expect{\norm{\nabla \ell_i(\bfthe^{t};\xi_{i}^{t})-\nabla F_i(\bfthe^{t})}^2 \mid \calF^t} \le \sigma^2,
    ~\forall~
    i\in[n].
\] 
\end{assumption}
\begin{assumption}
    \label{ass: bounded sto gradient cl}
    Stochastic gradients 
    $\nabla \ell_i (\bfthe, \xi)$ are uniformly bounded in $l_2$ norm
    \(
        \norm{
        \nabla \ell_i (\bfthe, \xi)
        }
        \le
        C,~\forall \bfthe \in \reals^d 
        ~\text{and}~
        \xi.
    \)
\end{assumption}

Assumptions~\ref{ass: smmothness} and~\ref{ass: bounded variance client-wise} are common in FL analysis \cite{karimireddy2021learning,li_federated_2020,wang_tackling_2020}.
\prettyref{ass: bounded sto gradient cl} is widely adopted by DP algorithm analysis \cite{allouah2023privacy,bassily2014private,bassily2023user,ghazi2021user,levy2021learning,wang2022differentially}.
We conjecture that~\prettyref{ass: bounded sto gradient cl} is imposed primarily for technical convenience and leave its relaxation to future work.

\subsection{Privacy Results of~\byzalg}
\label{sec: privacy results}
We start from the privacy analysis of~\byzalg.
Since clients influence the global model through the robust aggregator, the key step is characterizing its user-level sensitivity.
\prettyref{def: user-level sensitivity robust aggregator} adapts the standard $\ell_2$ sensitivity \cite{dwork2014algorithmic} to the Byzantine setting.
As Byzantine clients may behave arbitrarily, we measure sensitivity with respect to replacing a single honest client's update.
\begin{definition}[Byzantine-robust user-level sensitivity]
\label{def: user-level sensitivity robust aggregator}
    Let $[n] = \calG \cup \calB$ be a union of honest clients $\calG$ and Byzantine clients $\calB$ such that $\abth{\calB} \le f$.
    The Byzantine-robust user-level sensitivity of a robust aggregator
    $R : \reals^{n \times d} \to \reals^d$,
    operating on $n$ inputs of which at most $f < n/2$ are Byzantine,
    is
    \begin{align*}
     \Delta(R)~ \triangleq~ \max_{\calB \subset [n], \abth{\calB} \le f}
    \sup_{\X_{\calB}} \sup_{\X_{\calG} \simeq \X_{\calG^\prime}}
    \|R(\X_{\calG} \cup \X_{\calB}) - R(\X_{\calG^\prime} \cup \X_{\calB})\|_2,   
    \end{align*}
    where $\X_{\calG}, \X_{\calB}$ are input collections from the respective sets, and $\X_{\calG} \simeq \mathbf{X}_{\calG^{\prime}}$ denotes honest client collections differing by exactly one input.
\end{definition}
Our analysis extends to any robust aggregator in~\prettyref{def: robust agg}, 
but {\tt CWTM} enables us to attain the optimal convergence rate in~\prettyref{sec: convergence results}.
\begin{lemma}[Per-dimension {\tt CWTM}]
    \label{lmm: 1-lipschitz cwtm}
    Let $n \ge (2 + \nu) f$, where $\nu > 0$. 
    Fix a coordinate $r$, 
    for any $a_{1} \le a_{2} \le \cdots \le a_{n-1}$ in $\reals^d$,
    let ${\tt TM}_{r}(h)$ define the coordinate-$r$ output of {\tt CWTM} when inserting value $h$.
    For any $h, h^{\prime} \in \reals$, it holds that
    \[
    \abth{{\tt TM}_{r}(h) - {\tt TM}_{r}(h^{\prime})} \le \frac{\abth{h - h^{\prime}}}{n}
    \sqrt{1 + \frac{4 (2 + \nu)^2}{\nu^2} \frac{f}{n}} 
    .
    \]
\end{lemma}
The proof of~\prettyref{lmm: 1-lipschitz cwtm} characterizes how
inserting an honest client update affects the {\tt CWTM} output by distinguishing whether the update lies inside or outside the trimmed set.
Since it holds coordinate-wise,
and honest updates are bounded by $C$ in $\ell_2$ norm, aggregating across coordinates yields a sensitivity without explicit dependence on the dimension $d$.
This intuition is formalized in~\prettyref{prop: trimmed mean sensitivity}.
\begin{proposition}[{\tt CWTM} sensitivity]
    \label{prop: trimmed mean sensitivity}
    Let $n \in \naturals$.
    If the input to the robust aggregator is $ \x_i = \beta_t \tilde R^{t-1} +(1 - \beta_t) {\m}_i^{t}$, and $n \ge (2 + \nu) f$,
    where $\norm{\m_i^{t}} \le C$ for $i \in \calG$ and $\nu > 0$,
    the Byzantine-robust user-level sensitivity of {\tt CWTM} in round $t$ is 
    \[
    \Delta_{t} ({\tt CWTM})
    =
    \frac{2 (1 - \beta_t) C}{n}
    \sqrt{1 + \frac{4 (2 + \nu)^2}{\nu^2} \frac{f}{n}} 
    .
    \]
\end{proposition}
\begin{theorem}[User-level DP noise]
    \label{thm: cl rdp guarantees}
    Let $\delta \in (0,1)$ and $\epsilon > 0$ such that $\epsilon \le 8 \log (1 / \delta)$.
    \prettyref{alg: byzantine dp learning base stochastic v2} is a $(\epsilon, \delta)$ user-level DP algorithm with {\tt CWTM} as the robust aggregator when 
    \(
        \sigma_{dp,t}^2
        =
        \frac{32 (1 - \beta_t)^2 C^2 T \log (1 / \delta)}{n^2 \epsilon^2}
        \pth{1 + \frac{4 (2 + \nu)^2}{\nu^2} \frac{f}{n}} 
        .
    \)
\end{theorem}
\prettyref{thm: cl rdp guarantees} combines the per-round dimension-independent sensitivity in~\prettyref{prop: trimmed mean sensitivity} with the cross-round ~\renyi~DP composition~\cite{mironov2017renyi}, yielding the $\sigma_{dp,t}^2 \propto T$ scaling without explicit dependence on $d$.
The noise variance additionally inherits a $(1 - \beta_t)^2$ factor from server-momentum aggregation, since $\|\M_i^t - \M_j^t\|^2 = (1-\beta_t)^2 \|\m_i^t - \m_j^t\|^2$ for $i,j \in \calG$.
Such a dependence effectively reduces privacy sensitivity when $(1 - \beta_t)$ is small
and is crucial for convergence; see~\prettyref{rmk: nonconvex convergence} for details.

\subsection{Convergence Results of~\byzalg}
\label{sec: convergence results}
In this subsection, we provide convergence results for~\byzalg.
The non-convex case uses constant $\alpha_t \equiv \alpha$ and $\eta_t \equiv \eta$, while the strongly-convex case keeps time-varying schedules.
Let 
$\overline{\m}^t \triangleq \frac{1}{\abth{\calG}} \sum_{i \in \calG} \m_i^t$
and $F^{\ast}_{\calG} = \min_{\bfthe} F_{\calG}(\bfthe)$.
In~\prettyref{lmm: sc momentum drift v2}, we show that client-side momentum mitigates the stochastic gradient noise. 
Take non-convex objectives as an example.
\begin{lemma}[Non-convex local momentum drift]
    \label{lmm: sc momentum drift v2}
    Under Assumptions~\ref{ass: bounded similarity},~\ref{ass: smmothness}, and~\ref{ass: bounded variance client-wise}.
    When $\byzobj$ is non-convex, it holds for any $\alpha_t = \alpha \in (0,1)$ that
       \begin{small}
       \begin{align}
       \label{eq:momentum non-convex expansion}
        \frac{1}{\abth{\calG}}
        \sum_{i \in \calG}
        \Expect{
        \norm{\m_i^t - \overline{\m}^t}^2}
        &\le
        \pth{3 + \frac{3}{\abth{\calG}}}
        \pth{1 - \alpha^2}
        \sigma^2
        +
        3
        G^2
        +
        3 B^2 
        (1 - \alpha)
        \sum_{r=0}^t
        \alpha^{t - r}
        \Expect{\norm{
        \nabla \byzobj (\bfthe^r)
        }^2}
        .
    \end{align}
    \end{small}
\end{lemma}
\eqref{eq:momentum non-convex expansion} shows that the stochastic variance term scales as $(1 - \alpha^2)$.
Setting $1 - \alpha = \Theta(\eta L)$ makes this factor small, effectively dampening its effect.
The proof of~\prettyref{lmm: sc momentum drift v2} follows the standard Polyak momentum analysis
by unrolling across all rounds~\cite{allouah2023fixing,allouah2023privacy,karimireddy2021learning}.
An analogous strongly-convex result uses single-round unrolling; see~\prettyref{app: stochastic convergence strongly-convex v2} for details.
We establish our convergence bounds in Theorems~\ref{thm: sc stochastic gradient convergence v2 main text} and~\ref{thm: non-convex learning convergence v2 main text}
by constructing separate Lyapunov functions for strongly-convex and non-convex objectives to account for their different learning rate and momentum schedules.

    \begin{theorem}[Strongly-convex, Informal]
    \label{thm: sc stochastic gradient convergence v2 main text}
    Suppose that Assumptions~\ref{ass: bounded similarity},~\ref{ass: smmothness},~\ref{ass: bounded variance client-wise}, and \ref{ass: bounded sto gradient cl} hold.
    When $\eta_t = \frac{10}{\mu (t + a \frac{L}{\mu})}$ such that 
    $a \ge 49000$, 
    $1 - \alpha_t= \Theta(\eta_t L)$,
    $1 - \beta_t = \Theta(\eta_t L)$, 
    $B^2 \le 1/(600 \kappa)$, 
    and $n \ge (2 + \nu) f$, where $\nu > 0$, it holds for~\prettyref{alg: byzantine dp learning base stochastic v2} with {\tt CWTM} as the robust aggregator that
\begin{align}
    \label{eq:sc main text}
    \Expect{\big\|\bfthe^T - \bfthe^{\ast} \big\|_2^2}
    &\lesssim
    {\blue \frac{\kappa G^2}{\mu^2} }
    +
    {\color{olive}
    \pth{\frac{L }{\mu}}
    \frac{\log(1 / \delta) C^2 d}{\mu^2 \epsilon^2 n^2} \pth{1 + \frac{f}{n}}}
    +
    L (1 + \kappa)
    \pth{1 + \frac{1}{\abth{\calG}}}
    \frac{\sigma^2 }{\mu^3 T}\\
    \notag
    &
    +\frac{L^2 (\expect{\byzobj (\bfthe^0)} - \byzobj^{\ast})}{\mu^3 T^2} 
    +
    \frac{L \sigma^2}{\mu^3 T^2 \abth{\calG}} 
    +
    \frac{L \kappa}{\mu^3 T^2} 
  \qth{
     \pth{1 + \frac{1}{\abth{\calG}}}
    \sigma^2
    +
    G^2
    }
    .
\end{align}
\end{theorem}

\begin{theorem}[Non-convex, Informal]
    \label{thm: non-convex learning convergence v2 main text}
    Suppose that Assumptions~\ref{ass: bounded similarity},~\ref{ass: smmothness},~\ref{ass: bounded variance client-wise}, and~\ref{ass: bounded sto gradient cl} hold.
    Let $1 - \alpha = \Theta(\eta L)$, $1 - \beta = \Theta(\eta L)$, such that
    $\eta < 1/ (5100 L)$ and $B^2 \le 1/(23625 \kappa)$.
    When $n \ge (2 + \nu) f$, where $\nu > 0$ is an absolute constant,
    it holds for~\prettyref{alg: byzantine dp learning base stochastic v2} with {\tt CWTM} as the robust aggregator that
\begin{align}
\notag
    &\frac{1}{T}
    \sum_{t=0}^{T-1}
    \Expect{\norm{\nabla \byzobj (\bfthe^{t})}^2} 
    \lesssim
    \frac{ \expect{\byzobj (\bfthe^0)} - \byzobj^{\ast}}{\eta T}
    +
    \frac{\eta L \sigma^2}{\abth{\calG}} 
    +
    \eta L
    \kappa 
    (1 + \frac{1}{\abth{\calG}})
    \sigma^2
    +
    \kappa  G^2 
    +
    \frac{d \sigma_{dp}^2 }{ \eta L} \\
     \label{eq:non-convex main text}
    &\overset{\sigma_{dp} = \Theta(1 - \beta)}{\lesssim}
    \frac{ \expect{F (\bfthe^0)} - F^{\ast}}{\eta T}
    +
    \eta L \sigma^2
    \Big(
    \frac{1}{\abth{\calG}} 
    +
    \kappa 
    (1 + \frac{1}{\abth{\calG}})
    \Big)
    +
    \kappa  G^2 
    +
    \frac{(1 - \beta)^2 T}{\eta L}
    \frac{d C^2 \log(\frac{1}{\delta}) (1 + \frac{f}{n}) }{ n^2 \epsilon^2} \\
    \label{eq:non-convex main text final}
    &
    \overset{(1 - \beta) = \Theta(\eta L)}{\lesssim}
    \frac{ \expect{\byzobj (\bfthe^0)} - \byzobj^{\ast}}{\eta T}
    +
    \eta L \sigma^2
    (
    \frac{1}{\abth{\calG}} 
    +
    \kappa 
    (1 + \frac{1}{\abth{\calG}})
    )
    +
    {\blue \kappa  G^2 }
    +
    {\color{olive}
    \frac{\eta L T  d C^2 \log(1/\delta) (1 + \frac{f}{n}) }{ n^2 \epsilon^2}}
    .
\end{align}
\end{theorem}

\begin{corollary}[Informal]
    \label{cor: non-convex convergence v2 main text}
    Suppose that Assumptions~\ref{ass: bounded similarity},~\ref{ass: smmothness},~\ref{ass: bounded variance client-wise}, and~\ref{ass: bounded sto gradient cl} and the conditions in~\prettyref{thm: non-convex learning convergence v2 main text} hold.
    The choice of learning rate is formally defined in~\prettyref{app: nonconvex convergence learning rate}.
    When $n \ge (2 + \nu) f$, where $\nu > 0$ is an absolute constant,
    it holds for~\prettyref{alg: byzantine dp learning base stochastic v2} with {\tt CWTM} as the robust aggregator that

    \begin{align*}
        &\frac{1}{T}
        \sum_{t=0}^{T-1}
        \Expect{\norm{\nabla \byzobj (\bfthe^{t})}^2} 
        \lesssim
        {\blue \kappa  G^2} +
        {\color{olive}
        \sqrt{L (\expect{\byzobj (\bfthe^0)} - \byzobj^{\ast})
        \frac{ \log(1 / \delta) C^2 d }{\epsilon^2 n^2}
        \pth{1 + \frac{f}{n}}}}
        \\
        &\qquad \qquad \qquad \quad
        +
        \frac{\sigma}{\sqrt{T}} 
        \sqrt{ L ( \expect{\byzobj (\bfthe^0)} - \byzobj^{\ast})
        \pth{
        \frac{1}{\abth{\calG}}
        + \kappa
        \pth{1 + 
        \frac{1}{\abth{\calG}}}}} 
        + 
        \frac{L ( \expect{\byzobj (\bfthe^0)} - \byzobj^{\ast})}{T}
        .
    \end{align*}
\end{corollary}

\begin{remark}[\bf Optimal bound for strongly-convex objectives]
    \label{rmk: sc optimal}
    We remark on~\prettyref{thm: sc stochastic gradient convergence v2 main text}, which establishes the convergence guarantees for a $\mu$ strongly-convex objective $\byzobj$.
    \begin{itemize}[leftmargin=*]
        \item {\bf Order-optimal bound.}
        \prettyref{thm: sc stochastic gradient convergence v2 main text} establishes the full convergence bound for~\byzalg~under a $\mu$-strongly convex objective.
        The colored terms in~\eqref{eq:sc main text} are the non-vanishing errors.
        \begin{itemize}[leftmargin=*]
            \item {\em DP terms.}
            The hard instances in~\prettyref{lmm: dp lower bound non-adversarial cl} and in~\prettyref{lmm: adversarial dp lower bound cl} satisfy $G = \Theta (C)$ when $f = \Theta(n)$, yielding a matching lower bound $\Omega \left( \frac{ (1 + f/n) d C^2 \log(1/\delta)}{\mu^2 \epsilon^2 (n-f)^2} \right)$ up to constants and a factor $L/\mu$;
            see~\prettyref{app: matching optimality condition} for details.

            \item {\em Byzantine-only term.}
            {\tt CWTM} aggregator achieves a 
            robustness coefficient of $\kappa = \Theta (f/n)$~\cite{allouah2023fixing}, yielding a matching bound $\calO(\kappa G^2/ \mu^2) = \calO (f/n \cdot G^2/ \mu^2)$;
            see~\prettyref{cor:cwtm optimal} for details.
        \end{itemize}
        Together, our upper bound is order-optimal with respect to the lower bound in~\prettyref{sec: lower bound}.
        In the absence of the non-vanishing error terms, our $\calO (1 / T)$ rate for strongly-convex objectives also matches the optimal rate for stochastic first-order methods~\cite{rakhlin2012making}.

        \item {\bf Privacy-utility tradeoff.}
        For the non-vanishing errors, we have both DP error and Byzantine residual terms, where the former dominates when $d \gtrsim n$.
        Fortunately, the large scale of FL systems can effectively mitigate the impact of DP noise on the final model utility~\cite{kairouz2021advances}. 
        On the other hand, the Byzantine residual error becomes negligible when the number of Byzantine clients $f$ is small or when the clients' data is not highly heterogeneous.

        \item {\bf Comparisons with~\cite{allouah2025towards}.}
        Recall that \cite{allouah2025towards} guarantees 
        user-level DP without a trusted server in the presence of Byzantine faults. We establish an improved convergence bound in the central DP setting.
        Ignoring the Byzantine-only error and $\mu$, we have the following comparisons:
        \begin{align*}
            \text{\bf Ours}&:~\calO(\frac{d C^2 \log (1 / \delta)}{n^2 \epsilon^2} + \frac{f d C^2 \log (1 / \delta)}{n^3 \epsilon^2}); \\
            \text{\bf \cite{allouah2025towards}}&:~
            \calO (\frac{d C^2 \log (1/\delta)}{n^2 \epsilon^2} + \frac{f d C^2 \log (1/\delta)}{n^2 \epsilon^2} + \frac{f C^2 \log (1/\delta)}{n \epsilon^2}).
        \end{align*}
        
        The two bounds coincide in the Byzantine-free case $f=0$. 
        Yet, our bound strictly improves whenever $f \ge 1$.
        The dominant Byzantine-DP term improves from $fd/n^2$ to $fd/n^3$ when $d \gtrsim n$, 
        and from $f/n$ to $f/n^2$ when $d \lesssim n$.
        Moreover, their robust aggregator costs $\calO (f (nd^2 + d^3))$ to compute,
        while {\tt CWTM} runs efficiently in $\calO(nd \log(n))$.
    \end{itemize}
    
\end{remark}

\smallskip
\begin{remark}[\bf Results for non-convex objectives]
    \label{rmk: nonconvex convergence}
   We remark on \prettyref{thm: non-convex learning convergence v2 main text} and \prettyref{cor: non-convex convergence v2 main text}, which establish the convergence guarantees for a non-convex objective $\byzobj$.
    \begin{itemize}[leftmargin=*]
        \item {\bf Necessity of server-side momentum.}
        \prettyref{thm: cl rdp guarantees} establishes that $\sigma_{dp} = \Theta(1 - \beta)$.
        Substituting into~\prettyref{thm: non-convex learning convergence v2 main text}
        and setting $(1 - \beta) = \Theta(\eta L)$ reduces the DP error to a $\calO(\eta L T)$ scaling.
        Without server momentum ($\beta = 0$),
        this term scales as $\calO(T/ (\eta L))$ and diverges, \eg, as $\calO(T^{3/2})$ when $\eta = \Theta(1/\sqrt{T})$.
        Thus, server momentum is a structural requirement for a non-trivial DP guarantee.

        \item {\bf Non-convex bound.}
        Without the non-vanishing error terms, our $\calO(1/\sqrt{T})$ convergence rate for non-convex objectives matches the optimal bound for stochastic first-order methods~\cite{arjevani2023lower}.
        Similar to~\prettyref{rmk: sc optimal}, {\tt CWTM} achieves an order-optimal Byzantine residual error of $\calO(\kappa G^2) = \calO(f/n \cdot G^2)$.
        However, a gap remains regarding the DP residual error.
        Ignoring the Byzantine residual, 
        \eqref{eq:non-convex main text final} scales as $\calO(\frac{1}{\eta T} + \eta L (1 + T))$.
        Applying the AM-GM inequality yields an optimal learning rate of $\eta = \Theta(\frac{1}{\sqrt{L T(1 + T)}})$, subject to the constraint $\eta < 1/(5100 L)$.
        However, this optimal $\eta$ inevitably couples the DP noise term with the initialization error under a square root in~\prettyref{cor: non-convex convergence v2 main text}, preventing a strict match with our lower bound in~\prettyref{sec: lower bound}.
        The appearance of the DP noise term inside a square root is widely observed in the DP non-convex learning literature; see, \eg, \cite{allouah2023privacy,allouah2025towards,arora2023faster,bassily2021differentially,levy2021learning}.
    \end{itemize}
\end{remark}

\section{Numerical Experiments}
\label{sec: numerical experiment}
\begin{figure}[!t]
    \begin{subfigure}{\textwidth}
    \centering
    \includegraphics[width=\linewidth, trim=0.3cm 0.4cm 0.2cm 0.2cm, clip]{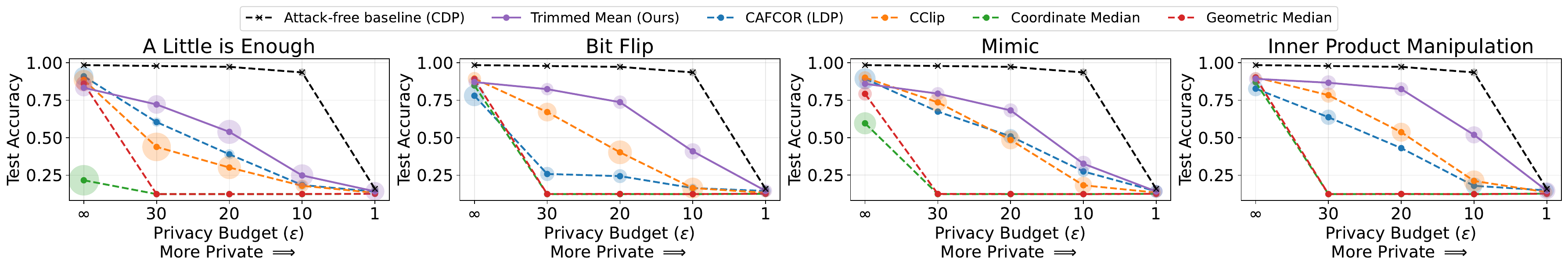}
    \caption{\small
    Test accuracy results on the MNIST dataset.
    }
    \label{fig: mnist robustness and privacy}    
    \end{subfigure}
    \begin{subfigure}{\textwidth}
    \centering
    \includegraphics[width=\linewidth]{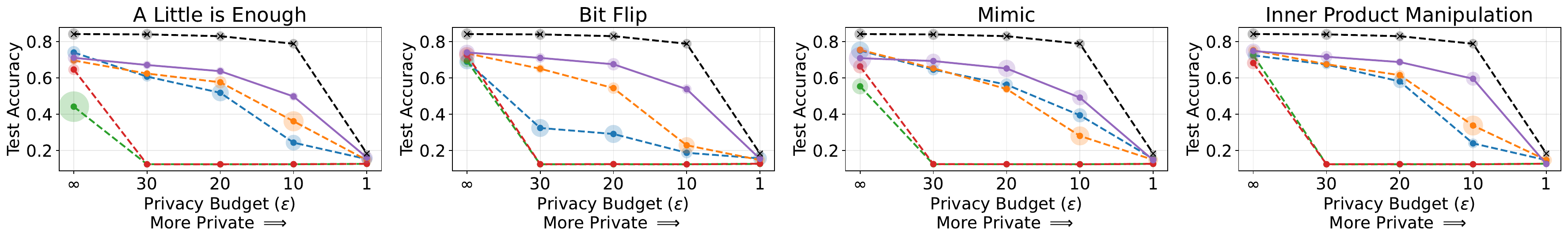}
    \caption{\small
    Test accuracy results on the Fashion-MNIST dataset.
    }
    \label{fig: fmnist robustness and privacy}    
    \end{subfigure}
    \caption{\small
    Test accuracy results and comparisons on real-world datasets.
    The results are averaged over the last 50 rounds across $4$ different random seeds.
    All the experiments run for a total of $800$ rounds.
    The radius of the shaded circle denotes the standard deviation at a given privacy budget $\epsilon$.
    }
    \label{fig:final results main text}
\end{figure}
{\bf Overview.}
In this section, we evaluate the performance of~\byzalg~with {\tt CWTM} against other state-of-the-art (SOTA) robust aggregators.
The hyperparameter setups and additional experiment details are deferred to~\prettyref{app: experiments}.
We consider an FL system of $1000$ clients with $250$ clients corrupted by a variety of Byzantine adversaries.
The image classification tasks use CNNs and are based on the MNIST~\cite{lecun2010mnist} and the Fashion-MNIST~\cite{xiao2017fashion} datasets.
While these tasks are relatively simple in non-adversarial settings, they become significantly more challenging in the presence of Byzantine clients~\cite{guerraoui2021distributed,karimireddy2021learning} and DP~\cite{allouah2023privacy,allouah2025towards}.
All of them include $10$ classes of images of different categories.
We partition the dataset to clients such that each client holds two classes of images.

{\bf Attacks and Benchmark algorithms.}
We evaluate four SOTA attacks,
including
a little is enough \cite{baruch2019little},
bit-flip attack~\cite{allenbyzantine},
mimic attack \cite{karimireddybyzantine},
and inner-product manipulation attack \cite{xie2020fall}.
The detailed attack setups are deferred to~\prettyref{app: experiments}.
Our~\byzalg~is a user-level DP framework, compatible with any robust aggregator satisfying~\prettyref{def: robust agg},
and it attains the optimal rate when instantiated with {\tt CWTM}.
Other than {\tt CWTM}, the SOTA robust aggregators
include 
geometric median~\cite{chen2017distributed}, 
coordinate-wise median~\cite{yin2018byzantine},
and {\tt CClip}~\cite{karimireddy2021learning}.
We additionally compare against two baselines:
a simple mean as an aggregator in the attack-free setting ($f=0)$,
and {\tt CAFCOR} \cite{allouah2025towards}.
{\tt CAFCOR} is a local DP algorithm that guarantees user-level DP without a trusted server.

{\bf Discussions.}
\prettyref{fig:final results main text} displays the final performance under different privacy budgets.
Observations are mostly consistent between the two datasets.
Compared with the attack-free baseline, a performance gap emerges due to the presence of Byzantine clients.
When user-level DP protection is not imposed, {\tt CWTM} is among the top-performing robust aggregators.
The other strong performers are {\tt CClip} and {\tt CAFCOR}, where the former relies on a clipping threshold that requires careful tuning, 
and the latter incurs high computational overhead (see~\prettyref{rmk: sc optimal} for details). 
As a byproduct, we find that server-side momentum also improves the empirical performance of robust aggregators.
Intuitively, the shared anchor $\beta_t \tilde R^{t-1}$ pulls all clients' updates toward a common direction, thereby strengthening their concentration and improving the robust aggregator's effectiveness against Byzantine clients.
We defer the details to~\prettyref{tab:benefits of server momentum} in~\prettyref{app: experiments}.

Under user-level DP protections, {\tt CWTM} consistently outperforms the other baselines in the presence of Byzantine attacks.
We attribute this to its dimension-free sensitivity, which yields the optimal convergence rate.
Stronger privacy requires higher DP noise, degrading performance.
Each nonlinear robust aggregator induces a different Byzantine-robust user-level sensitivity (\prettyref{def: user-level sensitivity robust aggregator}) due to the differences in structure; see~\prettyref{tab: cl byzantine sensitivity} for details.
Since such sensitivity determines the variance of the DP noise injected at the server, it dictates the signal-to-noise ratio and directly shapes the privacy-utility tradeoff.
Among the robust aggregators within our~\byzalg~framework, only {\tt CClip} and {\tt CWTM} achieve meaningful test accuracy, and they are also the only two whose sensitivity denominators grow with the number of clients.
However, the sensitivity of {\tt CClip} is proportional to a heuristically chosen clipping threshold $\tau$, typically set as $\tau = 10/(1 - \alpha)$~\cite{karimireddybyzantine,karimireddy2021learning}, which can be significantly larger than the corresponding $C$ in {\tt CWTM}.
{\tt CAFCOR} adopts an interesting correlated-noise mechanism to inject DP noise at clients and lies outside our~\byzalg~framework.
Nevertheless, it still falls behind, which we attribute to its suboptimal convergence rate; see~\prettyref{rmk: sc optimal} for details.
We also show that server-side momentum is essential to guarantee performance.
The results are deferred to~\prettyref{fig:beta and utility} in~\prettyref{app: experiments}.

\section{Conclusion}
\label{sec: conclusion}
In this paper, we study Byzantine-resilient federated learning with user-level DP. 
We have established an information-theoretic lower bound for any Byzantine-resilient empirical risk minimization problem under user-level DP and heterogeneous clients.
We have proposed~\byzalg, a central DP algorithm featuring two-level momentum across client and server.
We have shown that~\byzalg~achieves the order-optimal convergence bound for strongly-convex objectives.
Numerical experiments have corroborated our analysis.

\newpage
\begin{ack}
We gratefully acknowledge the support
from the National Science Foundation under grants 2106891, 2107062, and the National Science Foundation CAREER award under grant 2340482.
The views and conclusions contained in this document are those of the authors and should not be interpreted as representing the official policies, either expressed or implied,
of 
the National Science Foundation, 
or the U.S. Government. The U.S. Government is authorized to reproduce and distribute reprints for Government purposes notwithstanding any copyright notation herein.
\end{ack}

\bibliographystyle{plain}
\bibliography{bib,byzantine_bib}

\clearpage
\newpage
\appendix 

\section*{Appendices}
\label{toc}

Here, we provide an overview of the Appendix.
In particular, the proofs of the main results are presented and backed by supporting intermediate results.

{
\setlength{\parskip}{3pt}
\setlength{\itemsep}{0pt}
\startcontents[sections]
\printcontents[sections]{l}{1}{\setcounter{tocdepth}{3}}
}

\clearpage
\newpage

\setcounter{lemma}{0} %
\renewcommand{\thelemma}{\Alph{section}.\arabic{lemma}}
\renewcommand{\thetheorem}{\Alph{section}.\arabic{theorem}}

\section{Limitations}
\label{app: limitations}

The limitations of our work are twofold:
\begin{enumerate}[leftmargin=*]
    \item 
    We assume a trustworthy parameter server.
    While this is the standard setting in regulated domains and achieves the minimax-optimal convergence rate, it only applies to central DP.
    Recent work by Allouah et al. \cite{allouah2025towards} makes important progress in user-level DP without a trusted server in the presence of Byzantine adversaries.
    It is interesting to explore the minimax-optimal convergence rate when the server cannot be trusted.

    \item
    We assume that stochastic gradients are uniformly bounded for technical convenience. 
    Although this assumption is widely adopted in the DP literature~\cite{allouah2023privacy,bassily2014private,bassily2023user,ghazi2021user,levy2021learning,wang2022differentially}, real-world data distributions may exhibit heavy tails.
    Relaxing this assumption but applying gradient clipping can introduce bias that complicates the convergence analysis.
    It is interesting to explore the trade-off between the clipping threshold, privacy, and utility.

\end{enumerate}

\section{Broader Impacts}
\label{app: broad impact}
Federated learning has reshaped distributed ML/AI in recent years and has driven commercial applications such as autonomous vehicles, the Internet of Things, and natural language processing.
Our paper focuses on the fault-tolerance aspect of real-world federated learning systems in the presence of both malicious clients and privacy leakage.
Our work has substantially advanced the theory and algorithms for federated learning by bringing together insights from statistics, optimization, distributed computing 
and engineering practices.
In addition, our research is important for federated learning systems to expand their outreach to more adversarial deployment environments.
We are unaware of any potential negative social impacts of our work.

\section{Nomenclatures}
\label{app: nomenclatures}
In this section, we provide the notations and nomenclatures used throughout
our proofs for a comprehensive presentation.
It is worth noting that all notation has been properly introduced before its first use.
\begin{table}[!ht]
    \centering
    \caption{Notation table}
    \label{tab: notation table}
    \begin{tabular}{c  p{12cm}}
    \toprule
       $\norm{\bm{v}}$  
       &  
       The $l_2$ norm of a given vector $\bm{v}$.\\
       \midrule
       $\calF^t$ &
       The sigma algebra generated by randomness up to round $t$.\\
       \midrule
       $\reals^d$ &
       A $d$-dimensional vector space, where $d$ denotes the dimension. \\
       \midrule
       $[n]$ &
       A set $\sth{k \mid k\in\naturals, k\in[1,n]}$. \\
       \midrule
       $\calG$ &
       A set of honest clients. \\
       \midrule
       $\calB$ &
       A set of clients suffering from Byzantine attacks $\calB \triangleq [n] \setminus \calG$. \\
       \midrule
       $\Indc$ &
       An all one vector $\Indc \in \reals^d$. \\
       \midrule
       $\lesssim$ &
       $f(n) \lesssim g(n)$,
       if there exists a constant $c_{\calO}>0$ and an integer $n_0 \in \naturals$, $f(n)\le c_{\calO} g(n)$ for all $n \ge n_0$. \\
       \midrule
       $\gtrsim$ &
       $f(n) \gtrsim g(n)$,
       if there exists a constant $c_{\Omega}>0$ and an integer $n_0 \in \naturals$, $f(n)\ge c_{\Omega} g(n)$ for all $n \ge n_0$. \\
       \midrule 
       $\calO$ &
       $f(n) = \calO(g(n))$,
       if there exists a constant $c_{\calO}>0$ and an integer $n_0 \in \naturals$, $f(n)\le c_{\calO} g(n)$ for all $n \ge n_0$. \\
       \midrule 
       $\Theta$ &
       $f(n) = \Theta(g(n))$,
       if there exists a constant $c_{\Theta}>0$ and an integer $n_0 \in \naturals$, $f(n)= c_{\Theta} g(n)$ for all $n \ge n_0$. \\
        \midrule 
       $\Omega$ &
       $f(n) = \Omega(g(n))$,
       if there exists a constant $c_{\Omega}>0$ and an integer $n_0 \in \naturals$, $f(n)\ge c_{\Omega} g(n)$ for all $n \ge n_0$. \\
    \bottomrule
    \end{tabular}
\end{table}

\begin{table}[!ht]
    \centering
    \caption{Algorithmic nomenclature table}
    \label{tab: alg nomenclature table}
    \begin{tabular}{p{1cm}  p{12cm}}
    \toprule
        $\bfthe^t$ &
       The global model at the {\bf beginning} of round $t$ in~\prettyref{alg: byzantine dp learning base stochastic v2}. \\ \midrule
       $F_i(\bfthe)$ &
       The local objective function at client $i$. \\
       \midrule
       $\byzobj (\bfthe)$ &
        The global objective function defined in~\eqref{eq: Byzantine global obj}:
       $\byzobj (\bfthe) \triangleq \sum_{i \in \calG} F_i(\bfthe)/\abth{\calG}$. \\
       \midrule
       $\nabla \ell_i(\bfthe)$ &
       The local stochastic gradient function at client $i$ taken with respect to $\bfthe$. \\
       \midrule
       $\hat \g_i^{t}$ &
       The clipped local stochastic gradient function at client $i$ taken with respect to $\bfthe$.
       $\hat{\g}_i^{t}
        \gets
        {\tt clip} (\nabla \ell_i (\bfthe^t; \xi_i^t), C)
        ,
        ~\text{where}~
        {\tt clip}(\bm{v}, C) = 
        \bm{v} \cdot \min \big\{
        1, C/\norm{\bm{v}}
        \big\}.$
       \\
       \midrule
       $\m_i^{t}$ &
       The client-side momentum at client $i$ in round $t$. \\
       \midrule
       $\overline{\m}^{t}$ &
       An average over the client-side momentum for clients in $\calG$ in round $t$.
       \\
       \midrule
       $\hat{\m}_i^{t}$ &
       The local update from client $i$ to the PS in round $t$. 
       It duplicates $\m_i^{t}$ for honest clients in $\calG$ but can be arbitrary for Byzantine clients.
       \\
       \midrule
       $\M_i^{t}$ &
       The server-side momentum in round $t$ for client $i$. \\
       \midrule
       $\overline{\M}^{t}$ &
       An average over the server-side momentum for clients in $\calG$ in round $t$, \ie, $\overline{\M}^{t} \triangleq \sum_{i \in \calG} \M_{i}^{t} / \abth{\calG}$.
       \\
       \midrule
       $\nabla F_i(\bfthe)$ &
       The local full-batch gradient function at client $i$ taken with respect to $\bfthe$. \\
       \midrule
       $\nabla \byzobj(\bfthe)$ &
       The global full-batch gradient function taken with respect to $\bfthe$. \\
       \midrule
       $R^{t}$ &
       The robust aggregation at the PS in round $t$ by a robust aggregator $R$. \\
       \midrule
       $\X_{dp}^{t}$ &
       The DP noise at the PS in round $t$. \\
       \midrule
       $\tilde R^{t}$ &
       The noisy robust aggregation at the PS in round $t$ is a sum of a robust aggregation $R$ and a DP noise $\X_{dp}^t$. \\
       \midrule
       $D_i$ &
       Client $i$'s local empirical data distribution. \\
       \midrule
       $\xi_i$ &
       An {\bf independent} stochastic sample drawn from client $i$'s local distribution $D_i$. \\ 
    \bottomrule
    \end{tabular}
\end{table}

\begin{table}[!t]
    \centering
    \caption{Variable table}
    \label{tab: constant table}
    \begin{tabular}{c  p{12cm}}
    \toprule
       $L$ &
       Lipschitz constant in~\prettyref{ass: smmothness}. \\
       \midrule
       $\sigma^2$ &
       The upper bound of the stochastic gradient variance. \\
       \midrule
       $\sigma_{dp,t}^2$ &
       The variance of DP noise in round $t$. \\
       \midrule
       $(B,~G)$ &
       Parameters that capture the averaged local-global gradient dissimilarity.
       \\
       \midrule
       $C$ &
       A uniform upper bound on the stochastic gradient norm. \\
       \midrule
       $n$ &
       The number of clients in the FL system.  \\
       \midrule
       $m$ &
       The number of local data points at each client in the FL system.  \\
    \bottomrule
    \end{tabular}
\end{table}

\section{Useful Inequalities}
\label{app: preliminaries}
For completeness and for ease of exposition, we present some common inequalities that will be frequently used in our proofs.
The followings hold for any $\bm{a}_i \in \reals^d$ and any $i\in[n]$.
\begin{enumerate}[leftmargin=*]
\item {Jensen's inequality.}
\begin{align}
\label{eq: jensen's inequality}
\norm{\frac{1}{n} \sum_{i=1}^n \bm{a}_i}^2
\le
\frac{1}{n} \sum_{i=1}^n \norm{\bm{a}_i}^2
~~~\text{and}~~~
\norm{\sum_{i=1}^n \bm{a}_i}^2
\le
n \sum_{i=1}^n \norm{\bm{a}_i}^2.
\end{align}
\item {Young's inequality (a.k.a. Peter-Paul inequality).}
\begin{align}
\label{eq: Young's inequality}
\iprod{\bm{a}_1}{\bm{a}_2}
\le
\frac{\norm{\bm{a}_1}^2}{2 \epsilon} + \frac{\epsilon \norm{\bm{a}_2}^2}{2}, ~~~ \text{for any }\epsilon>0. 
\end{align}
Equivalently,
we have
\begin{align}
\nonumber
\norm{\bm{a}_1 + \bm{a}_2}^2
&=
\norm{\bm{a}_1}^2
+
\norm{\bm{a}_2}^2
+
2 \iprod{\bm{a}_1}{\bm{a}_2} \\
\label{eq: Young's inequality 2}
&\le
\pth{1 + \frac{1}{\epsilon}}
\norm{\bm{a}_1}^2
+
\pth{1 + \epsilon}
\norm{\bm{a}_2}^2
, ~~~ \text{for any }\epsilon>0. 
\end{align}

\item {Cauchy-Schwartz inequality.}
\begin{align}
\label{eq: CS inequality}
\iprod{\bm{a}_1}{\bm{a}_2}
\le
\norm{\bm{a}_1}
\norm{\bm{a}_2}
. 
\end{align}

\item {Smoothness corollary.} 
Given~\prettyref{ass: smmothness}, it holds that
\begin{align}
\label{eq: smoothness corollary}
\nonumber
F(\bm{a}_1) - F(\bm{a}_2) 
&=
\iprod{\bm{a}_1 - \bm{a}_2}{\int_0^1 \nabla F(\bm{a}_2 + \tau (\bm{a}_1 - \bm{a}_2)) \diff \tau}
\\
\nonumber
&=
\iprod{\nabla F(\bm{a}_2)}{\bm{a}_1 - \bm{a}_2} +
\int_0^1 \iprod{\bm{a}_1 - \bm{a}_2}{\nabla F(\bm{a}_2 + \tau (\bm{a}_1 - \bm{a}_2)) - \nabla F(\bm{a}_2) }\diff \tau \\
\nonumber
&\overset{(a)}{\le}
\iprod{\nabla F(\bm{a}_2)}{\bm{a}_1 - \bm{a}_2} +
L \int_0^1 \tau \norm{\bm{a}_1 - \bm{a}_2}\norm{(\bm{a}_1 - \bm{a}_2) }\diff \tau
\\
&\le
\iprod{\nabla F(\bm{a}_2)}{\bm{a}_1 - \bm{a}_2}
+
\frac{L}{2} \norm{\bm{a}_1 - \bm{a}_2}^2
,
\end{align}
where $(a)$ follows from Cauchy-Schwartz inequality and~\prettyref{ass: smmothness}.
\end{enumerate}

\newpage
\section{Lower Bound}
\label{app: byzantine unlearning user-level lower bound}

\subsection{Proof of~\prettyref{lmm: dp lower bound non-adversarial cl} (Non-Byzantine)}
\label{app: non-adversarial lower bound}

\begin{lemma}
\label{lmm: dp lower bound non-adversarial cl app}
    Suppose that 
    $\epsilon, \delta \in (0,1)$ 
    and 
    $ 1/2^{n^{1 - \gamma}} \le 
    \delta 
    \le 1/n^{1 + \gamma} 
    $,
    where $\gamma \in (0, 1)$.
    For any given $L, B \ge 0$ and $G >0$, there exists a collection of $n$ functions $\{F_i\}_{i \in [n]}$,
    each defined over a dataset of $m$ samples, that are all
    i) $L$-smooth,  
    ii) $\mu$-strongly convex with $\mu = \Theta(L)$, 
    and iii) satisfy~\prettyref{ass: bounded similarity} with respect to $F$.
    If $\calA$ is $(\epsilon, \delta)$-DP, then 
    the output $\hat{\bfthe}$ of a $(\epsilon, \delta)$ user-level DP learning algorithm $\calA$ has an error at least
        \begin{small}
        \begin{align}
        \label{eq: private dp lower bound sc}
        \Expect \Big[\|\hat{\bfthe} - \bfthe^{\ast} \|_2^2\Big]
        &=
        \Omega
        \Big(
        \frac{d
         G^2 \log (1 / \delta)
        }{\mu^2 \epsilon^2 n^2}
        \Big)
        ,~
        \Expect \Big[\|\nabla F (\hat{\bfthe})\|_2^2\Big]
        =
        \Omega
        \Big(
        \frac{d
        G^2 \log (1 / \delta)
        }{\epsilon^2 n^2}
        \Big)
        .
        \end{align}
        \end{small}
\end{lemma}

\subsubsection*{Proof of~\prettyref{lmm: dp lower bound non-adversarial cl}}
    In this proof, we show the lower bound when we do not have any Byzantine adversaries, \ie, $f = 0$. 
    It is worth noting that our bound remains valid when we have $f > 0$. 
    We focus on the case of $G > 0$ only.
    We first construct $n$ local datasets, each consisting of $m$ elements: 
    Each client $i$ holds a local dataset $\calD_i$ of $m$ ordered elements 
    \begin{align*}
            \calD_i &= 
            \sth{\frac{\x_i^1 G_0}{
            L\sqrt{d}}, \ldots,  \frac{\x_i^m G_0}{L 
            \sqrt{d}}}
            ,
    \end{align*}
    where $\x_i^j \in \{\pm 1\}^d$ for 
    $j=1, \cdots, m$ and 
    $i \in [n]$, and $G_0 = \frac{G}{2}$, 
    such that 
    \[
    \sum_{i=1}^n \x_i^1 = \cdots = \sum_{i=1}^n \x_i^j = \cdots = \sum_{i=1}^n \x_i^m. 
    \]
    We define local functions based on these datasets as follows:   
    for any given $\mu\le L$,
    \begin{align*}
        F_i (\bfthe)
        &=
        \frac{\mu}{2}
        \norm{\bfthe - \frac{1}{m}\frac{G_0}{L\sqrt{d}}\sum_{j=1}^m \x_i^j}^2
        . 
    \end{align*}
The resulting global objective is 
    \begin{align*}
        F (\bfthe)
        &=
        \frac{1}{n}
        \sum_{i=1}^{n}
        F_i (\bfthe)
        .
    \end{align*}
    The optimality gaps $\|\nabla F(\hat{\bfthe})\|_2^2$ and $\norm{\bfthe - \bfthe^\ast}^2$ can be characterized as follows:   
        \begin{align*}
        \norm{\nabla F(\hat{\bfthe})}^2
        &
        =
        \mu^2
        \norm{
        \bfthe - 
        \frac{G_0}{L\sqrt{d}} \frac{1}{mn}
        \sum_{i=1}^n\sum_{j=1}^m \x_i^j
        }^2
        , 
        ~
        \text{and} ~
        \norm{\bfthe - \bfthe^\ast}^2
        =
       \norm{
        \bfthe - 
        \frac{G_0}{L\sqrt{d}} \frac{1}{mn}
        \sum_{i=1}^{n}\sum_{j=1}^{m} \x_i^j
        }^2. 
        \end{align*}

    Next, we verify that the constructed functions collectively satisfy the $(B, G)$-conditions. 
    It can be readily seen that 
    \begin{align*}
        \nabla F_i (\bfthe)
        =
        \mu \bfthe - 
        \frac{\mu G_0 }{L\sqrt{d}} \frac{1}{m}
        \sum_{j=1}^{m} \x_i^j, 
        ~ \text{and} ~~ 
        \nabla F (\bfthe)
        =
        \mu \bfthe - 
        \frac{\mu G_0 }{L\sqrt{d}} \frac{1}{mn}
        \sum_{i=1}^{n}\sum_{j=1}^{m} \x_i^j
        .
    \end{align*}
    It follows that
    \begin{align*}
        \frac{1}{n}
        \sum_{i=1}^{n}
        \norm{
        \nabla F_i (\bfthe)
        -
        \nabla F (\bfthe)
        }^2
        &= 
        \frac{\mu^2G_0^2}{L^2 d}
        \sum_{i=1}^{n}
        \norm{
        \frac{1}{m}\sum_{j=1}^{m}\x_i^j 
        -
        \frac{1}{nm}\sum_{i=1}^{n}\sum_{j=1}^{m}\x_i^j
        }^2
        \le
        4 G_0^2 
        = G^2
        .
    \end{align*}
    Thus, our construction satisfies the $(B,G)$-condition for arbitrary given $B\ge 0$ and $G>0$.
    
    Different from the item-level differential privacy guarantees in Definition 2.3 in~\cite{allouah2023privacy}, 
    \prettyref{def: dp federated learning} requires the indistinguishability of clients' local datasets. 
    At a high level, Definition \ref{def: dp federated learning} does not require each client to be differentially private with respect to its own datasets.

Recall that $\sum_{i=1}^{n} \x_i^1 = \cdots = \sum_{i=1}^{n} \x_i^j = \cdots = \sum_{i=1}^{n} \x_i^{m}$, which results in 
\begin{align*}
\frac{1}{mn} \sum_{i=1}^{n} \sum_{j=1}^{m} \x_i^j = \frac{1}{m} \sum_{j=1}^{m} \frac{1}{n} \sum_{i=1}^{n} \x_i^j =  \frac{1}{n} \sum_{i=1}^{n} \x_i^j, ~~~ \forall ~ j\in [m]. 
\end{align*}
Thus, when $\mu = \Theta(L)$, we have 
        \begin{align*}
        \expect{\norm{\nabla F(\hat{\bfthe})}^2}
        &=
        \mu^2 
        \norm{
        \hat{\bfthe}
        -
        \frac{1}{nm}
        \sum_{i=1}^{n}
        \sum_{j=1}^{m}
        \frac{\x_i^j G_0}{L \sqrt{d}}
        }^2 \\
        &= 
        \mu^2 \frac{1}{m}
        \norm{\Indc_{m} \otimes \hat{\bfthe} - 
        \frac{G_0}{L \sqrt{d}} 
        \begin{bmatrix}
         \frac{1}{n} \sum_{i=1}^{n}\x_i^1 \\
         \vdots\\ 
         \frac{1}{n}\sum_{i=1}^{n}\x_i^{m} 
        \end{bmatrix}}^2  \\
        &\overset{(a)}{=}
        \Omega
        \pth{
        \mu^2 \frac{1}{m} 
        \frac{G_0^2}{L^2}
        \cdot
        \frac{m d \log (1 / \delta)}{\epsilon^2 n^2}} 
        \overset{(b)}{=}
        \Omega
        \pth{\frac{d \log (1 / \delta)}{\epsilon^2 n^2} G^2}
        ,
        \end{align*}    
        where equality $(a)$ follows by invoking \prettyref{cor: dp sample complexity lower bound}, and equality $(b)$ follows from $\mu = \Theta(L)$ and $G^2 = 4 G_0^2$.
  Similarly, 
        \begin{align*}
        \expect{\norm{\bfthe - \bfthe^\ast}^2}
        &=
        \norm{
        \hat{\bfthe}
        -
        \frac{1}{nm}
        \sum_{i=1}^{n}
        \sum_{j=1}^{m}
        \frac{\x_i^j G_0}{L \sqrt{d}}
        }^2
        =
        \Omega
        \pth{\frac{d \log (1 / \delta)}{\epsilon^2 n^2} \frac{G_0^2}{L^2}} \\
        &= 
        \Omega
        \pth{\frac{d \log (1 / \delta)}{\epsilon^2 n^2} \frac{G^2}{\mu^2}}
        . 
    \end{align*}

\subsection{Proof of~\prettyref{lmm: non private lower bound cl} (Non-private)}
\label{sec: non-private lower bound cl}
\begin{lemma}
\label{lmm: non private lower bound cl app} 
For any given $L, B \ge 0$ and $G > 0$, there exists a collection of $n$ functions $\{F_i\}_{i \in [n]}$,
each defined over a dataset of $m$ samples, that are all
i) $L$-smooth,  
ii) $\mu$-strongly convex with $\mu = \Theta(L)$, 
and iii) satisfy~\prettyref{ass: bounded similarity} with respect to $\byzobj$.
For this collection, the output $\hat{\bfthe}$ of any distributed algorithm $\calA$ satisfies 
\begin{small}
\begin{align}
    \Expect \Big[\|\hat{\bfthe} - \bfthe^{\ast} \|_2^2\Big]
    & = \Omega 
    \Big(
    \frac{f}{n - f}\frac{G^2}{\mu^2}\Big),  
    ~\Expect \Big[\|\nabla \byzobj (\hat{\bfthe})\|_2^2\Big]
     = \Omega\Big(\frac{f}{n - f}G^2\Big).
\end{align}
\end{small}
    
\end{lemma}
        
\subsubsection*{Proof of~\prettyref{lmm: non private lower bound cl} \cite{karimireddy2021learning,allouah2023fixing}}
We construct local function families such that, regardless of the algorithm employed, the resulting observations are indistinguishable under different Byzantine client configurations.

The construction is similar to Theorem III in~\cite{karimireddybyzantine} and Proposition 3.2 in~\cite{allouah2023privacy}, but with adaptations to different optimality gaps.
Let $\tau = f / (n - f)$. 
Consider the following three groups of local datasets: 
    \begin{align*}
        \calD_{+} &= 
        \sth{\underbrace{+ \frac{G_0 \cdot \Indc}{L \sqrt{\tau d}}, \ldots, + \frac{G_0 \cdot \Indc}{L \sqrt{\tau d}}}_{n~\text{elements}}},~ 
        \calD_{-} = 
        \sth{\underbrace{- \frac{G_0 \cdot \Indc}{L \sqrt{\tau d}}, \ldots, - \frac{G_0 \cdot \Indc}{L \sqrt{\tau d}}}_{n~\text{elements}}},~
        \text{and}~
        \calD_{0} = 
        \sth{\underbrace{0 \cdot \Indc, \ldots, 0 \cdot \Indc}_{n~\text{elements}}}, 
    \end{align*}
    where $\Indc \in \reals^d$ and $G_0 > 0$ is a constant that can be chosen according to the given $G$.
    For each of the $n$ clients, its local dataset $\calD_i$ is constructed as: 
    \begin{align*}
        \calD_i &= 
        \begin{cases}
            \calD_{+},
            ~&\text{if $i \in [1, f]$};\\
            \calD_{-},
            ~&\text{if $i \in [f+1, 2f]$};\\
            \calD_{0},
            ~&\text{if $i \in [2 f + 1, n]$}.
        \end{cases}
    \end{align*}
    That is, each dataset is of $m$ rows chosen from its corresponding $d$-dimensional data universe. 
    
    Let $\mu\le L$ and $\mu = \Theta(L)$. Define each client's local objective as 
    \begin{align*}
        F_i (\bfthe) = \frac{\mu}{2 n} \sum_{\x \in \calD_i} \norm{\bfthe - \x}^2.
    \end{align*}
    It can be readily computed that each $F_i$ is $L$ smooth, $\mu$-strongly convex, and 
    \begin{align*}
    F_i(\bfthe) = 
    \begin{cases}
        & \frac{\mu}{2} \norm{\bfthe - \frac{G_0 \cdot \Indc}{L \sqrt{\tau d}}}^2, ~~~ \text{if } i\in [1, f], \\[0.5em]
        &  \frac{\mu}{2}\norm{\bfthe + \frac{G_0 \cdot \Indc}{L \sqrt{\tau d}}}^2, ~~~ \text{if } i\in [f+1, 2f], \\[0.5em]
        &  \frac{\mu}{2}\norm{\bfthe}^2, ~~~ \text{otherwise.}
        \end{cases}
\end{align*}
We consider the following two Byzantine client configurations induced by the adversary. 
\begin{itemize}[leftmargin=*]
    \item {\bf Configuration \#1:}
    $\calG_1 = \{1, \ldots, n\}$, i.e., $\calB^1 = \emptyset$
    \item {\bf Configuration \#2:}
    $\calG_2 = \{1, \ldots, f\} \cup \{2f+1, \ldots, n\},$ i.e., $\calB^2 = [f+1, 2f]$. 
\end{itemize}
Clearly, $|\calB^1| = 0\le f$, and $|\calB^2| = |[f+1, 2f]| = f$. 
For ease of exposition, let $\byzobj^1$ and $\byzobj^2$ denote the global objective functions of the two configurations, respectively, i.e., 
\begin{align*}
\byzobj^1 (\bfthe) = \frac{1}{\abth{\calG_1}}\sum_{i\in \calG_1} F_i (\bfthe), ~~\text{and} ~  \byzobj^2 (\bfthe) = \frac{1}{\abth{\calG_2}}\sum_{i\in \calG_2} F_i (\bfthe). 
\end{align*}
Let $\bfthe^{*1} = \arg\min_{\bfthe} \byzobj^1(\bfthe)$, and $\bfthe^{*2} = \arg\min_{\bfthe} \byzobj^2(\bfthe)$. 

Now, we verify the $(B, G)$ condition of the constructed functions. 
\begin{itemize}[leftmargin=*]
    \item {\bf Configuration \#1.}
    It is easy to see that 
    \begin{align*}
        \nabla \byzobj^1 (\bfthe)
        = \frac{1}{n}\sum_{i=1}^f \mu\pth{\bfthe - \frac{G_0 \cdot \Indc}{L \sqrt{\tau d}}} + \frac{1}{n}\sum_{i=f+1}^{2f}\mu\pth{\bfthe + \frac{G_0 \cdot \Indc}{L \sqrt{\tau d}}} + \frac{1}{n}\sum_{i=2f+1}^{n} \mu\bfthe
        = \mu \bfthe
        .
    \end{align*}
    It follows that
    \begin{align*}
        &\frac{1}{n}
        \sum_{i =1}^n 
        \norm{\nabla F_i (\bfthe) - \nabla \byzobj^1 (\bfthe)}^2 \\
        &=
        \frac{f}{n}
        \norm{
        \mu \bfthe
        -
        \frac{\mu G_0 \Indc}{L\sqrt{\tau d} }
        -
        \mu \bfthe
        }^2
        +
        \frac{n - 2f}{n}
        \norm{
        \mu \bfthe
        -
        \mu \bfthe
        }^2
        +
        \frac{f}{n}
        \norm{
        \mu \bfthe
        +
        \frac{\mu G_0 \Indc}{L\sqrt{\tau d} }
        -
        \mu \bfthe
        }^2 \\
        & \le 
        \frac{2 f}{n}
        \frac{G_0^2}{\tau}
        \le
        \frac{2 f}{n - f}
        \frac{G_0^2}{\tau}
        = 2 G_0^2
        .
    \end{align*}
    \item {\bf Configuration \#2.} 
    Similarly, we have 
    \begin{align*}
        \nabla \byzobj^2 (\bfthe)
        &=
        \frac{f}{n - f}
        \pth{
        \mu \bfthe
        -
        \frac{\mu G_0 \Indc}{L\sqrt{\tau d} }
        }
        +
        \frac{n - 2f}{n - f} 
        \mu \bfthe
        =
        \mu \bfthe 
        -
        \frac{f}{n - f}
        \frac{\mu G_0 \Indc}{L\sqrt{\tau d}}
        .
    \end{align*}
    It follows that
    \begin{align*}
        &\frac{1}{n - f}
        \sum_{i \in \calG_2}
        \norm{\nabla F_i (\bfthe) - \nabla \byzobj^2 (\bfthe)}^2 \\
        &=
        \frac{f}{n - f}
        \norm{\frac{\mu}{L}\pth{ 
        \frac{f}{n - f}
        \frac{G_0 \Indc}{\sqrt{\tau d} }
        -
        \frac{G_0 \Indc}{\sqrt{\tau d}}}
        }^2
        +
        \frac{n - 2f}{n - f}
        \norm{ \frac{\mu}{L}\pth{ 
        \frac{f}{n - f}
        \frac{G_0 \Indc}{\sqrt{\tau d}}}
        }^2 \\
        &\le 
        \frac{G_0^2}{\tau}
        \frac{f (n - 2f) (n - f)}{(n - f)^3}
        =
        \frac{f}{n - f}
        \frac{G_0^2}{\tau}
        =
        G_0^2
        .
    \end{align*}
\end{itemize}
Therefore, for any given $G>0$, we can choose $G_0 = \frac{1}{\sqrt{2}} G$.  
For any algorithm $\calA (F_1, \cdots, F_m)$, it cannot distinguish Configuration \#1 from Configuration \#2, since the Byzantine clients in Configuration \#2 can disguise themselves by mimicking the behaviors of the non-Byzantine clients $f+1, \cdots, 2f$ in Configuration \#1. Hence, $\calA$ should generate the same output $\hat{\bfthe}$ under these two configurations. 

We know that 
\begin{align*}
\norm{\nabla \byzobj^1(\hat{\bfthe})}^2  
= \norm{\mu \hat{\bfthe}}^2, 
~\text{and} ~ 
\norm{\nabla \byzobj^2(\hat{\bfthe})}^2 =
    \norm{
    \mu \hat{\bfthe} - 
    \sqrt{
    \frac{f}{n -f}
    }
    \frac{\mu G_0 \Indc }{L\sqrt{d}}}^2.  
\end{align*}    
In addition, 
\begin{align*}
\norm{\hat{\bfthe} - \bfthe^{*1}}^2 = \norm{\hat{\bfthe}}^2, 
~~ \text{and} ~ 
\norm{\hat{\bfthe} - \bfthe^{*2}}^2 = \norm{\hat{\bfthe} - \sqrt{\frac{f}{n -f}}\frac{G_0 \Indc }{L\sqrt{d}}}^2. 
\end{align*}

So, we have
\begin{align*}
    \max \sth{
    \expect{\norm{\hat{\bfthe} - \bfthe^{\ast 1 }}^2},
    \expect{\norm{\hat{\bfthe} - \bfthe^{\ast 2}}^2}
    }
    &\ge
    \frac{
    \expect{\norm{\hat{\bfthe} - \bfthe^{\ast 1 }}^2}
    +
    \expect{\norm{\hat{\bfthe} - \bfthe^{\ast 2}}^2}
    }{2} \\
    &=
    \frac{\expect{
    \norm{\hat{\bfthe}}^2
    +
    \norm{\hat{\bfthe} 
    -
    \sqrt{\frac{f}{n - f}}
    \frac{G_0 \Indc}{L \sqrt{d}}
    }^2
    }
    }{2} \\
    &\ge
    \frac{f}{n - f}
    \frac{G_0^2}{4 L^2} \\
    &=
    \frac{f}{n - f}
    \frac{G^2}{8 L^2}
    .
\end{align*}
In addition, 
\begin{align*}
    \max \sth{
    \expect{\norm{\nabla \byzobj^1(\hat{\bfthe})}^2},
    \expect{\norm{\nabla \byzobj^2(\hat{\bfthe})}^2}
    }
    &\ge
    \frac{
    \expect{\norm{\nabla \byzobj^1(\hat{\bfthe})}^2}
    +
    \expect{\norm{\nabla \byzobj^2(\hat{\bfthe})}^2}
    }{2} \\
    &\ge
    \mu^2
    \frac{
    \expect{\norm{\hat{\bfthe} - \bfthe^{1 \ast}}^2}
    +
    \expect{\norm{\hat{\bfthe} - \bfthe^{2 \ast}}^2}
    }{2} \\
    &\ge
    \frac{f}{n - f}
    \frac{\mu^2 G^2}{L^28}. 
\end{align*} 
We can get the desired results by invoking $L = \Theta(\mu)$.

\subsection{Proof of~\prettyref{lmm: adversarial dp lower bound cl} (Coupled)}
\label{app: adversarial lower bound}

\begin{lemma}
    \label{lmm: adversarial dp lower bound cl app} 
    Let $m, n, d\ge 1$ and $ 1 \le f < m/2$.  
    Suppose that 
    $\epsilon, \delta \in (0,1)$ 
    and 
    $ 1/2^{(n - f)^{1 - \gamma}} \le 
    \delta 
    \le 1/(n - f)^{1 + \gamma} 
    $,
    where $\gamma \in (0, 1)$.
    For any given $L, B\ge 0$ and $G>0$, there exists a collection of $n$ functions $\{F_i\}_{i \in [n]}$,
    each defined over a dataset of $m$ samples, that are all
    i) $L$-smooth,  
    ii) $\mu$-strongly convex with $\mu = \Theta(L)$, 
    and iii) satisfy~\prettyref{ass: bounded similarity} with respect to $\byzobj$.
    The output $\hat{\bfthe}$ of a $(\epsilon, \delta)$ user-level DP learning algorithm $\calA$ has an error at least
    \begin{small}
        \begin{align}
        \label{eq: adversarial dp lower bound sc}
        \Expect \Big[\|\hat{\bfthe} - \bfthe^{\ast} \|_2^2\Big]
        &=
        \Omega \Big(
        \frac{f}{n}
        \frac{d G^2 \log (1 / \delta)}{\mu^2 \epsilon^2 (n - f)^2}
        \Big)
        ,~
        \Expect \Big[\|\nabla \byzobj (\hat{\bfthe})\|_2^2\Big]
        =
        \Omega \Big(
        \frac{f}{n}
        \frac{d G^2 \log (1 / \delta)}{\epsilon^2 (n - f)^2}
        \Big)
        .
        \end{align}
    \end{small}
\end{lemma}

\subsubsection*{Proof of~\prettyref{lmm: adversarial dp lower bound cl}}
    We construct local function families such that, regardless of the algorithm adopted, the resulting observations are indistinguishable under different Byzantine client configurations.
    Consider the following three groups of local datasets:
    \begin{align*}
        \calD_{+} &= 
        \sth{\underbrace{+ \frac{G_0 \Indc}{L \sqrt{\tau d}}, \ldots, + \frac{G_0 \Indc}{L \sqrt{\tau d}}}_{m~\text{elements}}},~
        \calD_{-} = 
        \sth{\underbrace{- \frac{G_0 \Indc}{L \sqrt{\tau d}}, \ldots, - \frac{G_0 \Indc}{L \sqrt{\tau d}}}_{m~\text{elements}}},
        \text{and}~
        \calD_{0} = 
        \sth{\underbrace{0 \cdot \Indc, \ldots, 0 \cdot \Indc}_{m~\text{elements}}},
    \end{align*}
    where $\Indc \in \reals^d$, $\tau = f / (n -f)$, and $G_0 > 0$ is a constant that can be chosen according to a given $G$.
    For each of the $n$ clients, its local dataset $\calD_i$ is constructed as:
    \begin{align*}
        \calD_i &= 
        \begin{cases}
            \calD_{+},
            ~&\text{if $i \in [1, f]$};\\
            \calD_{-},
            ~&\text{if $i \in [f+1, 2f]$};\\
            \calD_{0},
            ~&\text{if $i \in [2 f + 1, n]$}.
        \end{cases}
    \end{align*}
    That is, each dataset is of $m$ rows chosen from its corresponding $d$-dimensional data universe.
    
    Let $\mu \le L$ and $\mu = \Theta (L)$.
    Define each client's local objective as
    \begin{align*}
        F_i (\bfthe) = \frac{\mu}{2 n} \sum_{\x \in \calD_i} \norm{\bfthe - \x}^2.
    \end{align*}
    It can be readily computed that each $F_i$ is $L$-smooth, $\mu$-strongly convex, and
    \begin{align*}
        F_i (\bfthe)
        &=
        \begin{cases}
        \frac{\mu}{2}
        \norm{\bfthe - \frac{G_0\Indc}{L\sqrt{\tau d}}}^2
        ,~\text{if}~i \in [1,f]; \\[0.5em]
        \frac{\mu}{2}
        \norm{\bfthe + \frac{G_0\Indc}{L \sqrt{\tau d}}}^2
        ,~\text{if}~i \in [f+1,2f]; \\[0.5em]
        \frac{\mu}{2}
        \norm{\bfthe}^2
        ,~\text{otherwise}.
        \end{cases}
    \end{align*}
    We consider the following three Byzantine client configurations induced by the adversary.
   \begin{itemize}[leftmargin=*]
        \item {\bf Configuration \#1:}
        $\calG_1 = \{1, \ldots, f\} \cup \{2f+1, n\}$, \ie, $\calB^1 = \{f+1, \ldots, 2f \}$;
        \item {\bf Configuration \#2:}
        $\calG_2 = \{f+1, \ldots, 2f\} \cup \{2f+1, n\}$, \ie, $\calB^2 = \{1, \ldots, f\}$;
        \item {\bf Configuration \#3:} 
        $\calG_3 = [n]$, \ie, $\calB^3 = \emptyset$.
   \end{itemize}
   Clearly, 
   $|\calB^1| = |[f+1, 2f]| = f$, 
   $|\calB^2| = |[1,f]| = f$,
   and $|\calB^3| = 0 \le f$.
   
   For any algorithm $\calA (F_1, \cdots, F_{n})$, it cannot distinguish Configuration \#1 from Configuration \#3, since the Byzantine clients in Configuration \#1 can disguise themselves by mimicking the behaviors of the non-Byzantine clients $f+1, \cdots, 2f$ in Configuration \#3.
   Similarly, any algorithm $\calA (F_1, \cdots, F_{n})$ cannot distinguish Configuration \#2 from Configuration \#3, since the Byzantine clients in Configuration \#2 can disguise themselves by mimicking the behaviors of the non-Byzantine clients $1, \cdots, f$ in Configuration \#3.
   Therefore, $\calA$ should generate the same output $\hat{\bfthe}$ under Configuration \#1 and Configuration \#2.

    For ease of exposition, let $\byzobj^1$ and $\byzobj^2$ denote the global objective functions of the configurations \#1 and \#2, respectively, \ie,
    \[
        \byzobj^1 (\bfthe) = 
        \frac{1}{\abth{\calG_1}}
        \sum_{i \in \calG_1} F_i (\bfthe)~
        \text{and}~
        \byzobj^2 (\bfthe) = 
        \frac{1}{\abth{\calG_2}}
        \sum_{i \in \calG_2} F_i (\bfthe).
    \]
    Let $\bfthe^{\ast 1} = \arg \min_{\bfthe} \byzobj^1(\bfthe),$ and $\bfthe^{\ast 2} = \arg \min_{\bfthe} \byzobj^2(\bfthe)$.
    
    Now, we verify the $(B, G)$ condition of the constructed functions.
    Recall that
    $n - f > f \Leftrightarrow f < \frac{n}{2} 
    \Leftrightarrow 
    n - f > \frac{n}{2}$.
    \begin{itemize}[leftmargin=*]
        \item {\bf Configuration \#1.}
        It is easy to see that
        \begin{align*}
        \nabla \byzobj^1 (\bfthe)
        &= 
        \frac{f}{n - f}
        \sum_{i=1}^{f}
        \mu
        \pth{
        \bfthe - \frac{G_0 \Indc}{L \sqrt{\tau d}}
        }
        +
        \frac{n-2f}{n - f}
        \sum_{i=2f+1}^{n}
        \mu
        \bfthe
        =
        \mu \bfthe 
        - 
        \sqrt{\frac{f}{n-f}} \frac{\mu G_0 \Indc}{L \sqrt{d}}
        .
        \end{align*}
        It follows that
        \begin{align*}
            \frac{1}{n - f}
            \sum_{i \in \calG}
            \norm{\nabla F_i (\bfthe) - \nabla \byzobj^1(\bfthe)}^2
            &=
            \underbrace{
            \frac{(n - 2 f) \mu^2}{(n - f) L^2}
            \norm{
            \sqrt{\frac{f}{n-f}} \frac{G_0 \Indc}{\sqrt{d}}
            }^2
            }_{(1.a)} 
            +
            \underbrace{\frac{f \mu^2}{(n - f) L^2}
            \norm{
            \sqrt{\frac{f}{n-f}} \frac{G_0 \Indc}{\sqrt{d}}
            - \frac{G_0 \Indc}{\sqrt{\tau d}}
            }^2}_{(1.b)} 
            .
        \end{align*}
        \paragraph{Bounding $(1.a)$.}
        It holds that
        \begin{align*}
            \frac{(n - 2 f) \mu^2}{(n - f) L^2}
            \norm{
            \sqrt{\frac{f}{n-f}} \frac{G_0 \Indc}{\sqrt{d}}
            }^2
            &=
            \frac{(n - 2 f) \mu^2}{(n - f) L^2}
            \sum_{i=1}^d
            \norm{
            \sqrt{\frac{f}{n-f}} \frac{G_0}{\sqrt{d}}
            }^2 
            =
            \frac{\mu^2 G_0^2}{L^2 \tau}
            \frac{f }{n - f}
            \frac{(n - 2f) f}{(n - f)^2}
            \le
            G_0^2
            .
        \end{align*}
        \paragraph{Bounding $(1.b)$.}
        \begin{align*}
            \frac{f \mu^2}{(n - f) L^2}
            \norm{
            \sqrt{\frac{f}{n-f}} \frac{G_0 \Indc}{\sqrt{d}}
            - \frac{G_0 \Indc}{\sqrt{\tau d}}
            }^2
            &=
            \frac{f \mu^2}{(n - f) L^2}
            \sum_{i=1}^d
            \norm{
            \sqrt{\frac{f}{n-f}} \frac{G_0}{\sqrt{d}}
            - \frac{G_0}{\sqrt{\tau d}}
            }^2 
            \le
            \frac{\mu^2 G_0^2}{L^2 \tau}
            \frac{f}{m - f}
            \frac{(n - 2 f)^2}{(n - f)^2}
            \le
            G_0^2
            .
        \end{align*}
        Combining them together, it holds that
        \begin{align*}
            \frac{1}{n - f}
            \sum_{i \in \calG}
            \norm{\nabla F_i (\bfthe) - \nabla \byzobj^1(\bfthe)}^2
            &\le
            2 G_0^2
            .
        \end{align*}
        \item {\bf Configuration \#2:}
        \begin{align*}
        \nabla \byzobj^2 (\bfthe)
        &= 
        \frac{f}{n - f}
        \sum_{i=f+1}^{2f}
        \mu
        \pth{
        \bfthe - \frac{G_0 \Indc}{L \sqrt{\tau d}}
        }
        +
        \frac{n - 2f}{n - f}
        \sum_{i=2f+1}^{n}
        \mu
        \bfthe
        =
        \mu \bfthe
        +
        \sqrt{\frac{f}{n-f}}
        \frac{\mu G_0 \Indc}{L \sqrt{d}}
        .
        \end{align*}
        It follows that
        \begin{align*}
            \frac{1}{n - f}
            \sum_{i \in \calG}
            \norm{\nabla F_i (\bfthe) - \nabla \byzobj^2(\bfthe)}^2
            &=
            \underbrace{
            \frac{(n - 2 f) \mu^2}{(n - f) L^2}
            \norm{
            - 
            \sqrt{\frac{f}{n-f}} \frac{G_0 \Indc}{\sqrt{d}}
            }^2
            }_{(2.a)} 
            +
            \underbrace{\frac{f \mu^2}{(n - f) L^2}
            \norm{
            - \sqrt{\frac{f}{n-f}} \frac{G_0 \Indc}{\sqrt{d}}
            + \frac{G_0 \Indc}{\sqrt{\tau d}}
            }^2}_{(2.b)} 
            .
        \end{align*}
        \paragraph{Bounding $(2.a)$.}
        \begin{align*}
            \frac{(n - 2 f) \mu^2}{(n - f) L^2}
            \norm{
            - \sqrt{\frac{f}{n-f}} \frac{G_0 \Indc}{\sqrt{d}} 
            }^2
            &=
            \frac{(n - 2 f) \mu^2}{(n - f) L^2}
            \sum_{i=1}^d
            \norm{
            G_0
            \sqrt{\frac{f}{(n-f){d}}} 
            }^2 
            =
            \frac{\mu^2 G_0^2}{L^2 \tau}
            \frac{f}{n -f}
            \frac{(n - 2f) f}{(n - f)^2}
            \le
            G_0^2
            .
        \end{align*}
        \paragraph{Bounding $(2.b)$.}
        \begin{align*}
            \frac{f \mu^2}{(n - f) L^2}
            \norm{
            - \sqrt{\frac{f}{n-f}} \frac{G_0 \Indc}{\sqrt{d}}
            + \frac{G_0 \Indc}{\sqrt{\tau d}}
            }^2
            \le
            \frac{\mu^2 G_0^2}{L^2 \tau}
            \frac{f}{n - f}
            \frac{(n - 2 f)^2}{(n - f)^2}
            \le
            G_0^2
            .
        \end{align*}
        Combining them together, it holds that
        \begin{align*}
            \frac{1}{n - f}
            \sum_{i \in \calG}
            \norm{\nabla F_i (\bfthe) - \nabla \byzobj^2 (\bfthe)}^2
            &\le
            2 G_0^2
            .
        \end{align*}
    \end{itemize}
    Therefore, for any given $G > 0$, we can choose $G_0 = G / \sqrt{2}$.

    Let $\calD^\prime$ be an auxiliary dataset.
    \begin{align*}
        \calD^{\prime} &=
        \sth{
        \underbrace{\Indc_{m} \otimes
        \frac{G_0 \Indc_{d}}{L \sqrt{\tau n d}}, \ldots}_{f~\text{elements}},
        \underbrace{\Indc_{m} \otimes
        \frac{-G_0 \Indc_{d}}{L \sqrt{\tau n d}}, \ldots}_{f~\text{elements}},
        \underbrace{\Indc_{m} \otimes 0 \Indc_d}_{(n-2f)~\text{elements}}
        }
        .
    \end{align*} 
    Next, we show that the errors induced by $\calD$ and $\calD^{\prime}$ are the same in a special case.
    Suppose that an auxiliary global model parameter $\hat{\bfthe}^{\prime} \in \reals^{md}$
    It holds, when $\hat{\bfthe}^{\prime} = \Indc_{m} \otimes \hat{\bfthe} / \sqrt{m}$, that
    \begin{itemize}[leftmargin=*]
        \item {\bf Configuration \#1.}
        \begin{align*}
            \norm{\hat{\bfthe}^{\prime} - \bfthe^{\prime 1 \ast}}^2
            &=
            \norm{
            \hat{\bfthe}^{\prime}
            -
            \Indc_{m} \otimes
            \sqrt{\frac{f}{n -f}}
            \frac{G_0 \Indc}{\sqrt{m d}}^2
            } \\
            &=
            \norm{
            \Indc_{m} \otimes
            \frac{\hat{\bfthe}}{\sqrt{m}}
            -
            \Indc_{m} \otimes
            \sqrt{\frac{f}{n -f}}
            \frac{G_0 \Indc}{\sqrt{m d}}
            }^2 \\
            &=
            \frac{1}{m}
            \sum_{j=1}^{m}
            \norm{
            \hat{\bfthe}
            -
            \sqrt{\frac{f}{n -f}}
            \frac{G_0 \Indc}{\sqrt{d}}
            }^2
            =
            \norm{\hat{\bfthe} - \bfthe^{1 \ast}}^2
            .
        \end{align*}
        \item {\bf Configuration \#2.}
        \begin{align*}
            \norm{\hat{\bfthe}^{\prime} - \bfthe^{\prime 2 \ast}}^2
            &=
            \norm{
            \hat{\bfthe}^{\prime}
            +
            \Indc_{m} \otimes
            \sqrt{\frac{f}{n -f}}
            \frac{G_0 \Indc}{\sqrt{m d}}^2
            } \\
            &=
            \norm{
            \Indc_{m} \otimes
            \frac{\hat{\bfthe}}{\sqrt{m}}
            +
            \Indc_{m} \otimes
            \sqrt{\frac{f}{m -f}}
            \frac{G_0 \Indc}{\sqrt{m d}}
            }^2 \\
            &=
            \frac{1}{m}
            \sum_{j=1}^{m}
            \norm{
            \hat{\bfthe}
            +
            \sqrt{\frac{f}{n -f}}
            \frac{G_0 \Indc}{\sqrt{d}}
            }^2
            =
            \norm{\hat{\bfthe} - \bfthe^{2 \ast}}^2
            .
        \end{align*}
    \end{itemize}
    We construct another auxiliary dataset $\calD_{\pm}^{\prime}$ of $n - f$ data points.
    \begin{align*}
        \calD_{\pm}^{\prime} = 
        \sth{\underbrace{
        \Indc_{m} \otimes
        \sqrt{\frac{f}{n - f}}
        \frac{\x_1^{\pm} G_0 \Indc}{L \sqrt{n d}}, \ldots, 
        \Indc_n \otimes
        \sqrt{\frac{f}{n - f}}
        \frac{\x_{n-f}^{\pm} G_0 \Indc}{L \sqrt{n d}}}_{(n-f)~\text{elements}}},
    \end{align*}
    where $\Indc \in \reals^d$ and $\x_i^{\pm} \in \{-1, + 1\}$ for all $i \in [1, n - f]$.
    Let $\hat{\bfthe}_{(i)}$ denote the $i$-th coordinate of $\hat{\bfthe}$,
    $\overline{\x}^{\pm} \triangleq \sum_{i=1}^{n - f} \x_i^{\pm} / (n - f)$,
    and $\overline{\x}_{(i)}^{\pm}$ denote the $i$-th coordinate of $\overline{\x}^{\pm}$.
    Since we know that $\overline{\x}_{(i)}^{\pm} \in [-1, 1]$, it holds that $\overline{\x}_{(i)}^{\pm} = -1 \cdot \lambda_{(i)}  + 1 \cdot (1 - \lambda_{(i)})$ for some $\lambda_{(i)} \ge 0$.
    Further, it holds that $\lambda_i = \lambda$ because $\overline{\x}_{(i)}^{\pm}$ is by definition identical for all $i \in \{1, \ldots, d\}$.
    It follows that 
    \begin{align*}
    &\hat{\bfthe}_{(i)}^{\prime} - 
    \sqrt{\frac{f}{n-f}}
    \frac{\overline{\x}_{(i)}^{\pm} G_0}{L \sqrt{m d}} 
    = 
    \lambda
    \pth{\hat{\bfthe}_{(i)}^{\prime} + 
    \sqrt{\frac{f}{n-f}}\frac{G_0}{L \sqrt{m d}}
    }
    +
    (1 - \lambda) 
    \pth{\hat{\bfthe}_{(i)}^{\prime} - 
    \sqrt{\frac{f}{n-f}}\frac{G_0}{L \sqrt{m d}}
    } \\
    &\Rightarrow
    \norm{\hat{\bfthe}_{(i)}^{\prime} - 
    \sqrt{\frac{f}{n-f}}
    \frac{\overline{\x}_{(i)}^{\pm} G_0}{L \sqrt{m d}} 
    }^2
    \le
    \lambda
    {\norm{\hat{\bfthe}_{(i)}^{\prime} + 
    \sqrt{\frac{f}{n-f}}\frac{G_0}{L \sqrt{m d}}
    }^2}
    +
    (1 - \lambda) 
    {\norm{\hat{\bfthe}_{(i)}^{\prime} - 
    \sqrt{\frac{f}{n-f}}\frac{G_0}{L \sqrt{m d}}
    }^2} 
    .     
    \end{align*}
    Since this holds for each coordinate $i \in \{ 1,\ldots, d\}$, it holds that
    \begin{align*}
    &\sum_{i = 1}^{m d}
    \norm{\hat{\bfthe}_{(i)}^{\prime} - 
    \sqrt{\frac{f}{n-f}}
    \frac{\overline{\x}_{(i)}^{\pm} G_0}{L \sqrt{m d}} 
    }^2
    \le
    \lambda
    \sum_{i = 1}^{m d}
    {\norm{\hat{\bfthe}_{(i)}^{\prime} + 
    \sqrt{\frac{f}{n-f}}\frac{G_0}{L \sqrt{m d}}
    }^2}
    +
    (1 - \lambda)
    \sum_{i = 1}^{m d}
    {\norm{\hat{\bfthe}_{(i)}^{\prime} - 
    \sqrt{\frac{f}{n-f}}\frac{G_0}{L \sqrt{m d}}
    }^2} \\
    &\Leftrightarrow 
    ~~
    \norm{\hat{\bfthe}^{\prime} - 
    \Indc_{m} \otimes
    \sqrt{\frac{f}{n-f}}
    \frac{\overline{\x}^{\pm} G_0 \Indc}{L \sqrt{m d}}
    }^2
    \le
    \lambda
    \norm{\hat{\bfthe}^{\prime} + 
    \Indc_{m} \otimes
    \sqrt{\frac{f}{n-f}}\frac{G_0}{L \sqrt{m d}}
    }^2
    +
    (1 - \lambda)
    \norm{\hat{\bfthe}^{\prime} - 
    \Indc_{m} \otimes
    \sqrt{\frac{f}{n-f}}\frac{G_0}{L \sqrt{m d}}
    }^2
    .
    \end{align*}
    Taking expectation with respect to the randomness in the algorithm, we have
    \begin{align}
    \notag
    &\expect{\norm{\hat{\bfthe}^{\prime} - 
    \Indc_{m} \otimes
    \sqrt{\frac{f}{n-f}}
    \frac{\overline{\x}^{\pm} G_0 \Indc}{L \sqrt{m d}}}^2
    }^2 \\
    \notag
    &\le
    \lambda
    \expect{\norm{\hat{\bfthe}^{\prime} +
    \Indc_{m} \otimes
    \sqrt{\frac{f}{n-f}}\frac{G_0 \Indc}{L \sqrt{m d}}
    }^2}
    +
    (1 - \lambda)
    \expect{\norm{\hat{\bfthe}^{\prime} -
    \Indc_{m} \otimes
    \sqrt{\frac{f}{n-f}}\frac{G_0 \Indc}{L \sqrt{m d}}
    }^2} \\
    \label{eq: convex combination pm cl}
    &\le
    \max \sth{
    \expect{\norm{\hat{\bfthe}^{\prime} +
    \Indc_{m} \otimes
    \sqrt{\frac{f}{n-f}}\frac{G_0 \Indc}{L \sqrt{m d}}
    }^2},
    \expect{\norm{\hat{\bfthe}^{\prime} - 
    \Indc_{m} \otimes
    \sqrt{\frac{f}{n-f}}\frac{G_0 \Indc}{L \sqrt{m d}}
    }^2}
    }
    .
    \end{align}

    Recall that we have shown the indistinguishability of Configurations \#1 and \#2.
    Since the algorithm should generate the same output for all possible collections of Byzantine clients,
    we have for any $\hat{\bfthe}^{\prime}$ that
    \begin{align}
    \notag
        &\max \sth{
        \expect{\norm{\hat{\bfthe}^{\prime} - \bfthe^{\prime 1 \ast}}^2},
        \expect{\norm{\hat{\bfthe}^{\prime} - \bfthe^{\prime 2 \ast}}^2}
        } \\
    \notag
        &\qquad \qquad \qquad
        =
        \max \sth{
        \expect{\norm{\hat{\bfthe}^{\prime} -
        \Indc_{m} \otimes
        \sqrt{\frac{f}{n-f}}\frac{G_0}{L \sqrt{m d}}
        }^2}
        ,
        \expect{\norm{\hat{\bfthe}^{\prime} + 
        \Indc_{m} \otimes
        \sqrt{\frac{f}{n-f}}\frac{G_0}{L \sqrt{m d}}
        }^2}
        }
        \\
        &\qquad \qquad \qquad
        \overset{\eqref{eq: convex combination pm cl}}{\ge}
        \label{eq: coupled byzantine lower bound cl}
        \expect{
        \norm{
        \hat{\bfthe}^{\prime} 
        -
        \Indc_{m} \otimes
        \sqrt{\frac{f}{n-f}}
        \frac{\overline{\x}^{\pm} G_0 \Indc}{L \sqrt{m d}}
        }^2
        }
        .
    \end{align}
    In our data universe below, $\Delta = \sqrt{\frac{f}{m (n - f)}} \frac{G_0}{L}$ with dimension $m d$,
    and the one-way marginal (average) is computed over $n-f$ data points in the auxiliary dataset $\calD_{\pm}^\prime$. 
    It holds that
    \begin{align*}
        \expect{
        \norm{
        \hat{\bfthe}^{\prime} 
        -
        \Indc_{m} \otimes
        \sqrt{\frac{f}{n-f}}
        \frac{\overline{\x}^{\pm} G_0 \Indc}{L \sqrt{m d}}
        }^2
        }
        &\overset{(a)}{=}
        \Omega \pth{
        \frac{f}{n -f}
        \frac{n d \log (1 / \delta)}{m \epsilon^2 (n -f)^2}
        \frac{G_0^2}{L^2}
        }
        \overset{(b)}{=}
        \Omega \pth{
        \frac{f}{n -f}
        \frac{d \log (1 / \delta)}{\epsilon^2 (n -f)^2}
        \frac{G^2}{L^2}
        }
        ,
    \end{align*}
    where equality $(a)$ holds because of~\prettyref{cor: dp sample complexity lower bound}, and equality $(b)$ holds because $G^2 = 2 G_0^2$.
    It follows that
    \begin{align*}
        \max \sth{
        \expect{\norm{\hat{\bfthe} - \bfthe^{1 \ast}}^2},
        \expect{\norm{\hat{\bfthe} - \bfthe^{2 \ast}}^2}
        } 
        &=
        \max \sth{
        \expect{\norm{\hat{\bfthe}^{\prime} - \bfthe^{\prime 1 \ast}}^2},
        \expect{\norm{\hat{\bfthe}^{\prime} - \bfthe^{\prime 2 \ast}}^2}
        } \\
        &=
        \Omega \pth{
        \frac{f}{n -f}
        \frac{d \log (1 / \delta)}{\epsilon^2 (n -f)^2}
        \frac{G^2}{L^2}
        }
        .
    \end{align*}
    In addition,
    \begin{align*}
    \max \sth{
    \expect{\norm{\nabla \byzobj^1(\hat{\bfthe})}^2},
    \expect{\norm{\nabla \byzobj^2(\hat{\bfthe})}^2}
    }
    &=
    \mu^2
    \max \sth{
    \expect{\norm{\hat{\bfthe} - \bfthe^{1 \ast}}^2},
    \expect{\norm{\hat{\bfthe} - \bfthe^{2 \ast}}^2}
    } \\
    &\ge
    \Omega \pth{
    \frac{f}{n -f}
    \frac{d \log (1 / \delta) G^2}{\epsilon^2 (n -f)^2}
    }
    .
    \end{align*}
    We can get the desired results by invoking $L = \Theta (\mu)$.

\subsection{Proof of~\prettyref{thm: dp final lower bound cl}}
\label{app: lower bound full}
\prettyref{thm: dp final lower bound cl} can be obtained by summing up Lemmas~\ref{lmm: dp lower bound non-adversarial cl},~\ref{lmm: non private lower bound cl}, and~\ref{lmm: adversarial dp lower bound cl}.

\subsection{Supporting Lemmas of~\prettyref{app: byzantine unlearning user-level lower bound}}
\label{app: lower bound}

\begin{lemma}[\cite{steinke2016between}]
\label{lmm: dp lower bound ref}
Let $n, d \ge 1$, 
$\calX = \sth{\pm 1 / \sqrt{d}}^d$, and $\calY = [\pm 1 / \sqrt{d}]^d$. 
Let $\calD \in \calX^n$ denote a dataset of size $n$, and $\bar{\calD} \triangleq \frac{1}{n}\sum_{\x \in \calD} \x$ denote the average of the samples in $\calD$.
Let $\calA: \calX^n \to \calY$ be an arbitrary mapping used to estimate $\bar{\calD}$. 
Suppose that $\epsilon = O(1)$ and $2^{- o(n)} \le \delta \le 1 / n^{1 + \Omega (1)}$.
If $\calA$ is $(\epsilon, \delta)$-DP and $\expect{\lnorm{\calA (\calD) - \bar{\calD}}{1}} \le \sqrt{d} \varrho$, 
then 
\begin{align*}
    n = \Omega \pth{\frac{\sqrt{d \log (1 / \delta)}}{\varrho \epsilon}}
    .
\end{align*}
\end{lemma}
\begin{lemma}
\label{lmm: one-way marginal}
    Let $n, d \ge 1$, %
$\calX = \sth{\pm 1 / \sqrt{d}}^d$, and $\calY = [\pm 1 / \sqrt{d}]^d$. 
Let $\calD \in \calX^n$ denote a dataset of size $n$, and $\overline{\calD} \triangleq \frac{1}{n}\sum_{\x \in \calD} \x $ denote the average of the samples in $\calD$.
Let $\calA: \calX^n \to \calY$ be an arbitrary mapping used to estimate $\overline{\calD}$. 
Suppose that $\epsilon = O(1)$ and $2^{- o(n)} \le \delta \le 1 / n^{1 + \Omega (1)}$.
If $\calA$ is $(\epsilon, \delta)$-DP, then 
    \begin{align}
    \label{eq: lower bound non-adversarial}
    \expect{\norm{\calA(\calD) - \overline{\calD}}^2 }
    &= \Omega \pth{\frac{d \log(1/\delta)}{\epsilon^2 n^2}} 
    ,
    \end{align}  
    where the expectation is taken over the randomness of $\calA$. 
\end{lemma}  
\begin{proof}[\bf Proof of~\prettyref{lmm: one-way marginal}]
    We prove this lemma by contradiction.
    Suppose that~\eqref{eq: lower bound non-adversarial} is not true. Then  
    \[
    \expect{\norm{\calA(\calD) - \overline{\calD}}^2 }
    = o \pth{\frac{d \log(1/\delta)}{\epsilon^2 n^2}}
    .
    \]
On the other hand, we know that
    \begin{align*}
        \expect{\lnorm{\calA(\calD) - \overline{\calD}}{1}}
        &\le
        \sqrt{d}
        \expect{\lnorm{\calA(\calD) - \overline{\calD}}{2}}
        \le
        \sqrt{d}
        \sqrt{\expect{\lnorm{\calA(\calD) - \overline{\calD}}{2}^2}} 
        = o \pth{\frac{d \sqrt{\log(1/\delta)} }{\epsilon n}}  
        .
    \end{align*}     
    It further implies that there exists $0< \varrho = o (\frac{\sqrt{d\log(1/\delta)} } {\epsilon n}) $ (i.e., $n = o (\frac{\sqrt{d\log(1/\delta)} } {\epsilon \varrho}) $) such that  
    \begin{align*}
        \expect{\lnorm{\calA(\calD) - \overline{\calD}}{1}}
        &\le 
        \sqrt{d} \varrho
        .
    \end{align*}
    Invoking Lemma \ref{lmm: dp lower bound ref}, we obtain 
    $n = \Omega({\sqrt{d \log (1 / \delta)}}/{\varrho \epsilon})$, 
    contradicting with $n = o ({\sqrt{d \log (1 / \delta)}}/{\varrho \epsilon}).$ 
    Therefore,~\eqref{eq: lower bound non-adversarial} must be true. 

\end{proof}

\begin{corollary}
    \label{cor: dp sample complexity lower bound}
    Fix $\Delta>0$, and let $n, d \ge 1$. Define  
    $\calX = \sth{\pm \Delta / \sqrt{d}}^d $ and $\calY = [\pm \Delta / \sqrt{d}]^d$.
    Let $\calA: k\calX^n \to k\calY$ be an arbitrary scale-invariant mapping for any $k>0$, \ie, 
   $\calA(k \cdot \calD) = k \cdot \calA(\calD)$ for any dataset $\calD\in \calX^n$ and $k > 0$. 
    For any $\epsilon = O(1)$, and $2^{- o(n)} \le \delta \le 1 / n^{1 + \Omega (1)}$, 
    if $\calA$ is $(\epsilon, \delta)$-DP, then  
    \begin{align*}
    \expect{\norm{\calA(\calD) - \overline{\calD}}^2 }
    &= 
    \Omega (\frac{\Delta^2 {d \log(1/\delta)}}{\epsilon^2 n^2})
    ,
    \end{align*}
    where $\overline{\calD}$ is defined in~\prettyref{lmm: one-way marginal}. 
\end{corollary}
~\prettyref{cor: dp sample complexity lower bound} follows immediately from~\prettyref{lmm: one-way marginal} and the fact that $\calA$ is scale invariant.

\newpage
\section{Privacy Analysis}
\label{app: cl privacy analysis}

\begin{claim}
    \label{clm: coefficient order}
    When $n \ge (2 + \nu) f$, where $\nu > 0$, it holds that
    \[
    \pth{1 - \frac{2 f}{n}}^{-2} 
    \asymp 
    1 + \frac{f}{n}.
    \]
\end{claim}
\begin{proof}[\bf Proof of~\prettyref{clm: coefficient order}]
It is easy to see that
\[  
    \frac{1}{(n - 2f)^2} 
    = \frac{1}{n^2} \frac{1}{(1 - \frac{2 f}{n})^2}
    = \frac{1}{n^2} \pth{1 - \frac{2 f}{n}}^{-2}.
\]
By Bernoulli inequality, it holds that $\pth{1 - \frac{2 f}{n}}^{-2} \ge 1 + \frac{4 f}{n}.$
We know that
\begin{align*}
    \frac{1}{(1-x)^2} 
    &= \frac{(1-x)^2 + 1 - (1-x)^2}{(1-x)^2}
    = 1 + \frac{x (2 - x)}{(1-x)^2}
    \le 1 + \frac{2 x}{(1-x)^2}.
\end{align*}
Therefore,
\begin{align*}
     1 + \frac{4 f}{n} \le
    \pth{1 - \frac{2 f}{n}}^{-2}
    &\le
    1 + \frac{4 (2 + \nu)^2}{\nu^2} \frac{f}{n}.
\end{align*}
It follows that
\[
    \pth{1 - \frac{2 f}{n}}^{-2} 
    \asymp 
    1 + \frac{f}{n}.
\]
    
\end{proof}

\subsection{Proof of~\prettyref{lmm: 1-lipschitz cwtm}}
    We consider {\tt TM} as a continuous function of $h$. For ease of notation, let $L (h) \triangleq \abth{\sth{j: a_j < h}}$ be the rank of $h$ among the unchanged values. After insertion, its position becomes $L(h) + 1$. The value remains untrimmed iff
    \[
        f+1 \le L (h) + 1 \le n - f~~~
        \Leftrightarrow
        f \le L (h) \le n - f - 1.
    \]

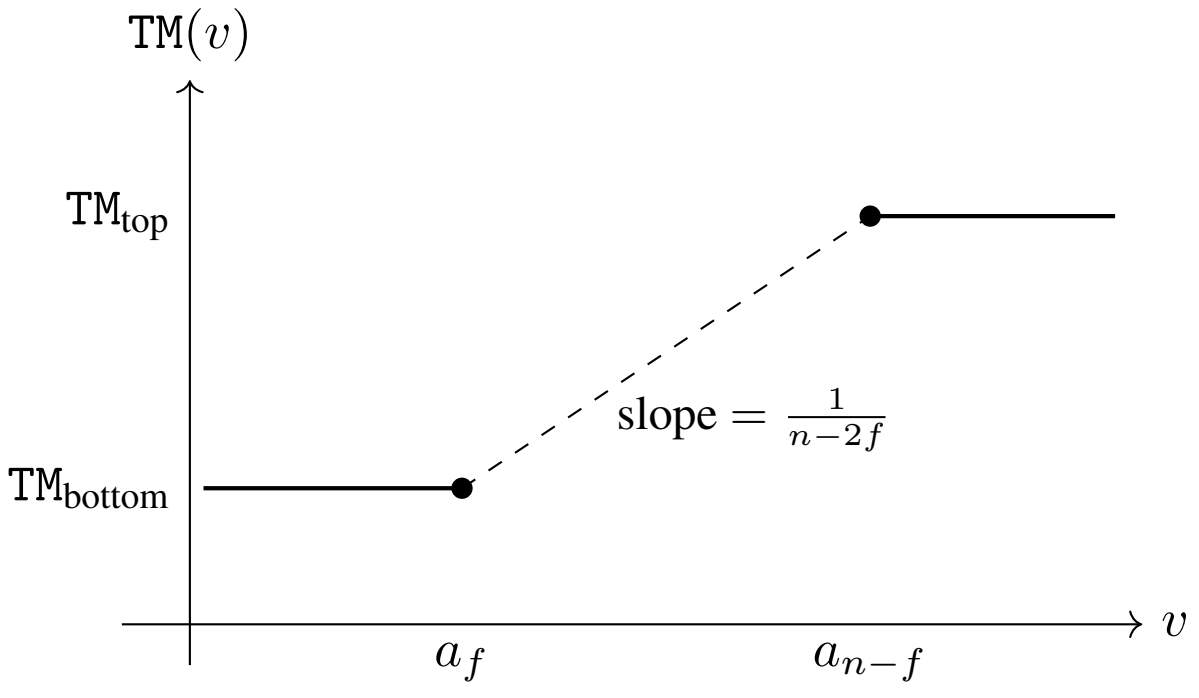
\begin{figure}[!ht]
        \centering
        \begin{tikzpicture}[scale=0.9]
  \draw[->] (-0.5,0) -- (7,0) node[right] {$v$};
  \draw[->] (0,-0.3) -- (0,4) node[above] {${\tt TM}(v)$};
  
  \draw[thick] (0.1,1) -- (2,1);
  \draw[thick] (5,3) -- (6.8,3);
  \draw[dashed] (2,1) -- (5,3);
  
  \filldraw (2,1) circle (2pt);
  \filldraw (5,3) circle (2pt);
  
  \node[left] at (0,1) {${\tt TM}_{\text{bottom}}$};
  \node[left] at (0,3) {${\tt TM}_{\text{top}}$};
  \node[below] at (2,0) {$a_f$};
  \node[below] at (5,0) {$a_{n-f}$};
  
  \node[right] at (3,1.5) {\small slope $= \frac{1}{n-2f}$};
\end{tikzpicture}
        \caption{Demonstration of {\tt CWTM} with respect to a coordinate-wise insertion value.}
        \label{fig:cwtm piecewise linear}
    \end{figure}

    \paragraph{Case \#1: $L(h) < f$.}
     $h$ is trimmed from below.
     \[
        {\tt TM} (h) = \frac{1}{n - 2f} \sum_{j=f}^{n-f-1} a_j ={\tt TM}_{\text{bottom}}.
     \]
    \paragraph{Case \#2: $f \le L(h) \le n - f $.}
    $h$ is kept.
    \[
        {\tt TM} (h) = \frac{1}{n - 2f} \pth{h + \sum_{j=f+1}^{n-f-1} a_j} = {\tt TM}_{\text{middle}} (h).
    \]
    \paragraph{Case \#3: $L(h) > n - f$}
    $h$ is trimmed from above.
    \[
        {\tt TM} (h) = \frac{1}{n - 2f} \sum_{j=f+1}^{n-f} a_j ={\tt TM}_{\text{top}}.
    \]
    It is easy to see that
    \begin{align*}
        {\tt TM}_{\text{middle}} (h)
        &=
        \frac{1}{n - 2f} \pth{h + \sum_{j=f+1}^{n-f-1} a_j} 
        \ge
        \frac{1}{n - 2f} \pth{a_{f} + \sum_{j=f+1}^{n-f-1} a_j}
        ={\tt TM}_{\text{bottom}},
    \end{align*}
    and
    \begin{align*}
        {\tt TM}_{\text{middle}} (h)
        &=
        \frac{1}{n - 2f} \pth{h + \sum_{j=f+1}^{n-f-1} a_j} 
        \le
        \frac{1}{n - 2f} \pth{a_{n - f} + \sum_{j=f+1}^{n-f-1} a_j}
        ={\tt TM}_{\text{top}},
    \end{align*}
    where both inequalities hold because $f \le L (h) \le n - f$.
    Therefore, we have
    \begin{align*}
        {\tt TM}(h)
        &=
        \begin{cases}
            {\tt TM}_{\text{bottom}}, &~\text{if}~h < a_{f};\\
            \frac{1}{n - 2f} \pth{h + \sum_{j=f+1}^{n-f-1} a_j}, &~\text{if}~a_{f } \le h \le a_{n - f};\\
            {\tt TM}_{\text{top}}, &~\text{if}~h > a_{n - f}.\\
        \end{cases}
    \end{align*}
    With a bit of abuse of notations, it follows that
    \begin{align*}
        \frac{\mathd {\tt TM}(h)}{\mathd h}
        &=
        \begin{cases}
            \frac{1}{n - 2f},
            &~\text{if}~a_{f } \le h \le a_{n - f};\\
            0, &~\text{otherwise}.
        \end{cases}
    \end{align*}
    \prettyref{fig:cwtm piecewise linear} confirms this pictorially.
    Without loss of generality, we assume $h < h^{\prime}$. 
    Therefore,
    \begin{align*}
        \abth{{\tt TM} (h) - {\tt TM}(h^{\prime})}
        &=
        \abth{\int_{h}^{h^{\prime}} \frac{\mathd {\tt TM}(h)}{\mathd h} \mathd h}
        \le
        \frac{1}{n - 2f}
        \int_{h}^{h^{\prime}} \mathd h
        =
        \frac{\abth{h - h^{\prime}}}{n - 2f}
        \le
        \frac{\abth{h - h^{\prime}}}{n}
        \sqrt{1 + \frac{4(2+ \nu)^2}{\nu^2} \frac{f}{n}}
        ,
    \end{align*}
    where the last inequality follows from~\prettyref{clm: coefficient order}.

\subsection{Proof of~\prettyref{prop: trimmed mean sensitivity}}
Let $i$ denote the replaced client in $\calG_1$, 
$j$ denote the replaced client in its adjacent healthy client set $\calG_2$.
Further, define ${\tt Diff}({\tt CWTM})$ as the difference between the {\tt CWTM} outputs of $\calG_1$ and $\calG_2$.

\paragraph{Step 1: One-dimensional analysis.}
From~\prettyref{lmm: 1-lipschitz cwtm}, we know that 
\begin{small}
\begin{align*}
    [\norm{{\tt Diff} ({\tt CWTM})}]_{r}
    &\le
    \frac{\abth{[\x_i]_r - [\x_j]_r}}{n - 2f}
    =
    \frac{(1 - \beta_t) \abth{[ \m_i]_r - [\m_j]_r}}{n - 2f}.
\end{align*}
\end{small}

\paragraph{Step 2: High-dimensional analysis.}
By the definition of Byzantine user-level sensitivity, it holds that
\begin{small}
\begin{align*}
    \norm{{\tt Diff} ({\tt CWTM})}^2 
    &\le
    \frac{(1 - \beta_t)^2 \sum_{r=1}^{d} 
    \abth{[\m_i]_r - [\m_j]_r}^2}{(n - 2f)^2} \\
    &\le
    \frac{2 (1 - \beta_t)^2 \sum_{r=1}^{d} 
    \pth{\abth{[\m_i]_r}^2 + \abth{[\m_j]_r}^2}}{(n - 2f)^2}  \\
    &=
    \frac{2 (1 - \beta_t)^2 
    \pth{\norm{\m_i}^2 + \norm{\m_j}^2}}{(n - 2f)^2} \\
    &\le
    \frac{4 (1 - \beta_t)^2 C^2}{(n - 2f)^2}.
\end{align*}
\end{small}

\paragraph{Step 4: Combining together.}
\begin{align*}
    \norm{{\tt Diff} ({\tt CWTM})}^2 
    &\le
    \frac{4 (1 - \beta_t)^2 C^2}{(n - 2f)^2}
    \le
    \frac{4 (1 - \beta_t)^2 C^2}{n^2}
    \pth{1 + \frac{4 (2 + \nu)^2}{\nu^2} \frac{f}{n}}.
\end{align*}

Therefore, we have the Byzantine user-level sensitivity of {\tt CWTM} as
\[
    \pth{\Delta ({\tt CWTM})}^2 
    =
    \max \norm{{\tt Diff} ({\tt CWTM})}^2 
    =
    \frac{4 (1 - \beta_t)^2 C^2}{n^2}
    \pth{1 + \frac{4 (2 + \nu)^2}{\nu^2} \frac{f}{n}}.
\]

\subsection{Supporting~\renyi~DP Analysis}
\label{app: supporting renyi}

Our privacy analysis is based on \renyi~differential privacy (RDP)
for its tight composition guarantees.
Its guarantee can be translated back to DP.
First, we introduce the notions of RDP.
\begin{definition}[$(\alpha, \epsilon)$-RDP, Definition 4 in \cite{mironov2017renyi}]
    \label{def: rdp}
    A randomized mechanism $\calM: \calD \rightarrow \calR$ is said to have $\epsilon$-\renyi~differential privacy of order $\alpha$, or $(\alpha, \epsilon)$-RDP for short, if for any adjacent $\calD, \calD^\prime \in \X$, it holds that
    \[
        D_{\alpha}
        \pth{
        \calM (\calD) \| \calM (\calD^\prime)
        }
        \le \epsilon,
    \]
    where $D_{\alpha} (\calM (\calD) \| \calM (\calD^\prime)) \triangleq \frac{1}{\alpha - 1}\log \Expect_{x \sim \calM (\calD^\prime)} (\calM (\calD)(x) / \calM (\calD^\prime) (x))^\alpha$ defines the \renyi~divergence of order $\alpha > 1$ for two probability distributions $\calM (\calD)$ and $\calM (\calD^\prime)$.
\end{definition}
Next, we introduce the connections between \renyi~differential privacy and differential privacy \cite{dwork2014algorithmic}.
\begin{lemma}[RDP to DP, Proposition 3 in~\cite{mironov2017renyi}]
    \label{lmm: rdp to dp}
    If $\calM$ is an $(\alpha, \epsilon)$-RDP mechanism, it also satisfies $(\epsilon + \log (1 / \delta) / (\alpha - 1), \delta)$-differential privacy for any $\delta \in (0,1)$.
\end{lemma}
It is well known that the Gaussian mechanism can also work under \renyi~differential privacy.
\begin{lemma}[Gaussian mechanism, Corollary 3 in~\cite{mironov2017renyi}]
    \label{lmm: rdp gaussian}
    Define Gaussian mechanism $\bm{G}_{\sigma_{dp}} f = f + \calN (0, \sigma^2_{dp} \identity)$, where $f$ is a function with $l_2$ sensitivity $\Delta$.
    The gaussian mechanism $\bm{G}_{\sigma_{dp}} f$ satisfies $(\alpha, \alpha \Delta^2 / 2 \sigma_{dp}^2)$-RDP.
\end{lemma}

\prettyref{lmm: rdp adaptive composition} characterizes the adaptive composition in RDP.
\begin{lemma}[Adaptive composition, Proposition 1 in \cite{mironov2017renyi}]
    \label{lmm: rdp adaptive composition}
    Let $\calM_1: \calD \rightarrow \calR_1$ be $(\alpha, \epsilon_1)$-RDP and $g: \calR_1 \times \calD \rightarrow \calR_2$ be $(\alpha, \epsilon_2)$-RDP, then the mechanism  defined as $(X,Y)$, where $X \sim \calM_1 (\calD)$ and $Y \sim \calM_2 (X, \calD)$, satisfies $(\alpha, \epsilon_1 + \epsilon_2)$-RDP.
\end{lemma}

\begin{lemma}
    \label{lmm: cl privacy guarantees}
    Let $\delta \in (0,1)$,~\prettyref{alg: byzantine dp learning base stochastic v2} is a $(\epsilon^{\ast}, \delta)$ differentially private distributed learning algorithm with
    \[
        \epsilon^{\ast}
        =
        \inf_{\alpha > 1}
        \pth{
        T \epsilon(\alpha) + \frac{\log (1 / \delta)}{\alpha - 1}
        },
    \]
    where $\epsilon (\alpha) = \frac{\alpha \Delta_r^2}{2 \sigma_{dp,r}^2} = \frac{\alpha \Delta^2}{2 \sigma_{dp}^2}$ is the per-round RDP budget.
\end{lemma}

\begin{proof}[Proof of~\prettyref{lmm: cl privacy guarantees}]
    We characterize the privacy guarantees of~\prettyref{alg: byzantine dp learning base stochastic v2} based on RDP.
    This includes two steps:
    i) derive guarantees in a single global round for all clients, and
    ii) compose the single-round guarantees to account for the final privacy loss.
    \paragraph{Step 1: single-round guarantee.}
    \prettyref{def: dp federated learning} defines the notion of differential privacy at a system level.
    It shows that the output model parameter should be nearly indistinguishable when two adjacent client sets differ by only one client.

    We track the privacy loss through the robust gradient descent step (line 7) in~\prettyref{alg: byzantine dp learning base stochastic v2}.
    In~\prettyref{prop: trimmed mean sensitivity}, we have derived the Byzantine user-level sensitivity.

    \paragraph{Step 2: $T$ rounds composition.}
    By~\prettyref{lmm: rdp adaptive composition}, we know that~\prettyref{alg: byzantine dp learning base stochastic v2} is $(\alpha, T \epsilon (\alpha))$-RDP after $T$ rounds of execution, where $\epsilon (\alpha) = \alpha \Delta^2 / 2 \sigma_{dp}^2$ is the per-round RDP budget.
    
    Now, we convert RDP guarantees back to differential privacy guarantees.
    By~\prettyref{lmm: rdp to dp}, it holds that~\prettyref{alg: byzantine dp learning base stochastic v2} is $(\epsilon_{dp}, \delta)$-differentially private, where $\delta \in (0,1)$ and
    \begin{align}
        \label{eq: cl dp budget from rdp}
        \epsilon_{dp} 
        =
        \sum_{r=0}^{T-1}
        \frac{\alpha \Delta_{r}^2}{2 \sigma_{dp,r}^2} + \frac{\log (1 / \delta)}{\alpha - 1}
        =
        \frac{\alpha T \Delta^2}{2 \sigma_{dp}^2} + \frac{\log (1 / \delta)}{\alpha - 1}
        ,
    \end{align}
    where the last equality follows from $\frac{\Delta_{r}^2}{\sigma_{dp,r}^2} = \frac{\Delta^2}{\sigma_{dp}^2}$ for $r \in [T]$.
\end{proof}

\subsection{Proof of~\prettyref{thm: cl rdp guarantees}}
    By~\prettyref{lmm: cl privacy guarantees}, since~\eqref{eq: cl dp budget from rdp} holds for any $\alpha > 1$, we can optimize it to obtain a tighter budget.
    \begin{align*}
        \frac{\partial \epsilon_{dp}}{\partial \alpha} 
        =
        \frac{T \Delta_{\text{Byz},r}^2}{2 \sigma_{dp,r}^2} - \frac{\log (1 / \delta)}{(\alpha - 1)^2}
        = 0 
        \Leftrightarrow
        \alpha = 1 +
        \frac{\sigma_{dp,r}}{\Delta_{\text{Byz},r}}
        \sqrt{
        \frac{2 \log (1 / \delta)}{T}
        } > 1
        ,
    \end{align*}
    which implies that
    \begin{align*}
        \epsilon_{dp}^{\ast}
        &=
        \frac{T \Delta_{\text{Byz},r}^2}{2 \sigma_{dp,r}^2} 
        \pth{
        1 +
        \frac{\sigma_{dp,r}}{\Delta_{\text{Byz},r}}
        \sqrt{
        \frac{2 \log (1 / \delta)}{T}
        }
        }
        + 
        \frac{\log (1 / \delta)}{
        \frac{\sigma_{dp,r}}{\Delta_{\text{Byz},r}}
        \sqrt{
        \frac{2 \log (1 / \delta)}{T}
        }}
        =
        \frac{T \Delta_{\text{Byz},r}^2}{2 \sigma_{dp,r}^2} 
        +
        \frac{\Delta_{\text{Byz},r} \sqrt{
        2 T \log (1 / \delta)
        }}{ \sigma_{dp,r}} 
        .
    \end{align*}
    When $\sigma_{dp,r} \ge 2 \Delta_{\text{Byz},r} \sqrt{2 T \log (1 / \delta)} / \epsilon$ and $\epsilon \le 8 \log (1 / \delta)$, it holds that
    \begin{align*}
        \epsilon_{dp}^{\ast}
        &=
        \frac{T \Delta_{\text{Byz},r}^2}{2 \sigma_{dp,r}^2} 
        +
        \frac{\Delta_{\text{Byz},r} \sqrt{
        2 T \log (1 / \delta)
        }}{ \sigma_{dp,r}} 
        \le
        \frac{\epsilon^2}{16 \log (1 / \delta)}
        +
         \frac{\epsilon}{2}
        \le
        \frac{\epsilon}{2}
        +
        \frac{\epsilon}{2}
        =
        \epsilon.
    \end{align*}

\subsection{Byzantine-robust User-level Sensitivity Analysis of Additional Robust Aggregators}
\label{app: additional sensitivity}

\begin{table}[!t]
    \centering
    \caption{\small
    Byzantine-robust user-level sensitivity when the honest inputs to the corresponding robust aggregator are bounded by $C$ or $\tau$ in the case of CClip.
    }
    \label{tab: cl byzantine sensitivity}
    \resizebox{\textwidth}{!}{
    \begin{tabular}{ccccccc}
    \toprule 
    \begin{tabular}{c}
     Robust \\
     Aggregator  
     \end{tabular}& 
        Trimmed Mean &
        Coordinate Median &
        Geometric Median &
        Krum &
        CClip &
        Average (fault-free)\\[0.05cm]
        \hline\\
        Sensitivity &  
        $\frac{2 C}{n - 2 f}$ &
        $2 C \sqrt{d}$ &
        $2 C$ &
        $2 C$ &
        $\frac{2 \tau}{n}$ &
        $\frac{2 C}{n}$ \\
        \bottomrule
    \end{tabular}}
\end{table}

In this section, we characterize the sensitivity of robust aggregators when a single honest client is replaced.
When updates from honest clients are bounded, updates from Byzantine clients must also be bounded in the worst case. Otherwise, it is easy for the aggregator to filter out their presence.
In the remainder of this section, we consider $\{\x_i\}_{i \in [n]}$ as the inputs to the robust aggregator.
Note that we have $\x_i^{t} = \M_i^t = \beta_t \tilde R^{t-1} + (1 - \beta_t) \hat \m_i^t$ in~\prettyref{alg: byzantine dp learning base stochastic v2}.

\paragraph{Krum.}
Let $\calN_j$ denote the set of indices of the $n - f$ nearest neighbors of $\x_j$ in $\{\x_1, \ldots, \x_n \}$, with ties arbitrarily broken.
{\tt Krum} outputs the vector $\x_{k^\ast}$ such that
\[
    k^{\ast} \in \arg\min_{j \in [n]} \sum_{i \in \calN_j}
    \norm{\x_j - \x_i}^2,
\]
with ties arbitrarily broken if the set of minimizers above includes more than one element.

\begin{example}
    \label{eg: krum sensitivity change}
    Suppose that we have a collection of five clients' updates $\{\x_1, \ldots, \x_5\}$ such that $\norm{\x_i} = C$ for $i \in [5]$.
    Further, we assume $f=1$ Byzantine client is $\x_5$.
    There exist client placements where the Byzantine user-level sensitivity of ${\tt Krum}$ is $2C$.
\end{example}
\begin{proof}[\bf Proof of~\prettyref{eg: krum sensitivity change}]
    We consider two adjacent configurations, where clients are divided into two clusters $\calC_1$ and $\calC_2$.
    Let the clients in each cluster reside at the same point, \ie, $\norm{\x_i - \x_j} = 0$ for $i, j \in \calC_1$ and for $i, j \in \calC_2$, respectively.
    Moreover, let $\norm{\x_i - \x_j} = 2 C$ for all $i \in \calC_1$ and all $j \in \calC_2$.
    \paragraph{Configuration 1.}
    Let $\calC_1 = \{ \x_1, \x_2, \x_3 \}$, and $\calC_2 = \{ \x_4, \x_5 \}$.
    It can be readily checked that 
    \[
    {\tt Krum}(\x_1, \ldots, \x_5) = \x_{\calC_1},
    \]
    which denotes any points in $\calC_1$.
    \paragraph{Configuration 2.}
    Without loss of generality, we replace client $\x_1$ in $\calC_1$ with an auxiliary client $\x^{\prime} \in \calC_2$.
    Therefore, it holds that $\calC_1 = \{\x_2, \x_3\}$ and $\{\x_4, \x_5, \x^\prime \}$.
    It can be readily checked that 
    \[
    {\tt Krum}(\x^\prime, \ldots, \x_5) = \x_{\calC_2},
    \]
    which denotes any points in $\calC_2$.
    
    Therefore, the Byzantine user-level sensitivity of {\tt Krum} in this instance is $2C$.
\end{proof}

\begin{proposition}
\label{prop: krum sensitivity}
Let $m \in \naturals$ and $f < n/2$.
The Byzantine-robust user-level sensitivity of ${\tt Krum}$ is $2 C$ if $\norm{\x_i} \le C$ for $i \in [n]$.
\end{proposition}
\begin{proof}[\bf Proof of~\prettyref{prop: krum sensitivity}]
    We prove this by contradiction based on~\prettyref{eg: krum sensitivity change}.
    Suppose that we have the Byzantine user-level sensitivity of {\tt Krum} is smaller than $2 C$.
    Contradiction with~\prettyref{eg: krum sensitivity change}.
    Therefore, it must be true that $\Delta({\tt Krum}) = 2 C$.

    It is a rather trivial bound. By the triangle inequality, we have
    \[
        \Delta ({\tt Krum})
        = 
        \norm{\x_{k_1} - \x_{k_2}}
        \le
        \norm{\x_{k_1}}
        +
        \norm{\x_{k_2}}
        \le 2 C
        ,
    \]
    where $k_1$ and $k_2$ denote the indices of the minimizers of two adjacent honest client sets, respectively.
\end{proof}
\paragraph{Median.}
For input vectors $\x_1, \ldots, \x_n$, their coordinate-wise median, denoted by ${\tt CWMed}(\x_1, \ldots, \x_n)$, is defined to be a vector whose $k$-th coordinate, for all $k \in [d]$, is defined to be
\[
    [{\tt CWMed}(\x_1, \ldots, \x_n)]_{k}
    \triangleq
    {\tt Median}
    ([\x_1]_{k}. \ldots, [\x_n]_{k})
    .
\]

\begin{example}
    \label{eg: median sensitivity change}
    Suppose that we have a collection of five clients' updates $\{\x_1, \ldots, \x_5\}$ such that $\norm{\x_i} = C$ for $i \in [5]$.
    Further, we assume $f=1$ Byzantine client is $\x_5$.
    There exist client placements where the Byzantine user-level sensitivity of ${\tt CWMedian}$ is $2C$.
\end{example}
\begin{proof}[\bf Proof of~\prettyref{eg: median sensitivity change}]
    We consider two adjacent configurations, where clients are divided into two clusters $\calC_1$ and $\calC_2$.
    Let the clients in each cluster reside at the same point, \ie, $\norm{\x_i - \x_j} = 0$ for $i, j \in \calC_1$ and for $i, j \in \calC_2$, respectively.
    Moreover, let $\norm{\x_i - \x_j} = 2 C$ for all $i \in \calC_1$ and all $j \in \calC_2$.
    More specifically, let 
    \begin{align*}
        \x_i = 
        \begin{cases}
        (\underbrace{C, 0, \ldots, 0}_{d~\text{dimensions}}), 
        ~&\text{if}~ i \in \calC_1;\\
        (\underbrace{-C, 0, \ldots, 0}_{d~\text{dimensions}}), 
        ~&\text{if}~ i \in \calC_2.
        \end{cases}
    \end{align*}
    \paragraph{Configuration 1.}
    Let $\calC_1 = \{ \x_1, \x_2, \x_3 \}$, and $\calC_2 = \{ \x_4, \x_5 \}$.
    It can be readily checked that 
    \[
    {\tt CWMed}(\x_1, \ldots, \x_5) = (C, 0, \ldots, 0).
    \]
    \paragraph{Configuration 2.}
    Without loss of generality, we replace client $\x_1$ in $\calC_1$ with an auxiliary client $\x^{\prime} \in \calC_2$.
    Therefore, it holds that $\calC_1 = \{\x_2, \x_3\}$ and $\{\x_4, \x_5, \x^\prime \}$.
    It can be readily checked that 
    \[
    {\tt CWMed}(\x^\prime, \ldots, \x_5) = (-C, 0, \ldots, 0).
    \]
    Therefore, the Byzantine user-level sensitivity of {\tt CWMed} in this instance is $2C$.
\end{proof}
\begin{proposition}
\label{prop: cwmed sensitivity}
Let $m \in \naturals$ and $f < n/2$.
The Byzantine-robust user-level sensitivity of ${\tt CWMed}$ is at least $2 C$ if $\norm{\x_i} \le C$ for $i \in [n]$.
\end{proposition}
\begin{proof}[\bf Proof of~\prettyref{prop: cwmed sensitivity}]
    We prove this by contradiction based on~\prettyref{eg: median sensitivity change}.
    Suppose that we have the Byzantine user-level sensitivity of {\tt CWMed} is smaller than $2 C$.
    Contradiction with~\prettyref{eg: median sensitivity change}.
    Therefore, it must be true that $\Delta({\tt CWMed})$ is at least $2 C$.
\end{proof}

\paragraph{Geometric median.}
The geometric median of $\x_1, \ldots, \x_n \in \reals^d$, denoted by ${\tt GM} (\x_1, \ldots, \x_n)$, is defined to be a vector that minimizes the sum of the $l_2$ distances to these vectors:
\[
    {\tt GM} (\x_1, \ldots, \x_n)
    \in 
    \arg \min_{\bm{y} \in \reals^d}
    \sum_{i=1}^n
    \norm{\bm{y} - \x_i}
    .
\]
\begin{example}
    \label{eg: geo median sensitivity change}
    Suppose that we have a collection of five clients' updates $\{\x_1, \ldots, \x_5\}$ such that $\norm{\x_i} = C$ for $i \in [5]$.
    Further, we assume $f=1$ Byzantine client is $\x_5$.
    There exist client placements where the Byzantine user-level sensitivity of ${\tt GM}$ is $2C$.
\end{example}
\begin{proof}[\bf Proof of~\prettyref{eg: geo median sensitivity change}]
    We reuse the instances from~\prettyref{eg: median sensitivity change} and consider two adjacent configurations, where clients are divided into two clusters $\calC_1$ and $\calC_2$.
    Let the clients in each cluster reside at the same point, \ie, $\norm{\x_i - \x_j} = 0$ for $i, j \in \calC_1$ and for $i, j \in \calC_2$, respectively.
    Moreover, let $\norm{\x_i - \x_j} = 2 C$ for all $i \in \calC_1$ and all $j \in \calC_2$.
    More specifically, let 
    \begin{align*}
        \x_i = 
        \begin{cases}
        (\underbrace{C, 0, \ldots, 0}_{d~\text{dimensions}}), 
        ~&\text{if}~ i \in \calC_1;\\
        (\underbrace{-C, 0, \ldots, 0}_{d~\text{dimensions}}), 
        ~&\text{if}~ i \in \calC_2.
        \end{cases}
    \end{align*}
    \paragraph{Configuration 1.}
    Let $\calC_1 = \{ \x_1, \x_2, \x_3 \}$, and $\calC_2 = \{ \x_4, \x_5 \}$.
    It can be readily checked that 
    \[
    {\tt GM}(\x_1, \ldots, \x_5) = (C, 0, \ldots, 0).
    \]
    \paragraph{Configuration 2.}
    Without loss of generality, we replace client $\x_1$ in $\calC_1$ with an auxiliary client $\x^{\prime} \in \calC_2$.
    Therefore, it holds that $\calC_1 = \{\x_2, \x_3\}$ and $\{\x_4, \x_5, \x^\prime \}$.
    It can be readily checked that 
    \[
    {\tt GM}(\x^\prime, \ldots, \x_5) = (-C, 0, \ldots, 0).
    \]
    Therefore, the Byzantine user-level sensitivity of {\tt CWMed} in this instance is $2C$.
\end{proof}
\begin{proposition}
\label{prop: geo median sensitivity}
Let $n \in \naturals$ and $f < n/2$.
The Byzantine-robust user-level sensitivity of {\tt GM} is $2 C$ if $\norm{\x_i} \le C$ for $i \in [n]$.
\end{proposition}
\begin{proof}[\bf Proof of~\prettyref{prop: geo median sensitivity}]
    We prove this by contradiction based on~\prettyref{eg: geo median sensitivity change}.
    Suppose that we have the Byzantine user-level sensitivity of {\tt GM} is smaller than $2 C$.
    Contradiction with~\prettyref{eg: geo median sensitivity change}.
    Therefore, it must be true that $\Delta({\tt GM})$ is at least $ 2 C$.

    Next, we show that $2 C$ is achievable. 
    Since ${\tt GM}$ is a convex operation over a set of inputs within a convex set ($\norm{\x} \le C$).
    It holds that $\norm{{\tt GM} (\x_1, \ldots, \x_n)} \le C$.
    Without loss of generality, we assume $\x_1 \in \calG$ is replaceped with $\x^\prime$.  It follows that
    \[
        \Delta ({\tt GM})
        =
        \norm{{\tt GM} (\x_1, \ldots, \x_n)
        -
        {\tt GM} (\x^\prime, \ldots, \x_n)
        }
        \le
        \norm{{\tt GM} (\x_1, \ldots, \x_n)}
        +
        \norm{{\tt GM} (\x^\prime, \ldots, \x_n)}
        \le
        2 C.
    \]
\end{proof}

\newpage
\section{Convergence Analysis}
\label{app: convergence analysis}

\subsection{Shared Intermediate Results}
\label{app: shared analysis}
For ease of notation, 
let $\overline{\M}^t \triangleq \frac{1}{\abth{\calG}} \sum_{i \in \calG} \M_i^t$, 
$\overline{\m}^t \triangleq \frac{1}{\abth{\calG}} \sum_{i \in \calG} \m_i^t$,
and $\overline{\g}^t \triangleq \frac{1}{\abth{\calG}} \sum_{i \in \calG} \g_i^t$.

\subsubsection{Proof of Descent Lemma}
\begin{lemma}
    \label{lmm: stochastic v2 descent}
    Under~\prettyref{ass: smmothness}, it holds that
    \begin{align*}
        &\expect{\byzobj (\bfthe^{t+1}) - \byzobj (\bfthe^{t}) ~\big|~ \calF^t}
        \le
        - \eta_{t}
        \pth{\frac{3}{4} - 2 \eta_{t} L}
        \norm{\nabla \byzobj (\bfthe^{t})}^2 
        +
        \frac{\eta_{t}^{2} L d}{2}
        \sigma_{dp,t}^2 \\
        &
        +
        2 \eta_{t} (1 + \eta_{t} L)
        \expect{
        \norm{\nabla \byzobj (\bfthe^t) - \overline{\M}^t}^2
        \Big|~\calF^t
        }
        +
        \frac{2
        \kappa  
        \eta_{t} (1 + \eta_{t} L)
        }{\abth{\calG}}
        \sum_{i \in \calG}
        \expect{\norm{
        \M_i^{t}
        -
        \overline{\M}^t
        }^2
        \Big|~\calF^t}
        .
    \end{align*}
\end{lemma}

\begin{proof}[\bf Proof of~\prettyref{lmm: stochastic v2 descent}]
    By~\prettyref{ass: smmothness}, it holds that
    \begin{align*}
        \byzobj (\bfthe^{t+1}) - \byzobj (\bfthe^{t})
        &\le
        \underbrace{
        \iprod{\nabla \byzobj (\bfthe^t)}{\bfthe^{t+1} - \bfthe^{t}}
        }_{(A)}
        +
        \underbrace{
        \frac{L}{2}
        \norm{\bfthe^{t+1} - \bfthe^{t}}^2
        }_{(B)}.
    \end{align*}

    \paragraph{Bounding $(A)$.}
    \begin{align*}
    &\iprod{\nabla \byzobj (\bfthe^{t})}{\bfthe^{t+1} - \bfthe^{t}} \\
    &=
    \eta_{t}
    \iprod{\nabla \byzobj (\bfthe^{t})}{
    \overline{\M}^t - R^t 
    }
    -
    \eta_{t}
    \iprod{\nabla \byzobj (\bfthe^{t})}{
    \overline{\M}^t
    -
    \nabla \byzobj (\bfthe^t)
    } 
    -
    \eta_{t} 
    \norm{\nabla \byzobj (\bfthe^t)}^2 
    -
    \eta_{t}
    \iprod{\nabla \byzobj (\bfthe^{t})}{
    \X_{dp,t}^{t}
    }  \\
    &\overset{(a)}{\le}
    \frac{\eta_{t}}{8}
    \norm{\nabla \byzobj (\bfthe^{t})}^2
    +
    2
    \eta_{t}
    \norm{R^t - \overline{\M}^t}^2
    +
    \frac{\eta_{t}}{8}
    \norm{\nabla \byzobj (\bfthe^{t})}^2
    +
    2 \eta_{t}
    \norm{
    \overline{\M}^t
    -
    \nabla \byzobj (\bfthe^{t})
    }^2
    -
    \eta_{t} 
    \norm{\nabla \byzobj (\bfthe^t)}^2 \\
    &~~~-
    \eta_{t}
    \iprod{\nabla \byzobj (\bfthe^{t})}{
    \X_{dp,t}^{t}
    }  \\
    &=
    -
    \frac{3 \eta_{t}}{4} 
    \norm{\nabla \byzobj (\bfthe^t)}^2
    +
    2
    \eta_{t}
    \norm{R^t - \overline{\M}^t}^2
    +
    2 \eta_{t}
    \norm{
    \overline{\M}^t
    -
    \nabla \byzobj (\bfthe^{t})
    }^2
    -
    \eta_{t}
    \iprod{\nabla \byzobj (\bfthe^{t})}{
    \X_{dp,t}^{t}
    } 
    ,
\end{align*}
where inequality $(a)$ follows from Young's inequality.
Taking expectation conditional on $\calF^t$, it holds that
\begin{align*}
    &\expect{
    \iprod{\nabla \byzobj (\bfthe^{t})}{\bfthe^{t+1} - \bfthe^{t}} 
    \mid \calF^t
    } \\
    &\le
    -
    \frac{3 \eta_{t}}{4} 
    \norm{\nabla \byzobj (\bfthe^t)}^2
    +
    2
    \eta_{t}
    \expect{\norm{R^t - \overline{\M}^t}^2 \mid \calF^t}
    +
    2 \eta_{t}
    \expect{\norm{
    \overline{\M}^t
    -
    \nabla \byzobj (\bfthe^{t})
    }^2 \mid \calF^t}
    -
    \eta_{t}
    \iprod{\nabla \byzobj (\bfthe^{t})}{
    \expect{\X_{dp,t}^{t} \mid \calF^t}
    } \\
    &=
    -
    \frac{3 \eta_{t}}{4} 
    \norm{\nabla \byzobj (\bfthe^t)}^2
    +
    2
    \eta_{t}
    \expect{\norm{R^t - \overline{\M}^t}^2 \mid \calF^t}
    +
    2 \eta_{t}
    \expect{\norm{
    \overline{\M}^t
    -
    \nabla \byzobj (\bfthe^{t})
    }^2 \mid \calF^t}.
\end{align*}

\paragraph{Bounding $(B)$.}
\begin{align*}
    \frac{L}{2}
    \norm{\bfthe^{t+1} - \bfthe^{t}}^2
    &=
    \frac{\eta_{t}^2 L}{2}
    \norm{
    R^t 
    - \overline{\M}^t 
    + \overline{\M}^t 
    +
    \X_{dp,t}^{t}
    }^2 
    \\
    &=
    \frac{\eta_{t}^2 L}{2}
    \pth{
    \norm{
    R^t 
    - \overline{\M}^t 
    + \overline{\M}^t}^2
    +
    \norm{\X_{dp,t}^{t}}^2
    +
    2 \iprod{
    R^t 
    - \overline{\M}^t  + \overline{\M}^t
    }{
    \X_{dp,t}^{t}
    }
    }
    .
\end{align*}
Taking expectation conditional on $\calF^t$, it holds that
\begin{align*}
    &\frac{L}{2}
    \expect{\norm{\bfthe^{t+1} - \bfthe^{t}}^2 \mid \calF^t}
    =
    \frac{\eta_{t}^2 L}{2}
    \Big (
    \expect{\norm{
    R^t 
    - \overline{\M}^t 
    + \overline{\M}^t 
    - \nabla \byzobj(\bfthe^t)
    + \nabla \byzobj(\bfthe^t)
    }^2 \mid \calF^t}
    +
    \expect{\norm{\X_{dp,t}^{t}}^2 \mid \calF^t} \\
    &~~~+
    2 \expect{\iprod{
    R^t 
    - \overline{\M}^t 
    + \overline{\M}^t
    }{
    \X_{dp,t}^{t}
    } \mid \calF^t}
    \Big ) \\
    &\overset{(b)}{=}
    \frac{\eta_{t}^2 L}{2}
    \Big (
    \expect{\norm{
    R^t 
    - \overline{\M}^t 
    + \overline{\M}^t 
    - \nabla \byzobj(\bfthe^t)
    + \nabla \byzobj(\bfthe^t)
    }^2 \mid \calF^t}
    +
    \expect{\norm{\X_{dp,t}^{t}}^2 \mid \calF^t} 
    \\
    &~~~
    +
    2 \iprod{
    \expect{R^t 
    - \overline{\M}^t 
    + \overline{\M}^t \mid \calF^t}
    }{
    \expect{\X_{dp,t}^{t} \mid \calF^t}
    } 
    \Big ) \\
    &=
    \frac{\eta_{t}^2 L}{2}
    \pth{
    \norm{
    R^t 
    - \overline{\M}^t 
    + \overline{\M}^t
    - \nabla \byzobj(\bfthe^t)
    + \nabla \byzobj(\bfthe^t)
    }^2
    +
    d
    \sigma_{dp,t-1}^2}
    \\
    &\overset{(c)}{\le}
    \eta_{t}^2 L
    \pth{
    2 \norm{
    R^t 
    - \overline{\M}^t 
    }^2
    +
    2
    \norm{
    \overline{\M}^t
    - \nabla \byzobj(\bfthe^t)
    }^2
    +
    2
    \norm{
    \nabla \byzobj(\bfthe^t)
    }^2
    +
    \frac{d}{2}
    \sigma_{dp,t-1}^2} \\
    &\overset{(d)}{\le}
    \eta_{t}^2 L
    \pth{
    \frac{2\kappa}{\abth{\calG}}
    \sum_{i \in \calG}
    \norm{
    \M_i^t
    -
    \overline{\M}^t
    }^2
    +
    2
    \norm{
    \overline{\M}^t
    - \nabla \byzobj(\bfthe^t)
    }^2
    +
    2
    \norm{
    \nabla \byzobj(\bfthe^t)
    }^2
    +
    \frac{d}{2}
    \sigma_{dp,t-1}^2}
    ,
\end{align*}
where equality $(b)$ holds because $\X_{dp}^t$ and $R^t$ are independent,
inequality $(c)$ follows from Jensen's inequality,
and inequality $(d)$ follows from the definition of robust aggregator.
\paragraph{Putting them together.}
\begin{align*}
    \expect{\byzobj (\bfthe^{t+1}) - \byzobj (\bfthe^{t}) \mid \calF^t}
    &\le
    - \eta_{t}
    \pth{\frac{3}{4} - 2 \eta_{t} L}
    \norm{\nabla \byzobj (\bfthe^{t})}^2
    +
    \frac{2
    \kappa  
    \eta_{t} (1 + \eta_{t} L)
    }{\abth{\calG}}
    \sum_{i \in \calG}
    \expect{\norm{
    \M_i^{t}
    -
    \overline{\M}^t
    }^2
    \mid
    \calF^t}
    \\
    &~~~
    +
    2 \eta_{t} (1 + \eta_{t} L)
    \expect{
    \norm{\nabla \byzobj(\bfthe^t) - \overline{\M}^t}^2
    \mid \calF^t
    }
    +
    \frac{\eta_{t}^2 L d}{2}
    \sigma_{dp,t-1}^2
    .
\end{align*}

\end{proof}

\subsubsection{Proof of Global Momentum Estimation Error}
\label{app: momentum deviataion decay v2 coefficient ref}
\begin{lemma}
    \label{lm: momentum deviation decay v2}
    Let $\calF^{t-1}$ be the sigma algebra by the randomness up to round $t-1$.
Suppose that Assumptions~\ref{ass: smmothness}, and~\ref{ass: bounded sto gradient cl} hold, we have for some absolute constant $k_2 > 0$
\begin{align*}
        \expect{\norm{\overline{\M}^t - \nabla \byzobj (\bfthe^{t})}^2 \mid \calF^{t-1}}
        &\le
        (1 - \beta_{t})^2
        (1 - \alpha_{t})^2 
        \frac{\sigma^2}{\abth{\calG}} 
        +
        C_{\M,1} 
        \expect{\norm{\overline{\M}^{t-1} - \nabla \byzobj(\bfthe^{t-1})}^2  \mid \calF^{t-1}}\\
        &~~~+
        C_{\M,2} 
       \expect{\norm{
        \overline{\m}^{t-1}
        -
        \nabla \byzobj (\bfthe^{t-1})
        }^2 \mid \calF^{t-1}}
        +
        C_{\M,3}
        \norm{\nabla \byzobj (\bfthe^{t-1})}^2 \\
        &~~~+
        C_{\M,4}
        \frac{\kappa}{\abth{\calG}}
        \sum_{i \in \calG}
        \expect{\norm{\m_i^{t-1} - \overline{\m}^{t-1}}^2 \mid \calF^{t-1}} 
        +
        C_{\M,5}
        d \sigma_{dp,t-1}^2
        .
    \end{align*}
    where
    \begin{align*}
        C_{\M,1} (\alpha_t, \beta_t, \eta_{t-1} L) &\triangleq
        \beta_{t} \pth{
        \beta_{t}  +
        \frac{(1 - \beta_{t})\alpha_{t} +
        (1 - \beta_{t-1}) \alpha_{t}}{2 k_2}
        +
        \eta_{t-1} L
        (\beta_{t} + (1 - \beta_{t}) \alpha_{t})
        } \\
        &~~~+
        3 \eta_{t-1} L
        \pth{\beta_{t} + (1 - \beta_{t}) \alpha_{t}}
        \pth{\pth{\beta_{t} + (1 - \beta_{t}) \alpha_{t}}
        \pth{1 +\eta_{t-1} L} + \beta_{t} (1 - \beta_{t})}
        ,
        \\
        C_{\M,2} (\alpha_t, \beta_t, \eta_{t-1} L) &\triangleq
        (1 - \beta_{t}) \alpha_{t}
        \qth{\pth{(1 - \beta_{t}) \alpha_{t}
        +
        2 k_2 \beta_{t}}
        \pth{
         1
        +
        \eta_{t-1} L
        } + (1 - \beta_{t}) \beta_{t}}
        ,
        \\
        C_{\M,3} (\alpha_t, \beta_t, \eta_{t-1} L) &\triangleq
         3 \eta_{t-1} L
        \pth{\beta_{t} + (1 - \beta_{t}) \alpha_{t}}
        \pth{\pth{\beta_{t} + (1 - \beta_{t}) \alpha_{t}}
        \pth{1 +\eta_{t-1} L} + \beta_{t} (1 - \beta_{t})}
        ,
        \\
        C_{\M,4} (\alpha_t, \beta_t, \eta_{t-1} L) &\triangleq
         \beta_{t}
        (1 - \beta_{t-1})
        \pth{2 k_2 \beta_{t} + (1 - \beta_{t}) \alpha_{t} + \frac{\eta_{t-1} L(1 - \beta_{t})}{1 - \beta_{t-1}} (\beta_{t} + (1-\beta_{t})\alpha_{t})} \\
        &~~~+
        3 (1 - \beta_{t-1})^2
        \eta_{t-1} L
        \pth{\beta_{t} + (1 - \beta_{t}) \alpha_{t}}
        \pth{\pth{\beta_{t} + (1 - \beta_{t}) \alpha_{t}}
        \pth{1 +\eta_{t-1} L} + \beta_{t} (1 - \beta_{t})} \\
        C_{\M,5} (\alpha_t, \beta_t, \eta_{t-1} L) &\triangleq
        \pth{\beta_t^2
        +
        \eta_{t-1} L
        \pth{\beta_{t} + (1 - \beta_{t}) \alpha_{t}}
        \pth{\pth{\beta_{t} + (1 - \beta_{t}) \alpha_{t}}
        \pth{1 +\eta_{t-1} L} + \beta_{t} (1 - \beta_{t})}}
        .
    \end{align*}
\end{lemma}
\begin{proof}[\bf Proof of Lemma~\ref{lm: momentum deviation decay v2}]
    We know that
   \begin{align*}
        &\overline{\M}^t
        - 
        \nabla \byzobj (\bfthe^{t})
        =
        \beta_{t}
        \tilde R^{t-1}
        +
        (1 - \beta_{t})
        \overline{\m}^t
        - 
        \nabla \byzobj (\bfthe^{t}) \\
        &=
        \beta_{t} \X_{dp}^{t-1}
        +
        \beta_{t}
        \pth{R^{t-1}
        -
        \overline{\M}^{t-1}
        }
        +
        \beta_{t} 
        \pth{
        \overline{\M}^{t-1}
        -
        \nabla \byzobj (\bfthe^{t-1})
        }
        +
        (1 - \beta_{t})
        \pth{\overline{\m}^t
        -
        \nabla \byzobj (\bfthe^{t})
        }
        - 
        \beta_{t}
        \pth{
        \nabla \byzobj (\bfthe^{t})
        -
        \nabla \byzobj (\bfthe^{t-1})
        } \\
        &=
        \beta_{t} \X_{dp}^{t-1}
        +
        \beta_{t} 
        \pth{R^{t-1}
        -
        \overline{\M}^{t-1}
        }
        +
        \beta_{t}
        \pth{
        \overline{\M}^{t-1}
        -
        \nabla \byzobj (\bfthe^{t-1})
        }
        +
        (1 - \beta_{t})
        \pth{
        \alpha_{t} \overline{\m}^{t-1}
        +
        (1 - \alpha_{t}) \overline{\g}^t
        -
        \nabla \byzobj (\bfthe^{t})
        } \\
        &~~~
        - 
        \beta
        \pth{
        \nabla \byzobj (\bfthe^{t})
        -
        \nabla \byzobj (\bfthe^{t-1})
        } \\
        &=
        \beta_{t} \X_{dp}^{t-1}
        +
        \beta_{t} 
        \pth{R^{t-1}
        -
        \overline{\M}^{t-1}
        }
        +
        \beta_{t}
        \pth{
        \overline{\M}^{t-1}
        -
        \nabla \byzobj (\bfthe^{t-1})
        }
        +
        (1 - \beta_{t})
        \alpha_{t}
        \pth{
        \overline{\m}^{t-1}
        -
        \nabla \byzobj (\bfthe^{t-1})
        } \\
        &~~~
        +
        (1 - \beta_{t}) (1 - \alpha_{t})
        \pth{\overline{\g}^t
        -
        \nabla \byzobj (\bfthe^{t})}
        - 
        \pth{\beta_{t} + (1 - \beta_{t}) \alpha_{t}}
        \pth{
        \nabla \byzobj (\bfthe^{t})
        -
        \nabla \byzobj (\bfthe^{t-1})
        } 
        .
    \end{align*}
    It follows that
    \begin{align*}
        &\norm{\overline{\M}^t - \nabla \byzobj (\bfthe^{t})}^2 \\
        &=
        \norm{\beta_{t} \X_{dp}^{t-1}}^2
        +
        \norm{
        \beta_{t} 
        \pth{R^{t-1}
        -
        \overline{\M}^{t-1}
        }}^2
        +
        \norm{
        \beta_{t} 
        \pth{\overline{\M}^{t-1}
        -
        \nabla \byzobj (\bfthe^{t-1})
        }}^2 
        +
        \norm{
        (1 - \beta_{t}) \alpha_{t}
        \pth{\overline{\m}^{t-1}
        -
        \nabla \byzobj (\bfthe^{t-1})
        }}^2 \\
        &~~~
        +
        \norm{(1 - \beta_{t})(1 - \alpha_{t}) 
        \pth{\overline{\g}^t - \nabla \byzobj (\bfthe^t)}
        }^2
        +
        \norm{
        \pth{\beta_{t} + (1 - \beta_{t}) \alpha_{t}}
        \pth{
        \nabla \byzobj (\bfthe^{t-1})
        -
        \nabla \byzobj (\bfthe^{t})
        }}^2 \\
        &~~~+
        2
        \beta_{t} 
        (1 - \beta_{t}) \alpha_{t}
        \iprod{
        {\overline{\M}^{t-1}
        -
        \nabla \byzobj (\bfthe^{t-1})
        }
        }{
        {\overline{\m}^{t-1}
        -
        \nabla \byzobj (\bfthe^{t-1})
        }
        }  \\
        &~~~+
        2 
        \pth{\beta_{t} + (1 - \beta_{t}) \alpha_{t}} 
        (1 - \beta_{t}) \alpha_{t}
        \iprod{
        {\overline{\m}^{t-1}
        -
        \nabla \byzobj (\bfthe^{t-1})
        }}{
        {
        \nabla \byzobj (\bfthe^{t-1})
        -
        \nabla \byzobj (\bfthe^{t})
        }
        } \\
         &~~~+
        2
        \beta_{t} 
        (1 - \beta_{t}) \alpha_{t}
        \iprod{
        {R^{t-1}
        -
        \overline{\M}^{t-1}
        }
        }{
        {\overline{\m}^{t-1}
        -
        \nabla \byzobj (\bfthe^{t-1})
        }
        }  \\
        &~~~+
        2 
        \beta_{t} 
        (\beta_{t} + (1 - \beta_{t}) \alpha_{t})
        \iprod{
        R^{t-1}
        -
        \overline{\M}^{t-1}
        }{
        {
        \nabla \byzobj (\bfthe^{t-1})
        -
        \nabla \byzobj (\bfthe^{t})
        }
        } \\
        &~~~+
        2 
        \beta_{t}^2
        \iprod{
        R^{t-1}
        -
        \overline{\M}^{t-1}
        }{
        {
        \overline{\M}^{t-1}
        -
        \nabla \byzobj (\bfthe^{t-1})
        }
        } \\
        &~~~+
        2 \beta_{t}  
        \pth{\beta_{t} + (1 - \beta_{t}) \alpha_{t}}
        \iprod{
        {\overline{\M}^{t-1}
        -
        \nabla \byzobj (\bfthe^{t-1})
        }}{
        {
        \nabla \byzobj (\bfthe^{t-1})
        -
        \nabla \byzobj (\bfthe^{t})
        }} \\
        &~~~+
        2 
        (1 - \beta_{t}) (1 - \alpha_{t})
        \iprod{\overline{\g}^t - \nabla \byzobj (\bfthe^t)}{
        \beta_{t} R^{t-1} - \beta_{t} \nabla \byzobj(\bfthe^{t-1})
        +
        (1 - \beta_{t})\alpha_{t}
        \pth{\overline{\m}^{t-1} - \nabla \byzobj (\bfthe^{t-1})}
        } \\
        &~~~-
        2 
        (1 - \beta_{t}) (1 - \alpha_{t})
        \pth{\beta_{t} + (1 - \beta_{t}) \alpha_{t}}
        \iprod{\overline{\g}^t - \nabla \byzobj (\bfthe^t)}{
        \pth{\nabla \byzobj(\bfthe^t) - \nabla \byzobj (\bfthe^{t-1})}
        } \\
        &~~~+
        2 \iprod{\beta_{t} \X_{dp}^{t-1}}{
        \beta_{t} R^{t-1} 
        +
        (1 - \beta_{t})
        \overline{\M}^t
        - 
        \nabla \byzobj (\bfthe^{t}) 
        } \\
        &\overset{(a)}{\le}
        \norm{\beta_{t} \X_{dp}^{t-1}}^2
        +
        \norm{
        \beta_{t} 
        \pth{R^{t-1}
        -
        \overline{\M}^{t-1}
        }}^2
        +
        \norm{
        \beta_{t} 
        \pth{\overline{\M}^{t-1}
        -
        \nabla \byzobj (\bfthe^{t-1})
        }}^2
        +
        \norm{
        (1 - \beta_{t}) \alpha_{t}
        \pth{\overline{\m}^{t-1}
        -
        \nabla \byzobj (\bfthe^{t-1})
        }}^2 \\
        &~~~
        +
        \norm{(1 - \beta_{t})(1 - \alpha_{t}) 
        \pth{\overline{\g}^t - \nabla \byzobj (\bfthe^t)}
        }^2
        +
        \norm{
        \pth{\beta_{t} + (1 - \beta_{t}) \alpha_{t}}
        \pth{
        \nabla \byzobj (\bfthe^{t-1})
        -
        \nabla \byzobj (\bfthe^{t})
        }}^2 \\
        &~~~+
        2
        \beta_{t} 
        (1 - \beta_{t}) \alpha_{t}
        \norm{
        {\overline{\M}^{t-1}
        -
        \nabla \byzobj (\bfthe^{t-1})
        }
        }
        \norm{
        {\overline{\m}^{t-1}
        -
        \nabla \byzobj (\bfthe^{t-1})
        }
        }  \\
        &~~~+
        2 
        \pth{\beta_{t} + (1 - \beta_{t}) \alpha_{t}} 
        (1 - \beta_{t}) \alpha_{t}
        \norm{
        {\overline{\m}^{t-1}
        -
        \nabla \byzobj (\bfthe^{t-1})
        }}
        \norm{
        {
        \nabla \byzobj (\bfthe^{t-1})
        -
        \nabla \byzobj (\bfthe^{t})
        }
        } \\
         &~~~+
        2
        \beta_{t} 
        (1 - \beta_{t}) \alpha_{t}
        \norm{
        {R^{t-1}
        -
        \overline{\M}^{t-1}
        }
        }
        \norm{
        {\overline{\m}^{t-1}
        -
        \nabla \byzobj (\bfthe^{t-1})
        }
        }  \\
        &~~~+
        2 
        \beta_{t} 
        (\beta_{t} + (1 - \beta_{t}) \alpha_{t})
        \norm{
        R^{t-1}
        -
        \overline{\M}^{t-1}
        }
        \norm{
        {
        \nabla \byzobj (\bfthe^{t-1})
        -
        \nabla \byzobj (\bfthe^{t})
        }
        } \\
        &~~~+
        2 
        \beta_{t}^2
        \norm{
        R^{t-1}
        -
        \overline{\M}^{t-1}
        }
        \norm{
        \overline{\M}^{t-1}
        -
        \nabla \byzobj (\bfthe^{t-1})
        } \\
        &~~~+
        2 \beta_{t}  
        \pth{\beta_{t} + (1 - \beta_{t}) \alpha_{t}}
        \norm{\overline{\M}^{t-1}
        -
        \nabla \byzobj (\bfthe^{t-1})}
        \norm{
        {
        \nabla \byzobj (\bfthe^{t-1})
        -
        \nabla \byzobj (\bfthe^{t})
        }} \\
        &~~~+
        2 
        (1 - \beta_{t}) (1 - \alpha_{t})
        \iprod{\overline{\g}^t - \nabla \byzobj (\bfthe^t)}{
        \beta_{t} R^{t-1} - \beta_{t} \nabla \byzobj(\bfthe^{t-1})
        +
        (1 - \beta_{t})\alpha_{t}
        \pth{\overline{\m}^{t-1} - \nabla \byzobj (\bfthe^{t-1})}
        } \\
        &~~~-
        2 
        (1 - \beta_{t}) (1 - \alpha_{t})
        \pth{\beta_{t} + (1 - \beta_{t}) \alpha_{t}}
        \iprod{\overline{\g}^t - \nabla \byzobj (\bfthe^t)}{
        \pth{\nabla \byzobj(\bfthe^t) - \nabla \byzobj (\bfthe^{t-1})}
        } \\
        &~~~+
        2 \iprod{\beta_{t} \X_{dp}^{t-1}}{\beta_{t} R^{t-1}
        +
        (1 - \beta_{t}) \overline{\m}^t - \nabla \byzobj (\bfthe^{t})}
        ,
    \end{align*}
    where inequality $(a)$ follows from the Cauchy-Schwarz inequality.
    By the definition of a robust aggregator, it holds that
     \begin{align}
        \notag
        \norm{R^{t-1} - \overline{\M}^{t-1}}
        &=
        \sqrt{
        \norm{R^{t-1} - \overline{\M}^{t-1}}^2
        } 
        \le
        \sqrt{
        \frac{\kappa}{\abth{\calG}}
        \sum_{i \in \calG}
        \norm{\M_i^{t-1} - \overline{\M}^{t-1}}^2
        } \\
        &\overset{\eqref{eq: local global momentum mixing}}{\le}
        (1 - \beta_{t-1})
        \sqrt{
        \frac{\kappa}{\abth{\calG}}
        \sum_{i \in \calG}
        \norm{\m_i^{t-1} - \overline{\m}^{t-1}}^2
        } 
        .
        \label{eq:robust agg expansion}
    \end{align}
    Let $(A)$ define the terms that include $\|R^{t-1} - \overline{\M}^{t-1}\|_2$.
    \begin{equation*}
    \hspace{-1cm}
    \begin{aligned}
        &(A) \triangleq \norm{
        \beta_{t}
        \pth{R^{t-1}
        -
        \overline{\M}^{t-1}
        }}^2 
        +
        2
        \beta_{t} 
        (1 - \beta_{t}) \alpha_{t}
        \norm{
        {R^{t-1}
        -
        \overline{\M}^{t-1}
        }
        }
        \norm{
        {\overline{\m}^{t-1}
        -
        \nabla \byzobj (\bfthe^{t-1})
        }
        } \\
        &~~~+
        2 
        \beta_{t} 
        (\beta_{t} + (1 - \beta_{t}) \alpha_{t})
        \norm{
        R^{t-1}
        -
        \overline{\M}^{t-1}
        }
        \norm{
        {
        \nabla \byzobj (\bfthe^{t-1})
        -
        \nabla \byzobj (\bfthe^{t})
        }
        } 
        +
        2 
        \beta_{t}^2
        \norm{
        R^{t-1}
        -
        \overline{\M}^{t-1}
        }
        \norm{
        \overline{\M}^{t-1}
        -
        \nabla \byzobj (\bfthe^{t-1})
        } \\
        &\overset{(b)}{\le}
        2
        \beta_{t}
        (1 - \beta_{t})
        (1 - \beta_{t-1})
        \alpha_{t}
        \sqrt{
        \frac{\kappa}{\abth{\calG}}
        \sum_{i \in \calG}
        \norm{\m_i^{t-1} - \overline{\m}^{t-1}}^2
        }
        \norm{
        {\overline{\m}^{t-1}
        -
        \nabla \byzobj (\bfthe^{t-1})
        }
        }  \\
        &~~~+
        2 \eta_{t-1} L
        \beta_{t} (1 - \beta_{t})
        (\beta_{t} + (1 - \beta_{t}) \alpha_{t})
        \sqrt{
        \frac{\kappa}{\abth{\calG}}
        \sum_{i \in \calG}
        \norm{\m_i^{t-1} - \overline{\m}^{t-1}}^2
        }
        \norm{
        {
        R^{t-1} + X_{dp}^{t-1}
        }
        } \\
        &~~~+
        2 
        \beta_{t}^2 (1 - \beta_{t-1})
        \sqrt{
        \frac{\kappa}{\abth{\calG}}
        \sum_{i \in \calG}
        \norm{\m_i^{t-1} - \overline{\m}^{t-1}}^2
        }
        \norm{
        \overline{\M}^{t-1}
        -
        \nabla \byzobj (\bfthe^{t-1})
        } 
        +
        \beta_{t}^2 (1 - \beta_{t-1})^2
        \frac{\kappa}{\abth{\calG}}
        \sum_{i \in \calG}
        \norm{\m_i^{t-1} - \overline{\m}^{t-1}}^2
        \\
        &\overset{(c)}{\le}
        \beta_{t}^2 (1 - \beta_{t-1})^2
        \frac{\kappa}{\abth{\calG}}
        \sum_{i \in \calG}
        \norm{\m_i^{t-1} - \overline{\m}^{t-1}}^2
        +
        \beta_{t}
        (1 - \beta_{t})
        (1 - \beta_{t-1}) 
        \alpha_{t}
        \frac{\kappa}{\abth{\calG}}
        \sum_{i \in \calG}
        \norm{\m_i^{t-1} - \overline{\m}^{t-1}}^2 \\
        &~~~+
        \beta_{t}
        (1 - \beta_{t})
        (1 - \beta_{t-1}) 
        \alpha_{t}
        \norm{
        {\overline{\m}^{t-1}
        -
        \nabla \byzobj (\bfthe^{t-1})
        }
        }^2  \\
        &~~~+
        \eta_{t-1} L
        \beta_{t} (1 - \beta_{t})
        (\beta_{t} + (1 - \beta_{t}) \alpha_{t})
        \frac{\kappa}{\abth{\calG}}
        \sum_{i \in \calG}
        \norm{\m_i^{t-1} - \overline{\m}^{t-1}}^2
        +
        \eta_{t-1} L
        \beta_{t} (1 - \beta_{t})
        (\beta_{t} + (1 - \beta_{t}) \alpha_{t})
        \norm{
        {
        R^{t-1} + X_{dp}^{t-1}
        }
        }^2\\
        &~~~+
        2 k_2 \beta_{t}^2 (1 - \beta_{t-1})
        \frac{\kappa}{\abth{\calG}}
        \sum_{i \in \calG}
        \norm{\m_i^{t-1} - \overline{\m}^{t-1}}^2
        +
        \frac{\beta_{t}^2 }{2 k_2}
        (1 - \beta_{t-1})
        \norm{
        \overline{\M}^{t-1}
        -
        \nabla \byzobj (\bfthe^{t-1})
        }^2 \\
        &=
        \beta_{t} (1 - \beta_{t-1})
        \pth{2 k_2 \beta_{t}
        + (1 - \beta_{t})  \alpha_{t} 
        + \frac{\eta_{t-1} L (1 - \beta_t)}{1 - \beta_{t-1}}
        (\beta_{t} + (1-\beta_{t})\alpha_{t})}
        \frac{\kappa}{\abth{\calG}}
        \sum_{i \in \calG}
        \norm{\m_i^{t-1} - \overline{\m}^{t-1}}^2 \\
        &~~~+
        \beta_{t} 
        (1 - \beta_{t}) (1 - \beta_{t-1})
        \alpha_{t}
        \norm{
        {\overline{\m}^{t-1}
        -
        \nabla \byzobj (\bfthe^{t-1})
        }
        }^2
        +
        \frac{\beta_{t}^2}{2 k_2} (1 - \beta_{t-1})
        \norm{
        \overline{\M}^{t-1}
        -
        \nabla \byzobj (\bfthe^{t-1})
        }^2 \\
        &~~~+
        \eta_{t-1} L
        \beta_{t} (1 - \beta_{t})
        (\beta_{t} + (1 - \beta_{t}) \alpha_{t})
        \norm{
        {
        R^{t-1} + X_{dp}^{t-1}
        }
        }^2,
    \end{aligned}
    \end{equation*}
    where inequality $(b)$ follows from~\eqref{eq:robust agg expansion},
    and inequality $(c)$ follows from Young's inequality.
    Let $(B)$ define the terms that include $\|\overline{\m}^{t-1} - \nabla \byzobj(\bfthe^{t-1})\|_2$.
    \begin{equation*}
    \hspace{-1cm}
    \begin{aligned}
        &(B)  \triangleq
        2
        \beta_{t} 
        (1 - \beta_{t}) \alpha_{t}
        \norm{
        {\overline{\M}^{t-1}
        -
        \nabla \byzobj (\bfthe^{t-1})
        }
        }
        \norm{
        {\overline{\m}^{t-1}
        -
        \nabla \byzobj (\bfthe^{t-1})
        }
        }  \\
        &~~~+
        2 
        \pth{\beta_{t} + (1 - \beta_{t}) \alpha_{t}} 
        (1 - \beta_{t}) \alpha_{t}
        \norm{
        {\overline{\m}^{t-1}
        -
        \nabla \byzobj (\bfthe^{t-1})
        }}
        \norm{
        {
        \nabla \byzobj (\bfthe^{t-1})
        -
        \nabla \byzobj (\bfthe^{t})
        }
        } \\
        &~~~+
        2 \beta_{t}  
        \pth{\beta_{t} + (1 - \beta_{t}) \alpha_{t}}
        \norm{\overline{\M}^{t-1}
        -
        \nabla \byzobj (\bfthe^{t-1})}
        \norm{
        {
        \nabla \byzobj (\bfthe^{t-1})
        -
        \nabla \byzobj (\bfthe^{t})
        }} \\
        &\overset{(d)}{\le}
        2
        \beta_{t}
        (1 - \beta_{t}) \alpha_{t}
        \norm{
        {\overline{\M}^{t-1}
        -
        \nabla \byzobj (\bfthe^{t-1})
        }
        }
        \norm{
        {\overline{\m}^{t-1}
        -
        \nabla \byzobj (\bfthe^{t-1})
        }
        }  \\
        &~~~+
        2 \eta_{t-1} L
        \pth{\beta_{t} + (1 - \beta_{t}) \alpha_{t}} 
        (1 - \beta_{t}) \alpha_{t}
        \norm{
        {\overline{\m}^{t-1}
        -
        \nabla \byzobj (\bfthe^{t-1})
        }}
        \norm{
        R^{t-1} + \X_{dp}^{t-1}
        } \\
        &~~~+
        2 \eta_{t-1} L
        \beta_{t}
        \pth{\beta_{t} + (1 - \beta_{t}) \alpha_{t}}
        \norm{\overline{\M}^{t-1}
        -
        \nabla \byzobj (\bfthe^{t-1})}
        \norm{
        {
        R^{t-1} + \X_{dp}^{t-1}
        }} \\
        &\overset{(e)}{\le}
        \beta_{t}
        (1 - \beta_{t}) \frac{\alpha_{t}}{2 k_2}
        \norm{\overline{\M}^{t-1}
        -
        \nabla \byzobj (\bfthe^{t-1})
        }^2
        +
        2 k_2 \beta_{t} 
        (1 - \beta_{t}) \alpha_{t}
        \norm{\overline{\m}^{t-1}
        -
        \nabla \byzobj (\bfthe^{t-1})
        }^2  \\
        &~~~+
        \eta_{t-1} L
        \pth{\beta_{t} + (1 - \beta_{t}) \alpha_{t}} 
        (1 - \beta_{t}) \alpha_{t}
        \norm{
        \overline{\m}^{t-1}
        -
        \nabla \byzobj (\bfthe^{t-1})
        }^2
        +
        \eta_{t-1} L
        \pth{\beta_{t} + (1 - \beta_{t}) \alpha_{t}} 
        (1 - \beta_{t}) \alpha_{t}
        \norm{
        R^{t-1} + \X_{dp}^{t-1}}^2 \\
        &~~~+
        \eta_{t-1} L
        \beta_{t}  
        \pth{\beta_{t} + (1 - \beta_{t}) \alpha_{t}}
        \norm{\overline{\M}^{t-1}
        -
        \nabla \byzobj (\bfthe^{t-1})}^2
        +
        \eta_{t-1} L
        \beta_{t}
        \pth{\beta_{t} + (1 - \beta_{t}) \alpha_{t}}
        \norm{
        R^{t-1} + \X_{dp}^{t-1}}^2 \\
        &=
        \beta_{t} 
        \pth{
        (1 - \beta_{t}) \frac{\alpha_{t}}{2 k_2}
        +
        \eta_{t-1} L
        (\beta_{t} + (1 - \beta_{t}) \alpha_{t})
        }
        \norm{\overline{\M}^{t-1} - \nabla \byzobj(\bfthe^{t-1})}^2 \\
        &~~~+
        (1 - \beta_{t}) \alpha_{t}
        \pth{
        2 k_2 \beta_{t} 
        +
        \eta_{t-1} L (\beta_{t} + (1 - \beta_{t}) \alpha_{t})
        }
       \norm{
        \overline{\m}^{t-1}
        -
        \nabla \byzobj (\bfthe^{t-1})
        }^2 
        +
        \eta_{t-1} L
        \pth{\beta_{t} + (1 - \beta_{t}) \alpha_{t}}^2
        \norm{
        R^{t-1} + \X_{dp}^{t-1}}^2
        ,
    \end{aligned}
    \end{equation*}
    where inequality $(d)$ follows from~\prettyref{ass: smmothness}, and inequality $(e)$ follows from Young's inequality and $k_2$ is an absolute constant.
    Putting them together, it follows, by taking expectation conditional on $\calF^t$, that
    \begin{equation*}
    \hspace{-1.2cm}
    \begin{aligned}
        &\expect{\norm{\overline{\M}^t - \nabla \byzobj (\bfthe^{t})}^2 \mid \calF^t}
        \le
         \beta_{t}
        \pth{
        \beta_{t}  +
        \frac{(1 - \beta_{t}) \alpha_{t} 
        + 
        (1 - \beta_{t-1})
        \alpha_{t}}{2 k_2}
        +
        \eta_{t-1} L
        (\beta_{t} + (1 - \beta_{t}) \alpha_{t})
        }
        \norm{\overline{\M}^{t-1} - \nabla \byzobj(\bfthe^{t-1})}^2 \\
        &~~~+
        (1 - \beta_{t}) \alpha_{t}
        \qth{\pth{(1 - \beta_{t}) \alpha_{t}
        +
        2 k_2 \beta_{t}}
        \pth{
         1
        +
        \eta_{t-1} L
        } + (1 - \beta_{t}) \beta_{t}}
       \norm{
        \overline{\m}^{t-1}
        -
        \nabla \byzobj (\bfthe^{t-1})
        }^2 
        \\
        &~~~
        +
        (1 - \beta_{t})^2
        (1 - \alpha_{t})^2 
        \frac{\sigma^2}{\abth{\calG}} 
        +
        \beta_t^2 \expect{\norm{\X_{dp,t}^{t-1}}^2 \mid \calF^{t-1}}\\
        &~~~+
        \eta_{t-1} L
        \pth{\beta_{t} + (1 - \beta_{t}) \alpha_{t}}
        \pth{\pth{\beta_{t} + (1 - \beta_{t}) \alpha_{t}}
        \pth{1 +\eta_{t-1} L} + \beta_{t} (1 - \beta_{t})}
        \expect{\norm{R^{t-1} + \X_{dp,t}^{t-1}}^2 \mid \calF^t} \\
        &~~~+
        \beta_{t} (1 - \beta_{t-1})
        \pth{2 k_2 \beta_{t} 
        + (1 - \beta_{t}) \alpha_{t} + \frac{\eta_{t-1} L (1 - \beta_{t})}{1 - \beta_{t-1}} (\beta_{t} + (1-\beta_{t})
        \alpha_{t})}
        \frac{\kappa}{\abth{\calG}}
        \sum_{i \in \calG}
        \norm{\m_i^{t-1} - \overline{\m}^{t-1}}^2 \\
        &~~~+
        2 \iprod{\beta_{t} \X_{dp,t}^{t-1}}{\beta_{t} R^{t-1}
        +
        (1 - \beta_{t}) \expect{\overline{\m}^t \mid \calF^t} - \nabla \byzobj (\bfthe^{t})}.
    \end{aligned}
    \end{equation*}
    Taking expectation conditional on $\calF^{t-1}$, it holds that
    \begin{align}
        \notag
        \expect{\norm{R^{t-1}
        +
        \X_{dp}^{t-1}
        }^2 \mid \calF^{t-1}}
        &\overset{(f)}{=}
        \expect{\norm{R^{t-1}}^2 \mid \calF^{t-1}}
        +
        \expect{
        \norm{
        \X_{dp,t}^{t-1}
        }^2 \mid \calF^{t-1}}
        +
        2 
        \iprod{
        \expect{
        R^{t-1}
        \mid \calF^{t-1}}
        }{
        \expect{
        \X_{dp,t}^{t-1}
        \mid \calF^{t-1}}
        } \\
        \label{eq: t-1 filtration}
        &\overset{(g)}{=}
        \expect{\norm{
        R^{t-1}
        }^2 \mid \calF^{t-1}}
        +
        d \sigma^2_{dp,t-1}
        ,
    \end{align}
    where equality $(f)$ follows from independent DP noise, 
    and equality $(g)$ follows from zero-mean DP noise.
    It remains to bound $\| R^{t-1} \|^2$
    \begin{align}
        \notag
        &\norm{R^{t-1} }^2
        =
        \norm{
        R^{t-1} 
        - \overline{\M}^{t-1}
        + \overline{\M}^{t-1}
        - \nabla \byzobj(\bfthe^{t-1})
        + \nabla \byzobj(\bfthe^{t-1})
        }^2 \\ 
        \notag
        &\overset{(g)}{\le}
        3 \norm{
        R^{t-1} 
        - \overline{\M}^{t-1}}^2
        +
        3 \norm{
        \overline{\M}^{t-1}
        - \nabla \byzobj(\bfthe^{t-1})
        }^2 
        +
        3 \norm{\nabla \byzobj(\bfthe^{t-1})}^2
        \\
        \notag
        &\overset{(h)}{\le}
        \frac{3 \kappa}{\abth{\calG}} 
        \sum_{i \in \calG}
        \norm{
        \M_i^{t-1} 
        - \overline{\M}^{t-1}}^2
        +
        3 \norm{
        \overline{\M}^{t-1}
        - \nabla \byzobj(\bfthe^{t-1})
        }^2 
        +
        3 \norm{\nabla \byzobj(\bfthe^{t-1})}^2 \\
        \label{eq: t-1 filtration R expansion}
        &\overset{(i)}{\le}
        \frac{3 (1 - \beta_{t-1})^2 \kappa}{\abth{\calG}} 
        \sum_{i \in \calG}
        \norm{
        \m_i^{t-1} 
        - \overline{\m}^{t-1}}^2
        +
        3 \norm{
        \overline{\M}^{t-1}
        - \nabla \byzobj(\bfthe^{t-1})
        }^2 
        +
        3 \norm{\nabla \byzobj(\bfthe^{t-1})}^2
        ,
    \end{align}
    where inequality $(g)$ follows from Jensen's inequality, 
    inequality $(h)$ follows from~\prettyref{def: robust agg},
    and inequality $(i)$ follows from~\eqref{eq:robust agg expansion}.
    Therefore, it further holds, by putting the above back, that
    \begin{equation*}
    \hspace{-1cm}
    \begin{aligned}
        &\expect{\norm{\overline{\M}^t - \nabla \byzobj (\bfthe^{t})}^2 \mid \calF^{t-1}}
        \le
        (1 - \beta_{t})^2
        (1 - \alpha_{t})^2 
        \frac{\sigma^2}{\abth{\calG}} 
        \\
        &+
        \beta_{t} \pth{
        \beta_{t}  +
        \frac{(1 - \beta_{t})\alpha_{t} +
        (1 - \beta_{t-1}) \alpha_{t}}{2 k_2}
        +
        \eta_{t-1} L
        (\beta_{t} + (1 - \beta_{t}) \alpha_{t})
        }
        \expect{\norm{\overline{\M}^{t-1} - \nabla \byzobj(\bfthe^{t-1})}^2  \mid \calF^{t-1}}\\
        &+ 3 \eta_{t-1} L
        \pth{\beta_{t} + (1 - \beta_{t}) \alpha_{t}}
        \pth{\pth{\beta_{t} + (1 - \beta_{t}) \alpha_{t}}
        \pth{1 +\eta_{t-1} L} + \beta_{t} (1 - \beta_{t})}
        \expect{\norm{\overline{\M}^{t-1} - \nabla \byzobj(\bfthe^{t-1})}^2  \mid \calF^{t-1}} \\
        &+
        (1 - \beta_{t}) \alpha_{t}
        \qth{\pth{(1 - \beta_{t}) \alpha_{t}
        +
        2 k_2 \beta_{t}}
        \pth{
         1
        +
        \eta_{t-1} L
        } + (1 - \beta_{t}) \beta_{t}}
       \expect{\norm{
        \overline{\m}^{t-1}
        -
        \nabla \byzobj (\bfthe^{t-1})
        }^2 \mid \calF^{t-1}}
        \\
        &+
        3 \eta_{t-1} L
        \pth{\beta_{t} + (1 - \beta_{t}) \alpha_{t}}
        \pth{\pth{\beta_{t} + (1 - \beta_{t}) \alpha_{t}}
        \pth{1 +\eta_{t-1} L} + \beta_{t} (1 - \beta_{t})}
        \norm{\nabla \byzobj (\bfthe^{t-1})}^2 \\
        &+
        \beta_{t}
        (1 - \beta_{t-1})
        \pth{2 k_2 \beta_{t} + (1 - \beta_{t}) \alpha_{t} + \frac{\eta_{t-1} L(1 - \beta_{t})}{1 - \beta_{t-1}} (\beta_{t} + (1-\beta_{t})\alpha_{t})}
        \frac{\kappa}{\abth{\calG}}
        \sum_{i \in \calG}
        \expect{\norm{\m_i^{t-1} - \overline{\m}^{t-1}}^2 \mid \calF^{t-1}} \\
        &+
        3 (1 - \beta_{t-1})^2
        \eta_{t-1} L
        \pth{\beta_{t} + (1 - \beta_{t}) \alpha_{t}}
        \pth{\pth{\beta_{t} + (1 - \beta_{t}) \alpha_{t}}
        \pth{1 +\eta_{t-1} L} + \beta_{t} (1 - \beta_{t})}
        \frac{\kappa}{\abth{\calG}}
        \sum_{i \in \calG}
        \expect{\norm{\m_i^{t-1} - \overline{\m}^{t-1}}^2 \mid \calF^{t-1}} \\
        &+
        \pth{\beta_t^2
        +
        \eta_{t-1} L
        \pth{\beta_{t} + (1 - \beta_{t}) \alpha_{t}}
        \pth{\pth{\beta_{t} + (1 - \beta_{t}) \alpha_{t}}
        \pth{1 +\eta_{t-1} L} + \beta_{t} (1 - \beta_{t})}}
        d \sigma_{dp,t-1}^2
        .
    \end{aligned}
    \end{equation*}

\end{proof}

\subsubsection{Proof of Client-side Momentum Estimation Error}

\begin{lemma}
    \label{lmm: true momentum decay v2}
    Under Assumptions~\ref{ass: smmothness}, and~\ref{ass: bounded variance client-wise}, 
    it holds that
    \begin{align*}
        &\expect{\norm{\overline{\m}^t - \nabla \byzobj (\bfthe^t)}^2 \mid \calF^{t-1}}
        \le
        \alpha_{t}^2 (1  + \eta_{t-1} L) 
        \expect{\norm{\overline{\m}^{t-1} - \nabla \byzobj (\bfthe^{t-1})}^2 \mid \calF^{t-1}}
        +
        (1 - \alpha_{t})^2 \frac{\sigma^2}{\abth{\calG}}
        \\
        &~~~+
        3 \alpha_{t}^2 
        \eta_{t-1} L (1  + \eta_{t-1} L)
        \norm{\nabla \byzobj (\bfthe^{t-1})}^2 
        +
        3 \alpha_{t}^2 \eta_{t-1} L (1 + \eta_{t-1} L)
        \expect{\norm{\overline{\m}^{t-1} - \nabla \byzobj (\bfthe^{t-1})}^2 \mid \calF^{t-1}}
        \\
        &~~~+
        3 \alpha_{t}^2 
        \eta_{t-1} L (1  + \eta_{t-1} L)
        \frac{(1 - \beta_{t-1})^2 \kappa}{\abth{\calG}}
        \sum_{i \in \calG}
        \expect{\norm{\m_i^{t-1} - \overline{\m}^{t-1}}^2 \mid \calF^{t-1}}
        +
        \alpha_{t}^2 
        \eta_{t-1} L (1  + \eta_{t-1} L)
        d \sigma_{dp,t-1}^2  
        .
    \end{align*}
\end{lemma}

\begin{proof}[\bf Proof of~\prettyref{lmm: true momentum decay v2}]
    \begin{align*}
        \overline{\m}^t - \nabla \byzobj (\bfthe^t)
        &=
        \alpha_{t} \overline{\m}^{t-1}
        +
        (1 - \alpha_{t}) \overline{\g}^t
        -
        \nabla \byzobj (\bfthe^t) \\
        &=
        \alpha_{t} \pth{\overline{\m}^{t-1} - \nabla \byzobj (\bfthe^{t-1})}
        +
        (1 - \alpha_{t}) 
        \pth{\overline{\g}^t - \nabla \byzobj (\bfthe^{t})} 
        -
        \alpha_{t}
        \pth{\nabla \byzobj (\bfthe^{t}) - \nabla \byzobj (\bfthe^{t-1})}
        .
    \end{align*}
    It follows that
    \begin{align*}
        \norm{\overline{\m}^t - \nabla \byzobj (\bfthe^t)}^2
        &=
        (1 - \alpha_{t})^2
        \norm{\overline{\g}^t - \nabla \byzobj (\bfthe^{t})}^2
        +
        \alpha_{t}^2
        \norm{
        \pth{{\overline{\m}^{t-1} - \nabla \byzobj (\bfthe^{t-1})}}
        -
        \pth{\nabla \byzobj (\bfthe^{t}) - \nabla \byzobj (\bfthe^{t-1})}}^2 \\
        &~~~+
        2 (1 - \alpha_{t}) \alpha_{t}
        \iprod{\overline{\g}^t - \nabla \byzobj (\bfthe^{t})}{
        \pth{{\overline{\m}^{t-1} - \nabla \byzobj (\bfthe^{t-1})}}
        -
        \pth{\nabla \byzobj (\bfthe^{t}) - \nabla \byzobj (\bfthe^{t-1})}}.
    \end{align*}
    Taking expectation conditional on $\calF^t$, it holds that
    \begin{align*}
        &\expect{\norm{\overline{\m}^t - \nabla \byzobj (\bfthe^t)}^2 \mid \calF^t}
        =
        (1 - \alpha_{t})^2
        \expect{\norm{\overline{\g}^t - \nabla \byzobj (\bfthe^{t})}^2 \mid \calF^t} \\
        &~~~+
        \alpha_{t}^2
        \norm{
        \pth{{\overline{\m}^{t-1} - \nabla \byzobj (\bfthe^{t-1})}}
        -
        \pth{\nabla \byzobj (\bfthe^{t}) - \nabla \byzobj (\bfthe^{t-1})}}^2 \\
        &~~~+
        2 (1 - \alpha_{t}) \alpha_{t}
        \iprod{\expect{\overline{\g}^t - \nabla \byzobj (\bfthe^{t}) \mid \calF^t}}{
        \pth{{\overline{\m}^{t-1} - \nabla \byzobj (\bfthe^{t-1})}}
        -
        \pth{\nabla \byzobj (\bfthe^{t}) - \nabla \byzobj (\bfthe^{t-1})}} \\
        &=
        \alpha_{t}^2
        \norm{
        \pth{{\overline{\m}^{t-1} - \nabla \byzobj (\bfthe^{t-1})}}
        -
        \pth{\nabla \byzobj (\bfthe^{t}) - \nabla \byzobj (\bfthe^{t-1})}}^2
        +
        (1 - \alpha_{t})^2 \frac{\sigma^2}{\abth{\calG}} \\
        &\overset{(a)}{\le}
        \alpha_{t}^2 (1  + \eta_{t-1} L)
        \norm{\overline{\m}^{t-1} - \nabla \byzobj (\bfthe^{t-1})}^2
        +
        \alpha_{t}^2 (1  + \frac{1}{\eta_{t-1} L})
        \norm{\nabla \byzobj (\bfthe^{t}) - \nabla \byzobj (\bfthe^{t-1})}^2
        +
        (1 - \alpha_{t})^2 \frac{\sigma^2}{\abth{\calG}} \\
        &\overset{(b)}{\le}
        \alpha_{t}^2 (1  + \eta_{t-1} L)
        \norm{\overline{\m}^{t-1} - \nabla \byzobj (\bfthe^{t-1})}^2
        +
        \alpha_{t}^2 
        \eta_{t-1} L (1  + \eta_{t-1} L)
        \norm{R^{t-1} + \X_{dp}^{t-1}}^2 
        +
        (1 - \alpha_{t})^2 \frac{\sigma^2}{\abth{\calG}}
        ,
    \end{align*}
    where inequality $(a)$ follows from Young's inequality, 
    and inequality $(b)$ follows from~\prettyref{ass: smmothness}.
    Plugging~\eqref{eq: t-1 filtration} and~\eqref{eq: t-1 filtration R expansion}, it holds, by taking expectation conditional on $\calF^{t-1}$, that
    \begin{align*}
        &\expect{\norm{\overline{\m}^t - \nabla \byzobj (\bfthe^t)}^2 \mid \calF^{t-1}}
        \overset{(c)}{\le}
        \alpha_{t}^2 (1  + \eta_{t-1} L) 
        \expect{\norm{\overline{\m}^{t-1} - \nabla \byzobj (\bfthe^{t-1})}^2 \mid \calF^{t-1}}
        +
        \alpha_{t}^2 
        \eta_{t-1} L (1  + \eta_{t-1} L)
        d \sigma_{dp,t-1}^2  \\
        &~~~+
        3 \alpha_{t}^2 
        \eta_{t-1} L (1  + \eta_{t-1} L)
        \norm{\nabla \byzobj (\bfthe^{t-1})}^2 
        +
        3 \alpha_{t}^2 \eta_{t-1} L (1 + \eta_{t-1} L)
        \expect{\norm{\overline{\m}^{t-1} - \nabla \byzobj (\bfthe^{t-1})}^2 \mid \calF^{t-1}}
        \\
        &~~~+
        3 \alpha_{t}^2 
        \eta_{t-1} L (1  + \eta_{t-1} L)
        \frac{(1 - \beta_{t-1})^2 \kappa}{\abth{\calG}}
        \sum_{i \in \calG}
        \expect{\norm{\m_i^{t-1} - \overline{\m}^{t-1}}^2 \mid \calF^{t-1}}
        +
        (1 - \alpha_{t})^2 \frac{\sigma^2}{\abth{\calG}}
        ,
    \end{align*}
    where inequality $(c)$ follows from zero-mean and independent DP noise $\X_{dp}^{t-1}$.
\end{proof}

\subsection{Strongly-convex objectives}
\label{app: stochastic convergence strongly-convex v2}
\subsubsection{Proof of Local Momentum Drift}
\begin{lemma}
    \label{lmm: sc momentum drift v2 sc}
    Under Assumptions~\ref{ass: smmothness},~\ref{ass: bounded variance client-wise} and~\ref{ass: bounded similarity}.
    When $\byzobj$ is strongly convex,
    for any $\alpha_t, \beta_t \in (0,1)$, it holds that
    \begin{align*}
        &\frac{1}{\abth{\calG}}
        \sum_{i \in \calG}
        \expect{
        \norm{\m_i^t - \overline{\m}^t}^2 \big| \calF^{t-1}} 
        \le
        2 (1 - \alpha_t) B^2 \eta_{t-1}^2 L^2
        d \sigma_{dp,t-1}^2
        +
        2(1 - \alpha_t)^2
        \pth{1 + \frac{1}{\abth{\calG}}}
        \sigma^2 \\
        &~~~+\pth{
        \alpha_t + 6 \kappa (1 - \alpha_t) (1 - \beta_{t-1})^2 B^2 \eta_{t-1}^2 L^2
        }
       \frac{1}{\abth{\calG}}
        \sum_{i \in \calG}
        \expect{\norm{\m_i^{t-1} - \overline{\m}^{t-1}}^2 \big| \calF^{t-1}}
        +
        (1 - \alpha_t) G^2
        \\
        &~~~+
        2 (1 - \alpha_t) B^2
        \pth{1 + 3 \eta_{t-1}^2 L^2}
        \norm{\nabla F (\bfthe^{t-1})}^2 
        + 6 (1 - \alpha_t) B^2 \eta_{t-1}^2 L^2
        \expect{\norm{
        \overline{\M}^{t-1}
        - \nabla F(\bfthe^{t-1})
        }^2 \big|\calF^{t-1}}
        .
    \end{align*}
\end{lemma}

\begin{proof}[\bf Proof of~\prettyref{lmm: sc momentum drift v2 sc}]
    We know that
    \[
        \m_i^t = \alpha_t \m_i^{t-1} + (1 - \alpha_t) \g_i^t.
    \]
    It follows that
    \begin{align*}
        &\frac{1}{\abth{\calG}}
        \sum_{i \in \calG}
        \norm{\m_i^t - \overline{\m}^t}^2
        =
        \frac{1}{\abth{\calG}}
        \sum_{i \in \calG}
        \big \|
        \alpha_t \pth{\m_i^{t-1} - \overline{\m}^{t-1}} 
        + (1 - \alpha_t) \pth{\pth{\g_i^t - \nabla F_i (\bfthe^t)} - \pth{\overline{\g}^t - \nabla \byzobj (\bfthe^t)}} \\
        &~~~+ (1 - \alpha_t) \pth{\nabla F_i (\bfthe^t) - \nabla \byzobj (\bfthe^t)}\big \|_2^2 \\
        &=
        \frac{1}{\abth{\calG}}
        \sum_{i \in \calG}
        \norm{\alpha_t \pth{\m_i^{t-1} - \overline{\m}^{t-1}} 
        +
        (1 - \alpha_t) \pth{\nabla F_i (\bfthe^t) - \nabla \byzobj (\bfthe^t)}
        }^2 \\
        &~~~+
        \frac{1}{\abth{\calG}}
        \sum_{i \in \calG}
        \norm{(1 - \alpha_t) \pth{\pth{\g_i^t - \nabla F_i (\bfthe^t)} - \pth{\overline{\g}^t - \nabla \byzobj (\bfthe^t)}}}^2 \\
        &~~~+
        \frac{1 - \alpha_t}{\abth{\calG}}
        \sum_{i \in \calG}
        \iprod{\alpha_t \pth{\m_i^{t-1} - \overline{\m}^{t-1}} 
        +
        (1 - \alpha_t) \pth{\nabla F_i (\bfthe^t) - \nabla \byzobj (\bfthe^t)}}{\pth{\g_i^t - \nabla F_i (\bfthe^t)} - \pth{\overline{\g}^t - \nabla \byzobj (\bfthe^t)}}.
    \end{align*}
    Taking expectation conditional on $\calF^t$, it holds that
    \begin{align*}
        &\frac{1}{\abth{\calG}}
        \sum_{i \in \calG}
        \expect{\norm{\m_i^t - \overline{\m}^t}^2 \mid \calF^t}
        =
        \frac{1}{\abth{\calG}}
        \sum_{i \in \calG}
        \Expect[\big \|
        \alpha_t \pth{\m_i^{t-1} - \overline{\m}^{t-1}} 
        + (1 - \alpha_t) \pth{\pth{\g_i^t - \nabla F_i (\bfthe^t)} - \pth{\overline{\g}^t - \nabla \byzobj (\bfthe^t)}} \\
        &~~~+ (1 - \alpha_t) \pth{\nabla F_i (\bfthe^t) - \nabla \byzobj (\bfthe^t)}\big \|_2^2 \mid \calF^t]\\
        &=
        \frac{1}{\abth{\calG}}
        \sum_{i \in \calG}
        \expect{\norm{\alpha_t \pth{\m_i^{t-1} - \overline{\m}^{t-1}} 
        +
        (1 - \alpha_t) \pth{\nabla F_i (\bfthe^t) - \nabla \byzobj (\bfthe^t)}
        }^2 \mid \calF^t}\\
        &~~~+
        \frac{1}{\abth{\calG}}
        \sum_{i \in \calG}
        \expect{\norm{(1 - \alpha_t) \pth{\pth{\g_i^t - \nabla F_i (\bfthe^t)} - \pth{\overline{\g}^t - \nabla \byzobj (\bfthe^t)}}}^2 \mid \calF^t}\\
        &~~~+
        \frac{1 - \alpha_t}{\abth{\calG}}
        \sum_{i \in \calG}
        \expect{\iprod{\alpha_t \pth{\m_i^{t-1} - \overline{\m}^{t-1}} 
        +
        (1 - \alpha_t) \pth{\nabla F_i (\bfthe^t) - \nabla \byzobj (\bfthe^t)}}{\pth{\g_i^t - \nabla F_i (\bfthe^t)} - \pth{\overline{\g}^t - \nabla \byzobj (\bfthe^t)}} \mid \calF^t} \\
        &\overset{(a)}{=}
        \frac{1}{\abth{\calG}}
        \sum_{i \in \calG}
        \expect{\norm{\alpha_t \pth{\m_i^{t-1} - \overline{\m}^{t-1}} 
        +
        (1 - \alpha_t) \pth{\nabla F_i (\bfthe^t) - \nabla \byzobj (\bfthe^t)}
        }^2 \mid \calF^t}\\
        &~~~+
        \frac{1}{\abth{\calG}}
        \sum_{i \in \calG}
        \expect{\norm{(1 - \alpha_t) \pth{\pth{\g_i^t - \nabla F_i (\bfthe^t)} - \pth{\overline{\g}^t - \nabla \byzobj (\bfthe^t)}}}^2 \mid \calF^t}\\
        &~~~+
        \frac{1 - \alpha_t}{\abth{\calG}}
        \sum_{i \in \calG}
        \iprod{\alpha_t \pth{\m_i^{t-1} - \overline{\m}^{t-1}} 
        +
        (1 - \alpha_t) \pth{\nabla F_i (\bfthe^t) - \nabla \byzobj (\bfthe^t)}}{\expect{\pth{\g_i^t - \nabla F_i (\bfthe^t)} - \pth{\overline{\g}^t - \nabla \byzobj (\bfthe^t)}\mid \calF^t}}  \\
        &\overset{(b)}{\le}
        \frac{\alpha_t}{\abth{\calG}}
        \sum_{i \in \calG}
        \norm{ \pth{\m_i^{t-1} - \overline{\m}^{t-1}} }^2
        +
        \frac{1 - \alpha_t}{\abth{\calG}}
        \sum_{i \in \calG}
        \norm{\nabla F_i (\bfthe^t) - \nabla \byzobj (\bfthe^t)}^2
        +
        2(1 - \alpha_t)^2
        \pth{1 + \frac{1}{\abth{\calG}}}
        \sigma^2 \\
        &\overset{(c)}{\le}
        \frac{\alpha_t}{\abth{\calG}}
        \sum_{i \in \calG}
        \norm{ \pth{\m_i^{t-1} - \overline{\m}^{t-1}} }^2
        +
        (1 - \alpha_t) B^2
        \norm{\nabla \byzobj(\bfthe^t)}^2
        +
        (1 - \alpha_t) G^2
        +
        2(1 - \alpha_t)^2
        \pth{1 + \frac{1}{\abth{\calG}}}
        \sigma^2 
        ,
    \end{align*}
    where equality $(a)$ follows from independence, 
    inequality $(b)$ follows from Jensen's inequality and~\prettyref{ass: bounded variance client-wise},
    and inequality $(c)$ follows from~\prettyref{ass: bounded similarity}.

    It follows that
    \begin{align*}
        (1 - \alpha_t) B^2
        \norm{\nabla \byzobj(\bfthe^t)}^2
        &=
        (1 - \alpha_t) B^2
        \norm{\nabla \byzobj(\bfthe^t) - \nabla \byzobj(\bfthe^{t-1}) + \nabla \byzobj(\bfthe^{t-1})}^2 \\
        &\le
        2(1 - \alpha_t) B^2
        \norm{\nabla \byzobj(\bfthe^t) - \nabla \byzobj(\bfthe^{t-1})}^2
        + 
        2(1 - \alpha_t) B^2
        \norm{\nabla \byzobj(\bfthe^{t-1})}^2 \\
        &\le
        2(1 - \alpha_t) B^2 \eta_{t-1}^2 L^2
        \norm{R^{t-1} + \X_{dp}^{t-1}}^2
        + 
        2(1 - \alpha_t) B^2
        \norm{\nabla \byzobj(\bfthe^{t-1})}^2 
        .
    \end{align*}
    We know that
        \begin{align*}
        \expect{\norm{R^{t-1}
        +
        \X_{dp}^{t-1}
        }^2 \mid \calF^{t-1}}
        &=
        \expect{\norm{R^{t-1}}^2 \mid \calF^{t-1}}
        +
        \expect{
        \norm{
        \X_{dp}^{t-1}
        }^2 \mid \calF^{t-1}}
        +
        2 
        \iprod{
        \expect{
        R^{t-1}
        \mid \calF^{t-1}}
        }{
        \expect{
        \X_{dp}^{t-1}
        \mid \calF^{t-1}}
        } \\
        &=
        \expect{\norm{
        R^{t-1}
        }^2 \mid \calF^{t-1}}
        +
        d \sigma^2_{dp}
        ,
    \end{align*}
    and
    \begin{align*}
        \norm{R^{t-1} }^2
        &=
        \norm{
        R^{t-1} 
        - \overline{\M}^{t-1}
        + \overline{\M}^{t-1}
        - \nabla \byzobj(\bfthe^{t-1})
        + \nabla \byzobj(\bfthe^{t-1})
        }^2 \\ 
        &\overset{\eqref{eq:robust agg expansion}}{\le}
        \frac{3 (1 - \beta_{t-1})^2 \kappa}{\abth{\calG}} 
        \sum_{i \in \calG}
        \norm{\m_i^{t-1} - \overline{\m}^{t-1}}^2
        +
        3 \norm{
        \overline{\M}^{t-1}
        - \nabla \byzobj(\bfthe^{t-1})
        }^2 
        +
        3 \norm{\nabla \byzobj(\bfthe^{t-1})}^2
        .
    \end{align*}
    Therefore,
    \begin{align*}
        &\expect{ \norm{R^{t-1} + \X_{dp}^{t-1}}^2 \mid \calF^{t-1}} \\
        &\le
        \frac{3 (1 - \beta_{t-1})^2 \kappa}{\abth{\calG}} 
        \sum_{i \in \calG}
        \expect{\norm{\m_i^{t-1} - \overline{\m}^{t-1}}^2 \mid \calF^{t-1}}
        +
        3 \expect{\norm{
        \overline{\M}^{t-1}
        - \nabla \byzobj(\bfthe^{t-1})
        }^2 \mid \calF^{t-1}}
        +
        3 \norm{\nabla \byzobj(\bfthe^{t-1})}^2 \\
        &~~~+ d \sigma_{dp,t-1}^2
        .
    \end{align*}
    Putting it back in, it holds that
    \begin{align*}
        &(1-\alpha_t) B^2 \expect{\norm{\nabla \byzobj (\bfthe^t)}^2 \mid \calF^{t-1}}
        \le
        2 (1 - \alpha_t) B^2 \norm{\nabla \byzobj (\bfthe^{t-1})}^2 \\
        &~~~+
        \frac{6 \kappa (1 - \alpha_t) B^2 \eta_t^2 L^2}{\abth{\calG}}
        \sum_{i \in \calG}
        \expect{\norm{\M_i^{t-1} - \overline{\M}^{t-1}}^2 \mid \calF^{t-1}}
        + 2 (1 - \alpha_t) B^2 \eta_t^2 L^2
        d \sigma_{dp,t-1}^2\\
        &~~~+ 6 (1 - \alpha_t) B^2 \eta_t^2 L^2
        \expect{\norm{
        \overline{\M}^{t-1}
        - \nabla \byzobj(\bfthe^{t-1})
        }^2 \mid \calF^{t-1}}
        + 6 (1 - \alpha_t) B^2 \eta_t^2 L^2
        \norm{\nabla \byzobj(\bfthe^{t-1})}^2  
         \\
        &\le
        2 (1 - \alpha_t) B^2 \norm{\nabla \byzobj (\bfthe^{t-1})}^2 
        +
        \frac{6 \kappa (1 - \alpha_t) (1 - \beta_{t-1})^2 B^2 \eta_t^2 L^2}{\abth{\calG}}
        \sum_{i \in \calG}
        \expect{\norm{\m_i^{t-1} - \overline{\m}^{t-1}}^2 \mid \calF^{t-1}}\\
        &~~~+ 6 (1 - \alpha_t) B^2 \eta_t^2 L^2
        \expect{\norm{
        \overline{\M}^{t-1}
        - \nabla \byzobj(\bfthe^{t-1})
        }^2 \mid \calF^{t-1}}
        + 6 (1 - \alpha_t) B^2 \eta_t^2 L^2
        \norm{\nabla \byzobj(\bfthe^{t-1})}^2  \\
        &~~~+ 2 (1 - \alpha_t) B^2 \eta_t^2 L^2
        d \sigma_{dp,t-1}^2
        .
    \end{align*}
    Combining everything together, it holds that
    \begin{align*}
        &\frac{1}{\abth{\calG}}
        \sum_{i \in \calG}
        \expect{
        \norm{\m_i^t - \overline{\m}^t}^2 \mid \calF^{t-1}} 
        \le
        2 (1 - \alpha_t) B^2 \eta_{t-1}^2 L^2
        d \sigma_{dp,t-1}^2
        +
        (1 - \alpha_t) G^2
        +
        2(1 - \alpha_t)^2
        \pth{1 + \frac{1}{\abth{\calG}}}
        \sigma^2 \\
        &~~~+\pth{
        \alpha_t + 6 \kappa (1 - \alpha_t) (1 - \beta_{t-1})^2 B^2 \eta_{t-1}^2 L^2
        }
       \frac{1}{\abth{\calG}}
        \sum_{i \in \calG}
        \expect{\norm{\m_i^{t-1} - \overline{\m}^{t-1}}^2 \mid \calF^{t-1}}
        \\
        &~~~+
        2 (1 - \alpha_t) B^2
        \pth{1 + 3 \eta_{t-1}^2 L^2}
        \norm{\nabla \byzobj (\bfthe^{t-1})}^2 
        + 6 (1 - \alpha_t) B^2 \eta_{t-1}^2 L^2
        \expect{\norm{
        \overline{\M}^{t-1}
        - \nabla \byzobj(\bfthe^{t-1})
        }^2 \mid \calF^{t-1}}
        .
    \end{align*}
\end{proof}

\subsubsection{Proof of~\prettyref{thm: sc stochastic gradient convergence v2 main text}}
\begin{theorem}[Strongly-convex, formal]
    \label{thm: sc stochastic gradient convergence v2}
    Suppose that Assumptions~\ref{ass: smmothness}, \ref{ass: bounded variance client-wise}, \ref{ass: bounded similarity}, and \ref{ass: bounded sto gradient cl} hold.
    When $\alpha_t = 1 - 4900 \eta_t L$, $\beta_t = 1 - 55 \eta_t L$,
    $\eta_t = \frac{10}{\mu (t + a \frac{L}{\mu})}$ such that $a \ge 
    49000$,
    and $B^2 \le \frac{1}{600 \kappa}$, it holds that
\begin{align*}
   \expect{\norm{\bfthe^T - \bfthe^{\ast}}^2}
    &\lesssim
     \frac{L^2 (\expect{\byzobj (\bfthe^0)} - \byzobj^{\ast})}{\mu^3 T^2} 
    +
    \frac{L \sigma^2}{\mu^3 T^2 \abth{\calG}} 
    +
    \frac{L \kappa}{\mu^3 T^2} 
  \pth{
     \pth{1 + \frac{1}{\abth{\calG}}}
    \sigma^2
    +
    G^2
    }
    \\
    &~~~+
    L (1 + \kappa)
    \pth{1 + \frac{1}{\abth{\calG}}}
    \frac{\sigma^2 }{\mu^3 T}
    +
    \pth{\frac{L }{\mu}}
    \frac{\log(1 / \delta) C^2 d}{\mu^2 \epsilon^2 m^2} \pth{1 + \frac{f}{m}}
    +
    \frac{\kappa G^2}{\mu^2}
    .
\end{align*}
\end{theorem}
\begin{proof}
        Define the Lyapunov function as 
        \begin{align*}
            W^t &\triangleq
            \expect{\byzobj (\bfthe^t) - \byzobj^{\ast} + \frac{1}{20 L} \norm{\nabla \byzobj (\bfthe^t) - \overline{\M}^t}^2 + 
            \frac{1}{15 L} \norm{\nabla \byzobj (\bfthe^t) - \overline{\m}^t}^2 +
            \frac{2 \kappa}{L} \frac{1}{\abth{\calG}}\sum_{i\in \calG} \norm{\m_i^t - \overline{\m}^t}^2};
            \\ 
            V^t &\triangleq 
            \pth{t + a \frac{L}{\mu}}^2
            W^t \\ 
            &=  
            \pth{t + a \frac{L}{\mu}}^2
            \expect{\byzobj (\bfthe^t) - \byzobj^{\ast} + \frac{1}{20 L} \norm{\nabla \byzobj (\bfthe^t) - \overline{\M}^t}^2 + \frac{1}{15 L} \norm{\nabla \byzobj (\bfthe^t) - \overline{\m}^t}^2 +
            \frac{2 \kappa}{L} \frac{1}{\abth{\calG}}\sum_{i\in \calG} \norm{\m_i^t - \overline{\m}^t}^2}.
        \end{align*}
        For ease of notation, let $\hat{t} = t +  \frac{a L}{\mu}$.
        It follows that
        \begin{align*}
            V^{t+1} - V^t 
            &=
            (\hat{t} + 1)^2 W^{t+1} - \hat{t}^2 W^{t} \\
            &=
            (\hat{t} + 1)^2 (W^{t+1} - W^{t}) + \pth{(\hat{t} + 1)^2 - \hat{t}^2} W^{t} \\
            &=
            (\hat{t} + 1)^2 (W^{t+1} - W^{t}) + \pth{2 \hat{t} + 1} W^{t}.
        \end{align*}
    Next, we have
    \begin{align*}
        \expect{\byzobj (\bfthe^{t+1}) - \byzobj (\bfthe^{t})}
        &\le
        - \eta_t
        \pth{\frac{3}{4} - 2 \eta_t L}
        \expect{\norm{\nabla \byzobj (\bfthe^{t})}^2}
        +
        \frac{2
        \kappa  
        \eta_t (1 + \eta_t L) (1 - \beta_t)^2
        }{\abth{\calG}}
        \sum_{i \in \calG}
        \expect{\norm{
        \m_i^{t}
        -
        \overline{\m}^t
        }^2}
        \\
        &~~~
        +
        2 \eta_t (1 + \eta_t L)
        \expect{
        \norm{\nabla \byzobj(\bfthe^t) - \overline{\M}^t}^2
        }
        +
        \frac{\eta_t^2 L d}{2}
        \sigma_{dp, t}^2
        ,
    \end{align*}
    It follows that
    \begin{equation*}
    \hspace{-2cm}
        \begin{aligned}
        &\frac{z_1}{L} \expect{\norm{\overline{\M}^{t+1} - \nabla \byzobj (\bfthe^{t+1})}^2 - \norm{\overline{\M}^{t} - \nabla \byzobj (\bfthe^{t})}^2\mid \calF^{t}}
        \le
        \frac{z_1}{L}(1 - \beta_{t+1})^2(1 - \alpha_{t+1})^2 \frac{\sigma^2}{\abth{\calG}} 
        \\
        &+
        \frac{z_1}{L}
        \pth{\beta_{t+1} \pth{
        \beta_{t+1}  +
        \frac{(1 - \beta_{t+1}) \alpha_{t+1} + (1 - \beta_{t}) \alpha_{t+1}}{2 k_2}
        +
        \eta_{t} L
        (\beta_{t+1} + (1 - \beta_{t+1}) \alpha_{t+1})
        } - 1}
        \expect{\norm{\overline{\M}^{t} - \nabla \byzobj(\bfthe^{t})}^2  \mid \calF^{t}}\\
        &+ 3 z_1 \eta_{t}
        \pth{\beta_{t+1} + (1 - \beta_{t+1}) \alpha_{t+1}}
        \pth{\pth{\beta_{t+1} + (1 - \beta_{t+1}) \alpha_{t+1}}
        \pth{1 +\eta_{t} L} + \beta_{t+1} (1 - \beta_{t+1})}
        \expect{\norm{\overline{\M}^{t} - \nabla \byzobj(\bfthe^{t})}^2  \mid \calF^{t}} \\
        &+ \frac{z_1}{L}
        (1 - \beta_{t+1}) \alpha_{t+1}
        \qth{\pth{(1 - \beta_{t+1}) \alpha_{t+1}
        +
        2 k_2 \beta_{t+1}}
        \pth{
        1  
        +
        \eta_{t} L
        } + (1 - \beta_{t+1}) \beta_{t+1}}
       \expect{\norm{
        \overline{\m}^{t}
        -
        \nabla \byzobj (\bfthe^{t})
        }^2 \mid \calF^{t}}
        \\
        &+
        3 z_1 \eta_{t}
        \pth{\beta_{t+1} + (1 - \beta_{t+1}) \alpha_{t+1}}
        \pth{\pth{\beta_{t+1} + (1 - \beta_{t+1}) \alpha_{t+1}}
        \pth{1 +\eta_{t} L} + \beta_{t+1} (1 - \beta_{t+1})}
        \norm{\nabla \byzobj (\bfthe^{t})}^2 \\
        &+
        \frac{z_1}{L}
        (1 - \beta_{t})
        \beta_{t+1}
        \pth{2 k_2 \beta_{t+1} + (1 - \beta_{t+1}) \alpha_{t+1}
        + \frac{\eta_{t} L (1 - \beta_{t+1})}{1 - \beta_{t}} (\beta_{t+1} + (1-\beta_{t+1})\alpha_{t+1})}
        \frac{\kappa}{\abth{\calG}}
        \sum_{i \in \calG}
        \expect{\norm{\m_i^{t} - \overline{\m}^{t}}^2 \mid \calF^{t}} \\
        &+
        3 z_1 (1 - \beta_{t})^2
        \eta_{t}
        \pth{\beta_{t+1} + (1 - \beta_{t+1}) \alpha_{t+1}}
        \pth{\pth{\beta_{t+1} + (1 - \beta_{t+1}) \alpha_{t+1}}
        \pth{1 +\eta_{t} L} + \beta_{t+1} (1 - \beta_{t+1})}
        \frac{\kappa}{\abth{\calG}}
        \sum_{i \in \calG}
        \expect{\norm{\m_i^{t} - \overline{\m}^{t}}^2 \mid \calF^{t}} \\
        &+
        \frac{z_1}{L}
        \pth{\beta_{t+1}^2
        +
        \eta_{t} L
        \pth{\beta_{t+1} + (1 - \beta_{t+1}) \alpha_{t+1}}
        \pth{\pth{\beta_{t+1} + (1 - \beta_{t+1}) \alpha_{t+1}}
        \pth{1 +\eta_{t} L} + \beta_{t+1} (1 - \beta_{t+1})}}
        d \sigma_{dp,t}^2,
        \end{aligned}
    \end{equation*}
    where the inequality follows from Lemma~\ref{lm: momentum deviation decay v2}.
    Also,
   \begin{align*}
        &\frac{z_2}{L}
        \expect{\norm{\overline{\m}^{t+1} - \nabla \byzobj (\bfthe^{t+1})}^2 - \norm{\overline{\m}^{t} - \nabla \byzobj (\bfthe^{t})}^2 \mid \calF^{t}}
        \le
        \frac{z_2}{L}
        \pth{\alpha_{t+1}^2 (1  + \eta_{t} L) - 1 }
        \expect{\norm{\overline{\m}^{t} - \nabla \byzobj (\bfthe^{t})}^2 \mid \calF^{t}}
        \\
        &~~~+
        3 z_2 \alpha_{t+1}^2 
        \eta_{t} (1  + \eta_{t} L)
        \norm{\nabla \byzobj (\bfthe^{t})}^2 
        +
        3 z_2 \alpha_{t+1}^2 \eta_{t} (1 + \eta_{t} L)
        \expect{\norm{\overline{\M}^{t} - \nabla \byzobj (\bfthe^{t})}^2 \mid \calF^{t}}
        \\
        &~~~+
        3 z_2 \alpha_{t+1}^2 
        \eta_{t}  (1  + \eta_{t} L)
        \frac{(1 - \beta_{t})^2 \kappa}{\abth{\calG}}
        \sum_{i \in \calG}
        \expect{\norm{\m_i^{t} - \overline{\m}^{t}}^2 \mid \calF^{t}}
        \\
        &~~~+ \frac{z_2}{L}
        (1 - \alpha_{t+1})^2 \frac{\sigma^2}{\abth{\calG}}
        +
        z_2
        \alpha_{t+1}^2 
        \eta_{t} (1  + \eta_{t} L)
        d \sigma_{dp,t}^2  
        ,
    \end{align*}
    where the inequality follows from~\prettyref{lmm: true momentum decay v2}.
    Furthermore,
     \begin{align*}
            &\frac{\kappa z_3}{L}
            \Expect \Big[
            \frac{1}{\abth{\calG}}
            \sum_{i \in \calG}
            \norm{\m_i^{t+1} - \overline{\m}^{t+1}}^2
            -
            \frac{1}{\abth{\calG}}
            \sum_{i \in \calG}
            \norm{\m_i^{t} - \overline{\m}^{t}}^2 \mid \calF^{t} \Big] \\
            &\le
            \frac{\kappa z_3}{L}
            \pth{\alpha_{t+1} + 6 \kappa (1 - \alpha_{t+1}) (1 - \beta_t)^2 B^2 \eta_{t}^2 L^2- 1}
            \frac{1}{\abth{\calG}}
            \sum_{i \in \calG}
            \norm{\m_i^{t} - \overline{\m}^{t}}^2 \\
            &~~~+
            \frac{\kappa z_3}{L}
            (1 - \alpha_{t+1}) G^2
            +
            2(1 - \alpha_{t+1})^2
            \frac{\kappa z_3}{L}
            \pth{1 + \frac{1}{\abth{\calG}}}
            \sigma^2 \\
            &~~~+
            2 (1 - \alpha_{t+1}) B^2 (1 + 3 \eta_{t}^2 L^2)
            \frac{\kappa z_3}{L}
            \norm{\nabla \byzobj (\bfthe^{t})}^2
            +
            6 (1 - \alpha_{t+1}) B^2 \eta_{t}^2 L
            \kappa z_3
            \norm{\overline{\M}^{t} - \nabla \byzobj (\bfthe^t)}^2 \\
            &~~~+
            2 (1 - \alpha_{t+1}) B^2 \eta_{t}^2 L
            \kappa z_3
            d \sigma_{dp, t}^2
            ,
        \end{align*}
        where the inequality follows from~\prettyref{lmm: sc momentum drift v2}.
        Putting them together and taking expectation over the remaining randomness, it holds that
        \begin{equation*}
        \hspace{-2.5cm}
        \begin{aligned}
            &W^{t+1} - W^t 
            \le
            - \eta_t
            \pth{\frac{3}{4} 
            - 2 \eta_t L
            - 3 z_1 
            \pth{\beta_{t+1} + (1 - \beta_{t+1}) \alpha_{t+1}}
            \pth{\pth{\beta_{t+1} + (1 - \beta_{t+1}) \alpha_{t+1}}
            \pth{1 +\eta_{t} L} + \beta_{t+1} (1 - \beta_{t+1})}
            }
            \expect{\norm{\nabla \byzobj (\bfthe^{t})}^2} \\
            &~~~- \eta_t
            \pth{
            -3 z_2 \alpha_{t+1}^2 
             (1  + \eta_{t} L)  
            -2 (1 - \alpha_{t+1}) B^2 (1 + 3 \eta_{t}^2 L^2) \frac{\kappa z_3}{\eta_t L}
            }
            \expect{\norm{\nabla \byzobj (\bfthe^{t})}^2} \\
            &~~~+
            \pth{2 \eta_t (1 + \eta_t L) + 3 z_2 \alpha_{t+1}^2 \eta_{t} (1 + \eta_{t} L) + 6 (1 - \alpha_{t+1}) B^2 \eta_{t}^2 L \kappa z_3}
        \expect{\norm{\overline{\M}^{t} - \nabla \byzobj(\bfthe^{t})}^2}\\    
        &~~~+
        \frac{z_1}{L}
        \pth{\beta_{t+1} \pth{
        \beta_{t+1}  +
         \frac{(1 - \beta_{t+1}) \alpha_{t+1} + (1 - \beta_{t})\alpha_{t+1}}{2 k_2}
        +
        \eta_{t} L
        (\beta_{t+1} + (1 - \beta_{t+1}) \alpha_{t+1})
        } - 1}
        \expect{\norm{\overline{\M}^{t} - \nabla \byzobj(\bfthe^{t})}^2 }\\
        &~~~+ 3 z_1 \eta_{t}
        \pth{\beta_{t+1} + (1 - \beta_{t+1}) \alpha_{t+1}}
        \pth{\pth{\beta_{t+1} + (1 - \beta_{t+1}) \alpha_{t+1}}
        \pth{1 +\eta_{t} L} + \beta_{t+1} (1 - \beta_{t+1})}
        \expect{\norm{\overline{\M}^{t} - \nabla \byzobj(\bfthe^{t})}^2 }\\
        &~~~+
        \frac{z_1}{L}
        (1 - \beta_{t+1}) \alpha_{t+1}
        \qth{\pth{(1 - \beta_{t+1}) \alpha_{t+1}
        +
        2 k_2 \beta_{t+1}}
        \pth{
         1
        +
        \eta_{t} L
        } + (1 - \beta_{t+1}) \beta_{t+1}}
       \expect{\norm{
        \overline{\m}^{t}
        -
        \nabla \byzobj (\bfthe^{t})
        }^2 }
        \\
        &~~~+
        \frac{z_2}{L}
        \pth{\alpha_{t+1}^2 (1  + \eta_{t} L)  - 1}
        \expect{\norm{\overline{\m}^{t} - \nabla \byzobj (\bfthe^{t})}^2 } 
        +
        \frac{2
        \kappa  
        \eta_t (1 + \eta_t L) (1 - \beta_t)^2
        }{\abth{\calG}}
        \sum_{i \in \calG}
        \expect{\norm{
        \m_i^{t}
        -
        \overline{\m}^t
        }^2} \\
        &~~~+\frac{z_1}{L}
        (1 - \beta_{t})
        \beta_{t+1}
        \pth{2 k_2 \beta_{t+1} + (1 - \beta_{t+1}) \alpha_{t+1}
        + \frac{\eta_{t} L (1 - \beta_{t+1})}{1 - \beta_{t}} 
        (\beta_{t+1} + (1-\beta_{t+1})\alpha_{t+1})}
        \frac{\kappa}{\abth{\calG}}
        \sum_{i \in \calG}
        \expect{\norm{\m_i^{t} - \overline{\m}^{t}}^2 } \\
        &~~~+
        3 z_1 (1 - \beta_{t})^2
        \eta_{t}
        \pth{\beta_{t+1} + (1 - \beta_{t+1}) \alpha_{t+1}}
        \pth{\pth{\beta_{t+1} + (1 - \beta_{t+1}) \alpha_{t+1}}
        \pth{1 +\eta_{t} L} + \beta_{t+1} (1 - \beta_{t+1})}
        \frac{\kappa}{\abth{\calG}}
        \sum_{i \in \calG}
        \expect{\norm{\m_i^{t} - \overline{\m}^{t}}^2 } \\
        &~~~+
        3 z_2 \alpha_{t+1}^2 
        \eta_{t}  (1  + \eta_{t} L)
        \frac{(1 - \beta_{t})^2 \kappa}{\abth{\calG}}
        \sum_{i \in \calG}
        \expect{\norm{\m_i^{t} - \overline{\m}^{t}}^2 \mid \calF^{t}} \\
        &~~~+\frac{\kappa z_3}{L}
        \pth{\alpha_{t+1} + 6 \kappa (1 - \alpha_{t+1}) (1 - \beta_t)^2 B^2 \eta_{t}^2 L^2 - 1}
        \frac{1}{\abth{\calG}}
        \sum_{i \in \calG}
        \expect{\norm{\m_i^{t} - \overline{\m}^{t}}^2 }\\
        &~~~+
        \frac{(1 - \alpha_{t+1})^2}{L}
        \pth{z_1
        (1 - \beta_{t+1})^2
        +
        z_2
        +
        2 \kappa z_3}
        \pth{1 + \frac{1}{\abth{\calG}}}
        \sigma^2 
        +
        \frac{\kappa z_3}{L}
        (1 - \alpha_{t+1}) G^2 \\
        &~~~+
        \pth{
        \frac{\eta_t^2 L}{2}
        +
        \frac{z_1}{L}
        \pth{\beta_{t+1}^2
        +
        \eta_{t} L
        \pth{\beta_{t+1} + (1 - \beta_{t+1}) \alpha_{t+1}}
        \pth{\pth{\beta_{t+1} + (1 - \beta_{t+1}) \alpha_{t+1}}
        \pth{1 +\eta_{t} L} + \beta_{t+1} (1 - \beta_{t+1})}}
        }
        d \sigma_{dp, t}^2\\
        &~~~+
        \pth{
        z_2
        \alpha_{t+1}^2 
        \eta_{t} (1  + \eta_{t} L)
        +
        2 (1 - \alpha_{t+1}) B^2 \eta_{t}^2 L
        \kappa z_3
        }
        d \sigma_{dp, t}^2
        .
        \end{aligned}
        \end{equation*}
        Let $\alpha_t = 1 - k_1 \eta_t L$ and $\beta_t = 1 - k_2 \eta_t L$, where $k_2 \le k_1$. 
        More specifically, $k_1 = 4900$, and $k_2 = 55$. 
        It follows that $\alpha_t \le \beta_t$.
        We know that
        \begin{align*}
            \frac{\eta_{t+1}}{\eta_t}
            &=
            \frac{\frac{10}{\mu \pth{t + 1 + a \frac{L}{\mu}}}}{\frac{10}{\mu \pth{t + a \frac{L}{\mu}}}}
            =
            \frac{t + a \frac{L}{\mu}}{t + 1 + a \frac{L}{\mu}}
            =
            1 - \frac{1}{t + 1 + a \frac{L}{\mu}}
            \in [\frac{a }{1 +a}, 1],
        \end{align*}
        where the last inequality follows from $t \ge 0$ and $L \ge \mu$.
        Recall that
        $\eta_t 
        \le \frac{1}{49000 L}$, and $B^2 \le \frac{1}{40 k_1 \kappa}$.
        Next, we bound the coefficients by applying the learning rate condition for $\eta_t$. 
        
        \paragraph{Bounding $C_1$ for $\expect{\norm{\nabla \byzobj (\bfthe^t)}^2}$.}
        \begin{equation*}
        \hspace{-2cm}
        \begin{aligned}
            &C_1 \triangleq \frac{3}{4} 
            - 2 \eta_t L
            - 3 z_1 
            \pth{\beta_{t+1} + (1 - \beta_{t+1}) \alpha_{t+1}}
            \pth{\pth{\beta_{t+1} + (1 - \beta_{t+1}) \alpha_{t+1}}
            \pth{1 +\eta_{t} L} + \beta_{t+1} (1 - \beta_{t+1})} \\
            &\qquad \qquad -3 z_2 \alpha_{t+1}^2 
             (1  + \eta_{t} L)  
            -2 (1 - \alpha_{t+1}) B^2 (1 + 3 \eta_{t}^2 L^2) \frac{\kappa z_3}{\eta_t L} \\
            &\ge \frac{3}{4} 
            - 2 \eta_t L
            - 3 z_1 
            \pth{1 + k_2 \eta_t L}
            \pth{\pth{1 + k_2 \eta_t L}
            \pth{1 +\eta_{t} L} +  k_2 \eta_t L} 
            -3 z_2 (1  + \eta_{t} L)  
            -2 k_1 B^2 (1 + 3 \eta_{t}^2 L^2) \kappa z_3 \\
            &\ge
            \frac{3}{4} - 2 \times \frac{1}{150}
            - \frac{3}{20} \times \pth{1 + \frac{8}{150}} \times \pth{\pth{1 + \frac{8}{150}} \times \pth{1 + \frac{1}{150}} + \frac{8}{150}}
            - \frac{1}{5} \times \pth{1 + \frac{1}{150}} - \frac{4}{40} \pth{1 + 3 \times (\frac{1}{150})^2} \\
            &\ge
            \frac{3}{4} - \frac{1}{75} - 0.19921 - 0.252 - 0.105
            \ge \frac{1}{10}.
        \end{aligned}
        \end{equation*}
        \paragraph{Bounding  $C_2$ for $\expect{\norm{\overline{\M}^{t} - \nabla \byzobj(\bfthe^{t})}^2}$.}
        \begin{equation*}
        \begin{aligned}
            &C_2 \triangleq
            \pth{2 \eta_t (1 + \eta_t L) + 3 z_2 \alpha_{t+1}^2 \eta_{t} (1 + \eta_{t} L) + 6 (1 - \alpha_{t+1}) B^2 \eta_{t}^2 L \kappa z_3} \\
            &~~~+
            \frac{z_1}{L}
            \pth{\beta_{t+1} \pth{
            \beta_{t+1}  +
            \frac{(1 - \beta_{t+1}) \alpha_{t+1} + (1 - \beta_{t+1}) \alpha_{t}}{2 k_2}
            +
            \eta_{t} L
            (\beta_{t+1} + (1 - \beta_{t+1}) \alpha_{t+1})
            } - 1} \\
            &~~~+
            3 z_1 \eta_{t}
            \pth{\beta_{t+1} + (1 - \beta_{t+1}) \alpha_{t+1}}
            \pth{\pth{\beta_{t+1} + (1 - \beta_{t+1}) \alpha_{t+1}}
            \pth{1 +\eta_{t} L} + \beta_{t+1} (1 - \beta_{t+1})} \\
            &\le
            \pth{2 \eta_t (1 + \eta_t L) + 3 z_2 \eta_{t} (1 + \eta_{t} L) + 6 k_1 \eta_t L B^2 \eta_{t}^2 L \kappa z_3} \\
            &~~~+
            \frac{\eta_t z_1}{L}
            \pth{\pth{
            L
            +
            L
            (1 + k_2 \eta_t L)
            } - k_2 L \frac{\eta_{t+1}}{\eta_t}} 
            +
            3 z_1 
            \pth{1 + k_2 \eta_t L}
            \pth{\pth{1 + k_2 \eta_t L}
            \pth{1 +\eta_{t} L} + k_2 \eta_t L} 
            \\
            &\le
            \eta_t \pth{2 (1 + \frac{1}{150}) + \frac{1}{5} (1 + \frac{1}{150}) + \frac{12}{40} \times \frac{1}{150}} 
            +
            \frac{\eta_t}{20}
            \pth{\pth{
            1+1 + \frac{1}{10}
            } - k_2 \frac{a}{1+a}} 
            \\&~~~
            +
            \frac{3}{20} 
            \pth{1 + \frac{1}{10}}
            \pth{\pth{1 + \frac{1}{10}}
            \pth{1 + \frac{1}{150}} + \frac{1}{10}} \\
            &\le
            \frac{\eta_t}{20}
            \pth{50.5 - \frac{a k_2}{1+a}}
            \le
            - \eta_t
            ,
        \end{aligned}
        \end{equation*}
        where the last inequality holds because $k_2 \ge 51.5 (1 + \frac{1}{a})$.
        
        \paragraph{Bounding $C_3$ for $\expect{\norm{\overline{\m}^{t} - \nabla \byzobj(\bfthe^{t})}^2}$.}
        \begin{align*}
        C_3 &\triangleq
        \frac{z_1}{L}
        (1 - \beta_{t+1}) \alpha_{t+1}
        \qth{\pth{(1 - \beta_{t+1}) \alpha_{t+1}
        +
        2 k_2 \beta_{t+1}}
        \pth{
         1
        +
        \eta_{t} L
        } + (1 - \beta_{t+1}) \beta_{t+1}} 
        +
        \frac{z_2}{L}
        \pth{\alpha_{t+1}^2 (1  + \eta_{t} L)  - 1}\\
        &\le 
        \eta_t \pth{
        z_1
        k_2
        \qth{\pth{k_2 \eta_t L
        +
        2 k_2 }
        \pth{
         1
        +
        \eta_{t} L
        } + k_2 \eta_t L}
        +
        z_2
        \pth{1  - \frac{\eta_{t+1} k_1}{\eta_{t}}}}\\
        &\le\frac{\eta_t}{20}
        \pth{k_2^2
        \qth{\pth{\frac{1}{150}+2}
        \pth{1+\frac{1}{150}} + \frac{1}{10}}
        +
        \frac{4}{3}
        \pth{1  - \frac{a k_1}{1 + a}}
        } \\
        &\le
        \frac{\eta_t}{15}
        \pth{1.6 k_2^2 +1  - \frac{a k_1}{1 + a}} 
        \le
        - \eta_t,
        \end{align*}
        where the last inequality holds because $k_1 \ge \pth{1.6 k_2^2 + 16} (1 + \frac{1}{a})$.

        \paragraph{Bounding $C_4$ for $\sum_{i \in \calG} \expect{\norm{\m_i^{t} - \overline{\m}^{t}}^2} / \abth{\calG}$.}
        \begin{equation*}
        \hspace{-1cm}
        \begin{aligned}
        C_4 & \triangleq
        2 \eta_t (1 + \eta_t L) (1 - \beta_t)^2 
        +
        \frac{z_1}{L}
        (1 - \beta_{t+1})
        \beta_{t+1}
        \pth{2 k_2 \beta_{t+1} + (1 - \beta_{t+1}) \alpha_{t+1} + \frac{\eta_{t} L (1 - \beta_{t+1}) }{1 - \beta_t}(\beta_{t+1} + (1-\beta_{t+1})\alpha_{t+1})} \\
        &~~~+
         3 z_1 (1 - \beta_{t})^2
        \eta_{t}
        \pth{\beta_{t+1} + (1 - \beta_{t+1}) \alpha_{t+1}}
        \pth{\pth{\beta_{t+1} + (1 - \beta_{t+1}) \alpha_{t+1}}
        \pth{1 +\eta_{t} L} + \beta_{t+1} (1 - \beta_{t+1})} \\
        &~~~+
        3 z_2 \alpha_{t+1}^2 
        \eta_{t}  (1  + \eta_{t} L)
        (1 - \beta_{t})^2 
        + \frac{z_3}{L}
        \pth{\alpha_{t+1} + 6 \kappa (1 - \alpha_{t+1}) (1 - \beta_t)^2 B^2 \eta_{t}^2 L^2- 1} \\
        &\le
        2 \eta_t (1 + \eta_t L) (k_2 \eta_t L)^2
        +
        z_1 k_2 \eta_t
        \pth{2 k_2 + k_2 \eta_t L + \eta_{t} L (1 + k_2 \eta_t L)} \\
        &~~~+
         3 z_1 k_2^2 \eta_t^2 L^2
        \eta_{t}
        \pth{1 + k_2 \eta_t L}
        \pth{\pth{1 + k_2 \eta_t L}
        \pth{1 +\eta_{t} L} + k_2 \eta_t L} \\
        &~~~+
        3 z_2 
        \eta_{t}  (1  + \eta_{t} L)
        k_2^2 \eta_t^2 L^2
        + z_3
        \pth{6 \kappa k_1 \eta_t k_2^2 \eta_t^2 L^2 B^2 \eta_{t}^2 L^2- k_1 \eta_{t+1}} \\
        &\le
        \eta_t \Bigg(
        2 (1 + \frac{1}{150}) (\frac{1}{10})^2 
        +
        \frac{k_2}{20} 
        \pth{2 k_2 + \frac{1}{10} + \frac{1}{150} (1 + \frac{1}{10})} \\
        &~~~+
        \frac{3}{20}(\frac{1}{10})^2
        \pth{1 + \frac{1}{10}}
        \pth{\pth{1 + \frac{1}{10}}
        \pth{1 + \frac{1}{150}} + \frac{1}{10}}
        +
        \frac{1}{5} 
         (1  + \frac{1}{10})
        (\frac{1}{10})^2
        + 2
        \pth{\frac{1}{40}  (\frac{1}{10})^2  (\frac{1}{150})^2- \frac{a k_1}{1+a}} \Bigg) \\
        &\le
        \eta_t \pth{
        0.021 + 0.1 k_2^2 + 5.4 \times 10^{-3} k_2 + 2 \times 10^{-3}
        + 2.2 \times 10^{-3} +  2.3 \times 10^{-8} - \frac{2 a k_1}{1+a}
        } \\
        &\le
        \eta_t \pth{
        0.0253 + 0.1 k_2^2 + 5.4 \times 10^{-3} k_2  - \frac{2 a k_1}{1+a}
        } 
        \le
        - \eta_t,
        \end{aligned}
        \end{equation*}
        where the last inequality holds because $k_1 \ge \pth{0.6265 + 0.05 k_2^2 + 2.7\times 10^{-3} k_2 } \pth{1 + \frac{1}{a}}$.
        Finally, we bound the remaining noise coefficients.
        \paragraph{Coefficient $C_5$ of $\sigma^2$.}
        \begin{align*}
            C_5 \triangleq
            \frac{(1 - \alpha_{t+1})^2}{L}
            \pth{z_1
            (1 - \beta_{t+1})^2
            +
            z_2
            +
            2 \kappa z_3}
            &\le
            k_1^2 \eta_t^2 L
            \pth{\frac{1}{20}
            \pth{\frac{1}{10}}^2
            +
            \frac{1}{15}
            +
            4 \kappa}
            \le
            k_1^2 \eta_t^2 L
            \pth{1 + 4 \kappa}
            ,
        \end{align*}
        
        \paragraph{Coefficient $C_6$ of $d \sigma_{dp,t}^2$.}
        \begin{small}
        \begin{align*}
            &C_6 \triangleq
            \pth{
        \frac{\eta_t^2 L}{2}
        +
        \frac{z_1}{L}
        \pth{\beta_{t+1}^2
        +
        \eta_{t} L
        \pth{\beta_{t+1} + (1 - \beta_{t+1}) \alpha_{t+1}}
        \pth{\pth{\beta_{t+1} + (1 - \beta_{t+1}) \alpha_{t+1}}
        \pth{1 +\eta_{t} L} + \beta_{t+1} (1 - \beta_{t+1})}}
        }\\
        &~~~+
        \pth{
        z_2
        \alpha_{t+1}^2 
        \eta_{t+1} (1  + \eta_{t} L)
        +
        2 (1 - \alpha_{t+1}) B^2 \eta_{t}^2 L
        \kappa z_3
        } \\
        &\le
        \frac{1}{L}
        \pth{\frac{1}{2} \pth{\frac{1}{49000}}^2
        + \frac{1}{20} \pth{1 + \frac{1}{49000} \pth{1 + \frac{1}{10}} \pth{\pth{1 + \frac{1}{10}}\pth{1 + \frac{1}{49000}}+ \frac{1}{10}}}} \\
        &~~~+
        \frac{1}{L}
        \pth{
        \frac{1}{15}
        \frac{1}{49000} (1  + \frac{1}{49000})
        +
        \frac{2\times 2}{40}  \pth{\frac{1}{49000}}^2
        }
        \le
        \frac{1}{L}\pth{0.051 + 1. 4 \times 10^{-6}} \le \frac{0.06}{L}.
        \end{align*}
        \end{small}
        Putting them together.
        \begin{align*}
            &W^{t+1} - W^t 
            \le
            - \frac{\eta_t}{10}
            \expect{\norm{\nabla \byzobj (\bfthe^{t})}^2} 
            - \eta_t
            \expect{
            \norm{\nabla \byzobj(\bfthe^t) - \overline{\M}^t}^2}
        - \eta_t
       \expect{\norm{
        \overline{\m}^{t}
        -
        \nabla \byzobj (\bfthe^{t})
        }^2 }
        \\
        &~~~-
        \frac{
        \kappa  
        \eta_t
        }{\abth{\calG}}
        \sum_{i \in \calG}
        \expect{\norm{
        \m_i^{t}
        -
        \overline{\m}^t
        }^2} 
        +
        k_1^2 \eta_t^2 L (1 + 4 \kappa)
        \pth{1 + \frac{1}{\abth{\calG}}}
        \sigma^2 
        +
        2 \kappa  
        k_1 \eta_t G^2 
        +
        \frac{0.06}{L}
        d \sigma_{dp, t}^2 \\
        &\le
        - \frac{\mu \eta_t}{5}
        \pth{\expect{\byzobj(\bfthe^t) - \byzobj^{\ast}}
        +
        \frac{z_1}{\mu}
        \expect{    
        \norm{\nabla \byzobj(\bfthe^t) - \overline{\M}^t}^2}
        +
        \frac{z_2}{\mu}
        \expect{\norm{
        \overline{\m}^{t} - \nabla \byzobj (\bfthe^{t})}^2 }
        +
        \frac{\kappa z_3}{\mu}
        \frac{1}{\abth{\calG}}
        \sum_{i \in \calG}
        \expect{\norm{
        \m_i^{t}
        -
        \overline{\m}^t
        }^2} 
        } \\
        &~~~+
        k_1^2 \eta_t^2 L (1 + 4 \kappa)
        \pth{1 + \frac{1}{\abth{\calG}}}
        \sigma^2 
        +
        2 \kappa  
        k_1 \eta_t G^2 
        +
        \frac{0.06}{L}
        d \sigma_{dp, t}^2 \\
        &\overset{(a)}{\le}
        - \frac{\mu \eta_t}{5}
        \pth{\expect{\byzobj(\bfthe^t) - \byzobj^{\ast}}
        +
        \frac{z_1}{L}
        \expect{    
        \norm{\nabla \byzobj(\bfthe^t) - \overline{\M}^t}^2}
        +
        \frac{z_2}{L}
        \expect{\norm{
        \overline{\m}^{t} - \nabla \byzobj (\bfthe^{t})}^2 }
        +
        \frac{\kappa z_3}{L}
        \frac{1}{\abth{\calG}}
        \sum_{i \in \calG}
        \expect{\norm{
        \m_i^{t}
        -
        \overline{\m}^t
        }^2} 
        } \\
        &~~~+
        k_1^2 \eta_t^2 L (1 + 4 \kappa)
        \pth{1 + \frac{1}{\abth{\calG}}}
        \sigma^2 
        +
        2 \kappa  
        k_1 \eta_t G^2 
        +
        \frac{0.06}{L}
        d \sigma_{dp, t}^2 \\
        &=
        - \frac{\mu \eta_t}{5}
        W^t
        +
        k_1^2 \eta_t^2 L (1 + 4 \kappa)
        \pth{1 + \frac{1}{\abth{\calG}}}
        \sigma^2 
        +
        2 \kappa  
        k_1 \eta_t G^2 
        +
        \frac{0.06}{L}
        d \sigma_{dp, t}^2 
        ,
        \end{align*}
        where inequality $(a)$ holds because $\mu \le L$.
        By plugging $\eta_t$ in, we have
        \begin{align*}
            V^{t+1} - V^{t}
            &\le
            - \pth{\frac{2 (\hat{t} + 1)^2}{\hat{t}} - (2 \hat{t} + 1)}
            W^t 
            +
            100 k_1^2 \frac{(\hat{t} + 1)^2}{\mu^2 (\hat{t})^2} L (1 + 4 \kappa)
            \pth{1 + \frac{1}{\abth{\calG}}}
            \sigma^2  \\
            &~~~+
            20  \kappa  
            k_1 \frac{(\hat{t} + 1)^2}{\mu \hat{t}} G^2 
            +
            \frac{0.06 (\hat{t} + 1)^2}{L} 
            d \sigma_{dp, t}^2 
            .
        \end{align*}
        We know that
        \[
            \frac{2 (\hat{t} + 1)^2}{\hat{t}} - (2 \hat{t}+1)
            \ge
            \frac{2 (\hat{t} + 1)^2}{\hat{t} + 1} - (2 \hat{t}+1)
            \ge
            1 >0.
        \]
        It can be readily checked that,
        \begin{align*}
            \pth{\frac{\hat{t} + 1}{\hat{t}}}^2
            &=
            \pth{1 + \frac{1}{\hat{t}}}^2
            \le
            \pth{1 + \frac{1}{a}}^2
            \le
            2.
        \end{align*}
        We also know that
    \begin{align*}
        d \sigma_{dp, r}^2 
        \le
        \frac{24 (1 - \beta_r)^2 T \log(1 / \delta) C^2 d^2 }{\epsilon^2 n^2}
        \pth{1 + \frac{4(2+\nu)^2}{\nu^2}\frac{f}{n}} 
        =
        T \eta_r^2 L^2
        \frac{24 k_2^2  \log(1 / \delta) C^2 d^2}{\epsilon^2 n^2}
        \pth{1 + \frac{4(2+\nu)^2}{\nu^2}\frac{f}{n}} 
        .
    \end{align*}
    For ease of presentation, let 
    \[
     C_{dp} \triangleq \frac{24 k_2^2 \log(1 / \delta) C^2 d}{\epsilon^2 n^2} \pth{1 + \frac{4 (2 + \nu)^2}{\nu^2} \frac{f}{n}}.
    \]
         It follows that
        \begin{align*}
            V^{t+1} - V^{t}
            &\le
            200 k_1^2 L (1 + 4 \kappa)
            \pth{1 + \frac{1}{\abth{\calG}}}
            \frac{\sigma^2 }{\mu^2}
            +
            20 \kappa  
            k_1 \hat{t} \frac{G^2 }{\mu}
            +
            \frac{0.12 T L C_{dp}}{\mu^2}
            .
        \end{align*}
        By telescoping sum, we have
        \begin{align*}
            V^{t+1} - V^{0}
            &=
            \sum_{k=0}^{t}
            \pth{V^{k+1} - V^{k}}
            \\
            &\le
            200 (t + 1) k_1^2 L (1 + 4 \kappa)
            \pth{1 + \frac{1}{\abth{\calG}}}
            \frac{\sigma^2 }{\mu^2}
            +
            \frac{0.12 T L (t+1) C_{dp}}{\mu^2}
            +
            \frac{20}{\mu}
            \pth{\sum_{r=0}^t \pth{r + a \frac{L}{\mu}}}
            \kappa  
            k_1  G^2 
            .
        \end{align*}
         We know that
        \[
        \sum_{r=0}^{t} \pth{r + a \frac{L}{\mu}}
        =
        \frac{t (t+1)}{2}
        +
        a (t+1) \frac{L}{\mu}
        .
        \]
        Therefore,
        \begin{align*}
            V^{t+1} - V^{0}
            &=
            \sum_{k=0}^{t}
            \pth{V^{k+1} - V^{k}}
            \\
            &\le
            200 (t + 1)
            \pth{k_1^2 L (1 + 2 \kappa)
            \pth{1 + \frac{1}{\abth{\calG}}}
            \frac{\sigma^2 }{\mu^2}
            +
            \frac{T L C_{dp}}{\mu^2}
            +
            \frac{1}{\mu}
            \pth{\frac{t}{2} + a \frac{L}{\mu}}
            \kappa  k_1  G^2 
            }
            .
        \end{align*}
         By definition, it holds that
        \[
            V^{t+1}
            \ge
            (t+1 + a \frac{L}{\mu})^2
            \expect{\byzobj(\bfthe^{t+1}) - \byzobj^{\ast}}
            .
        \]
        Rearrange the terms, we have
        \begin{align*}
            \expect{\byzobj(\bfthe^{t+1}) - \byzobj^{\ast}}
            &\le
            \frac{V_0}{(t+1 + a \frac{L}{\mu})^2} \\
            &~~~+
            \frac{200(t+1)}{(t+1 + a \frac{L}{\mu})^2}
            \pth{k_1^2 L (1 + 2 \kappa)
            \pth{1 + \frac{1}{\abth{\calG}}}
            \frac{\sigma^2 }{\mu^2}
            +
            \frac{T L C_{dp}}{\mu^2}
            +
            \frac{1}{\mu}
            \pth{\frac{t}{2} + a \frac{L}{\mu}}
            \kappa  k_1  G^2 
            } \\
            &\le
            \frac{V_0}{(t+1 + a \frac{L}{\mu})^2} 
            +
            \frac{200}{t+1 + a \frac{L}{\mu}}
            \pth{k_1^2 L (1 + 2 \kappa)
            \pth{1 + \frac{1}{\abth{\calG}}}
            \frac{\sigma^2 }{\mu^2}
            +
            \frac{T L C_{dp}}{\mu^2}
            } 
            +
            \frac{100 }{\mu} 
            \kappa  
            k_1  G^2 
            .
        \end{align*}
        It remains to bound $V_0.$
        \begin{align*}
            V_0 &=
            \pth{a \frac{L}{\mu}}^2
            \pth{
            \byzobj (\bfthe^0) - \byzobj^{\ast}
            +
            \frac{z_1}{L} \norm{\nabla \byzobj (\bfthe^0) - \overline{\M}^0}^2
            +
            \frac{z_2}{L} \norm{\nabla \byzobj (\bfthe^0) - \overline{\m}^0}^2
            +
            \frac{\kappa z_3}{L}
            \frac{1}{\abth{\calG}}
            \sum_{i \in \calG}
            \norm{\m_i^t - \overline{\m}^t}^2
            }
            .
        \end{align*}
By definition, we have
\begin{align*}
    \overline{\m}^0 
    &=
    \beta_{0} R^{-1} + (1 - \beta_{0}) \overline{\m}^0 
    =
    \beta_{0} R^{-1} + (1 - \beta_{0}) \alpha_{0}\overline{\m}^{-1} 
    +
    (1 - \beta_{0}) (1 - \alpha_{0}) \overline{\g}^{0}
    =
    (1 - \beta_{0}) (1 - \alpha_{0}) \overline{\g}^{0}
    ,
\end{align*}
where the last equality follows from the convention that $R^{-1} = \overline{\m}^{-1} = \bm{0}$.
It follows that
\begin{align*}
    \norm{\overline{\M}^0 - \nabla \byzobj (\bfthe^{0})}^2
    &=
    \norm{(1 - \beta_{0}) (1 - \alpha_{0}) \pth{\overline{\g}^{0} - \nabla \byzobj (\bfthe^{0})} - (1 - (1 - \beta_{0}) (1 - \alpha_{0}))\nabla \byzobj (\bfthe^{0})}^2 \\
    &\le
    (1 - \beta_{0}) (1 - \alpha_{0})
    \norm{\overline{\g}^{0} - \nabla \byzobj (\bfthe^{0})}^2 + 
    (1 - (1 - \beta_{0}) (1 - \alpha_{0}))
    \norm{\nabla \byzobj (\bfthe^{0})}^2,
\end{align*}
where the last inequality holds because of Jensen's inequality.
Take expectation, we have
\begin{align*}
    \expect{\norm{\overline{\M}^0 - \nabla \byzobj (\bfthe^{0})}^2}
    &\le
    (1 - \beta_{0}) (1 - \alpha_{0})
    \expect{\norm{\overline{\g}^{0} - \nabla \byzobj (\bfthe^{0})}^2} + 
    (1 - (1 - \beta_{0}) (1 - \alpha_{0}))
    \expect{\norm{\nabla \byzobj (\bfthe^{0})}^2} \\
    &\le
    \frac{\sigma^{2}}{\abth{\calG}}
    + 
    \expect{\norm{\nabla \byzobj (\bfthe^{0})}^2}
\end{align*}

Similarly, we have
\begin{align*}
    \expect{\norm{\overline{\m}^{0} - \nabla \byzobj (\bfthe^{0})}^2}
    &=
    \expect{\norm{\alpha_{0} \overline{\m}^{-1} + (1 - \alpha_{0})\pth{\overline{\g}^0 - \nabla \byzobj (\bfthe^{0})} - \alpha_{0} \nabla \byzobj (\bfthe^{0})}^2} \\
    &\overset{(a)}{=}
    \expect{\norm{(1 - \alpha_{0})\pth{\overline{\g}^0 - \nabla \byzobj (\bfthe^{0})} - \alpha_{0} \nabla \byzobj (\bfthe^{0})}^2} \\
    &\overset{(b)}{=}
    \expect{\norm{(1 - \alpha_{0})\pth{\overline{\g}^0 - \nabla \byzobj (\bfthe^{0})}}^2} + \expect{\norm{\alpha_{0} \nabla \byzobj (\bfthe^{0})}^2} \\
    &\le
    \frac{\sigma^2}{\abth{\calG}}
    +
    \expect{\norm{\nabla \byzobj (\bfthe^{0})}^2}
    ,
\end{align*}
where equality $(a)$ follows from the convention that $\overline{\m}^{-1} = \bm{0}$, and equality $(b)$ follows from the independence in stochastic gradients.
Next, we have
 \begin{align*}
    \norm{\nabla \byzobj (\bfthe^0) - \overline{\M}^0}^2 
    &=\norm{\nabla \byzobj (\bfthe^0) - \overline{\M}^0}^2 
    =\norm{\nabla \byzobj (\bfthe^0)}^2 
    \le 2 L (\byzobj (\bfthe^0) - \byzobj^{\ast})
    ,
\end{align*}
and
\begin{align*}
    &\frac{1}{\abth{\calG}}
    \sum_{i \in \calG}
    \expect{\norm{\m_i^{0} - \overline{\m}^{0}}^2 } \\
    &= 
    \frac{1}{\abth{\calG}}
    \sum_{i \in \calG}
    \expect{\norm{
    (1 - \alpha_{0}) \pth{(\g_i^0  - \nabla F_i (\bfthe^0) )- (\overline{\g}^0 - \nabla \byzobj(\bfthe^0)}
    +
    (1 - \alpha_0) (\nabla F_i (\bfthe^0) -  \nabla \byzobj (\bfthe^0))
    }^2 } \\
    &=
    \frac{1}{\abth{\calG}}
    \sum_{i \in \calG}
    \expect{\norm{
    (1 - \alpha_{0}) \pth{(\g_i^0  - \nabla F_i (\bfthe^0) )- (\overline{\g}^0 - \nabla \byzobj(\bfthe^0)}}^2
    +
    \norm{
    (1 - \alpha_0) (\nabla F_i (\bfthe^0) -  \nabla \byzobj (\bfthe^0))
    }^2 } \\
    &\le
    2 \pth{1 + \frac{1}{\abth{\calG}}}
    \sigma^2
    +
    B^2
    \expect{\norm{\nabla \byzobj (\bfthe^{0})}^2}
    +
    G^2
    .
\end{align*}
Combining everything together, it holds that
\begin{align*}
    V_0 &=
    \pth{a \frac{L}{\mu}}^2
    \pth{
    \expect{\byzobj (\bfthe^0)} - \byzobj^{\ast}
    +
    \frac{z_1}{L} \norm{\nabla \byzobj (\bfthe^0) - \overline{\M}^0}^2
    +
    \frac{z_2}{L} \norm{\nabla \byzobj (\bfthe^0) - \overline{\m}^0}^2
    +
    \frac{\kappa z_3}{L}
    \frac{1}{\abth{\calG}}
    \sum_{i \in \calG}
    \norm{\m_i^t - \overline{\m}^t}^2
    } \\
    &\le
    \pth{a \frac{L}{\mu}}^2
    \pth{
    \expect{\byzobj (\bfthe^0)} - \byzobj^{\ast}
    +
    \frac{z_1 + z_2}{L}
    \pth{\frac{\sigma^{2}}{\abth{\calG}}
    + 
    \expect{\norm{\nabla \byzobj (\bfthe^{0})}^2}}} \\
    &\qquad \qquad \qquad +
    \pth{a \frac{L}{\mu}}^2
    \frac{\kappa z_3}{L}
    \pth{
     2 \pth{1 + \frac{1}{\abth{\calG}}}
    \sigma^2
    +
    B^2
    \expect{\norm{\nabla \byzobj (\bfthe^{0})}^2}
    +
    G^2
    } \\
    &\le
    \pth{a \frac{L}{\mu}}^2
    \pth{
    6
    \pth{\expect{\byzobj (\bfthe^0)} - \byzobj^{\ast}}
    +
    \frac{2}{L}
    \frac{\sigma^{2}}{\abth{\calG}
    }
    +
    \frac{2 \kappa}{L}
    \pth{
     2 \pth{1 + \frac{1}{\abth{\calG}}}
    \sigma^2
    +
    G^2
    }
    } 
    .
\end{align*}
Putting everything together, we have
\begin{align*}
    \expect{\byzobj(\bfthe^{t+1}) - \byzobj^{\ast}}
    &\le
    \frac{6 a^2 L^2 (\expect{\byzobj (\bfthe^0)} - \byzobj^{\ast})}{\mu^2 (t+1 + a \frac{L}{\mu})^2} 
    +
    \frac{2 a^2 L \sigma^2}{\mu^2 (t+1 + a \frac{L}{\mu})^2 \abth{\calG}} 
    +
    \frac{2 a^2 L \kappa}{\mu^2 (t+1 + a \frac{L}{\mu})^2} 
      \pth{
     2 \pth{1 + \frac{1}{\abth{\calG}}}
    \sigma^2
    +
    G^2
    }
    \\
    &~~~+\frac{200}{t+1 + a \frac{L}{\mu}}
    \pth{k_1^2 L (1 + 2 \kappa)
    \pth{1 + \frac{1}{\abth{\calG}}}
    \frac{\sigma^2 }{\mu^2}
    +
    \frac{T L C_{dp}}{\mu^2}
    } 
    +
    \frac{100 }{\mu} 
    \kappa  
    k_1  G^2 \\
    &\le
    \frac{6 a^2 L^2 (\expect{\byzobj (\bfthe^0)} - \byzobj^{\ast})}{\mu^2 (t+1)^2} 
    +
    \frac{2 a^2 L \sigma^2}{\mu^2 (t+1)^2 \abth{\calG}} 
    +
    \frac{2 a^2 L \kappa}{\mu^2 (t+1)^2} 
      \pth{
     2 \pth{1 + \frac{1}{\abth{\calG}}}
    \sigma^2
    +
    G^2
    }
    \\
    &~~~+\frac{200}{t+1}
    \pth{k_1^2 L (1 + 2 \kappa)
    \pth{1 + \frac{1}{\abth{\calG}}}
    \frac{\sigma^2 }{\mu^2}
    +
    \frac{T L C_{dp}}{\mu^2}
    } 
    +
    \frac{100 }{\mu} 
    \kappa  
    k_1  G^2
    .
\end{align*}
For $\mu$-strongly convex objectives, we also know that
\begin{align*}
    \byzobj (\bfthe^t) - \byzobj^{\ast}
    &\ge
    \frac{\mu}{2}
    \norm{\bfthe^t - \bfthe^{\ast}}^2
    .
\end{align*}
Therefore, we have
\begin{align*}
    \expect{\norm{\bfthe^t - \bfthe^{\ast}}^2}
    &\lesssim
  \frac{L^2 (\expect{\byzobj (\bfthe^0)} - \byzobj^{\ast})}{\mu^3 t^2} 
    +
    \frac{L \sigma^2}{\mu^3 t^2 \abth{\calG}} 
    +
    \frac{L \kappa}{\mu^3 t^2} 
  \pth{
     \pth{1 + \frac{1}{\abth{\calG}}}
    \sigma^2
    +
    G^2
    }
    \\
    &~~~+\frac{1}{t}
    \pth{L (1 + \kappa)
    \pth{1 + \frac{1}{\abth{\calG}}}
    \frac{\sigma^2 }{\mu^3}
    +
    \frac{T L C_{dp}}{\mu^3}
    } 
    +
    \frac{\kappa  
G^2}{\mu^2} 
    .
\end{align*}
In particular, when $t = T$, it holds that
\begin{align*}
    \expect{\norm{\bfthe^T - \bfthe^{\ast}}^2}
    &\lesssim
  \frac{L^2 (\expect{\byzobj (\bfthe^0)} - \byzobj^{\ast})}{\mu^3 T^2} 
    +
    \frac{L \sigma^2}{\mu^3 T^2 \abth{\calG}} 
    +
    \frac{L \kappa}{\mu^3 T^2} 
  \pth{
     \pth{1 + \frac{1}{\abth{\calG}}}
    \sigma^2
    +
    G^2
    }
    \\
    &~~~+
    L (1 + \kappa)
    \pth{1 + \frac{1}{\abth{\calG}}}
    \frac{\sigma^2 }{\mu^3 T}
    +
    \frac{L C_{dp}}{\mu^3}
    +
    \frac{\kappa G^2}{\mu^2} \\
    &\lesssim
     \frac{L^2 (\expect{\byzobj (\bfthe^0)} - \byzobj^{\ast})}{\mu^3 T^2} 
    +
    \frac{L \sigma^2}{\mu^3 T^2 \abth{\calG}} 
    +
    \frac{L \kappa}{\mu^3 T^2} 
  \pth{
     \pth{1 + \frac{1}{\abth{\calG}}}
    \sigma^2
    +
    G^2
    }
    \\
    &~~~+
    L (1 + \kappa)
    \pth{1 + \frac{1}{\abth{\calG}}}
    \frac{\sigma^2 }{\mu^3 T}
    +
    \pth{\frac{L }{\mu}}
    \frac{\log(1 / \delta) C^2 d}{\mu^2 \epsilon^2 n^2} \pth{1 + \frac{f}{n}}
    +
    \frac{\kappa G^2}{\mu^2}
    .
\end{align*}
\end{proof}

\subsection{Non-convex objectives}
\label{app: stochastic convergence non-convex v2}
In the analysis of non-convex objectives, we adopt a constant learning rate, \ie, $\eta_t = \eta$, $\beta_t=\beta$, and $\alpha_t=\alpha$.

\subsubsection{Proof of
\prettyref{lmm: sc momentum drift v2}}

    It holds, for $i \in \calG$, that
    \begin{align*}
        \M_i^{t}
        -
        \overline{\M}^t
        &=
        \beta \tilde R^{t-1}
        +
        (1 - \beta)
        \hat \m_i^t
        -
        \pth{\beta \tilde R^{t-1} + (1 - \beta) \overline{\m}^t}
        =
        (1 - \beta) 
        \pth{\hat \m_i^t - \overline{\m}^t}.
    \end{align*}
    It follows that
    \begin{align}
        \label{eq: local global momentum mixing}
        \frac{1}{\abth{\calG}}
        \sum_{i \in \calG}
        \norm{\M_i^{t}
        -
        \overline{\M}^t}^2
        &=
        \frac{(1 - \beta)^2}{\abth{\calG}}
        \sum_{i \in \calG}
        \norm{\hat \m_i^t - \overline{\m}^t}^2
        =
        \frac{(1 - \beta)^2}{\abth{\calG}}
        \sum_{i \in \calG}
        \norm{\m_i^t - \overline{\m}^t}^2,
    \end{align}
    where the last equality holds from the fact that $\hat \m_{i}^{t} = \m_{i}^{t}$ for $i \in \calG$.
    Furthermore, it holds for $i \in \calG$ that
    \begin{align*}
        \m_i^t
        &=
        \alpha
        \m_i^{t-1}
        +
        \pth{1 - \alpha }
        \g_i^t 
        =
        \pth{1 - \alpha }
        \sum_{r=0}^t
        \alpha^{t - r}
        \g_i^r
        .
    \end{align*}
    It follows that
    \begin{align*}
        &\frac{1}{\abth{\calG}}
        \sum_{i \in \calG}
        \norm{\m_i^t - \overline{\m}^t}^2
        =
        \frac{\pth{1 - \alpha }^2
        }{\abth{\calG}}
        \sum_{i \in \calG}
        \norm{
        \sum_{r=0}^t
        \alpha^{t - r}
        \pth{\g_i^r - \overline{\g}^t}
        }^2 \\
        &\le
        \underbrace{\frac{3\pth{1 - \alpha }^2
        }{\abth{\calG}}
        \sum_{i \in \calG}
        \norm{
        \sum_{r=0}^t
        \alpha^{t - r}
        \pth{\g_i^r - \nabla F_i (\bfthe^r)}
        }^2}_{(A)} \\
        &~~~+
        \underbrace{\frac{3\pth{1 - \alpha }^2
        }{\abth{\calG}}
        \sum_{i \in \calG}
        \norm{
        \sum_{r=0}^t
        \alpha^{t - r}
        \pth{\nabla \byzobj (\bfthe^r) - \overline{\g}^r}
        }^2}_{(B)} 
        +
        \underbrace{\frac{3\pth{1 - \alpha }^2
        }{\abth{\calG}}
        \sum_{i \in \calG}
        \norm{
        \sum_{r=0}^t
        \alpha^{t - r}
        \pth{\nabla F_i (\bfthe^r) - \nabla \byzobj (\bfthe^r)}
        }^2 }_{(C)}
        ,
    \end{align*}
    where the inequality follows from Jensen's inequality.
    \paragraph{Bounding $(A)$.}
    \begin{align*}
        \frac{3\pth{1 - \alpha }^2
        }{\abth{\calG}}
        \sum_{i \in \calG}
        \expect{
        \norm{
        \sum_{r=0}^t
        \alpha^{t - r}
        \pth{\g_i^r - \nabla F_i (\bfthe^r)}
        }^2}
        &=
        \frac{3\pth{1 - \alpha }^2
        }{\abth{\calG}}
        \sum_{i \in \calG}
        \sum_{r=0}^t
        \alpha^{2(t - r)}
        \expect{
        \norm{
        {\g_i^r - \nabla F_i (\bfthe^r)}
        }^2}
        \le
        \frac{3\pth{1 - \alpha }^2}{1 - \alpha^2}
        \sigma^2
        ,
    \end{align*}
    where the equality and the inequality both follow from~\prettyref{ass: bounded variance client-wise}.
    \paragraph{Bounding $(B)$.}
    \begin{align*}
        \frac{3\pth{1 - \alpha }^2
        }{\abth{\calG}}
        \sum_{i \in \calG}
        \expect{
        \norm{
        \sum_{r=0}^t
        \alpha^{t - r}
        \pth{\nabla \byzobj (\bfthe^r) - \overline{\g}^r}
        }^2}
        &=
        \frac{3\pth{1 - \alpha }^2
        }{\abth{\calG}}
        \sum_{i \in \calG}
        \sum_{r=0}^t
        \alpha^{2(t - r)}
        \expect{
        \norm{
        {\nabla \byzobj (\bfthe^r) - \overline{\g}^r}
        }^2} 
        \le
        \frac{3\pth{1 - \alpha }^2}{1 - \alpha^2}
        \frac{\sigma^2}{\abth{\calG}}
        ,
    \end{align*}
    where the equality and the inequality both follow from~\prettyref{ass: bounded variance client-wise}.
    Furthermore, it holds that
    \begin{align*}
        \frac{(1 - \alpha)^2}{1 - \alpha^2}
        &=
        \frac{1 - \alpha}{1 + \alpha}
        =
        \frac{(1 - \alpha) (1 + \alpha)}{(1 + \alpha)^2}
        \le
        (1 - \alpha) (1 + \alpha)
        =
        1 - \alpha^2
        .
    \end{align*}
    \paragraph{Bounding $(C)$.}
    \begin{equation*}
    \hspace{-1cm}
    \begin{aligned}
        &\frac{3\pth{1 - \alpha}^2
        }{\abth{\calG}}
        \sum_{i \in \calG}
        \expect{
        \norm{
        \sum_{r=0}^t
        \alpha^{t - r}
        \pth{\nabla F_i (\bfthe^r) - \nabla \byzobj (\bfthe^r)}
        }^2 } 
        \le
        \frac{3\pth{1 - \alpha }^2
        }{\abth{\calG}}
        \pth{\sum_{r=0}^t
        \alpha^{t - r}}
        \sum_{i \in \calG}
        \sum_{r=0}^t
        \alpha^{t - r}
        \expect{
        \norm{
        {\nabla F_i (\bfthe^r) - \nabla \byzobj (\bfthe^r)}
        }^2 } \\
        &\qquad \qquad \qquad \qquad 
        \qquad
        \le
        {3\pth{1 - \alpha }
        }
        \sum_{r=0}^t
        \alpha^{t - r}
        \pth{
        B^2 
        \expect{\norm{
        \nabla \byzobj (\bfthe^r)
        }^2}
        + 
        G^2
        }
        \le
        3 G^2
        +
        3 
        B^2
        (1 - \alpha)
        \sum_{r=0}^t
        \alpha^{t - r}
        \expect{\norm{
        \nabla \byzobj (\bfthe^r)
        }^2}
        .
    \end{aligned}
    \end{equation*}
    Putting them together, it holds that
    \begin{equation*}
    \begin{aligned}
        \frac{1}{\abth{\calG}}
        \sum_{i \in \calG}
        \expect{
        \norm{\m_i^t - \overline{\m}^t}^2} 
        &\le
        3 
        \pth{1 + \frac{1}{\abth{\calG}}}
        \pth{1 - \alpha^2}
        \sigma^2
        +
        3 
        G^2 
        +
        3 B^2
        (1 - \alpha)
        \sum_{r=0}^t
        \alpha^{t - r}
        \expect{\norm{
        \nabla \byzobj (\bfthe^r)
        }^2}
        .
    \end{aligned}
    \end{equation*}

\subsubsection{Proof of~\prettyref{thm: non-convex learning convergence v2 main text}}
\begin{theorem}
    \label{thm: non-convex learning convergence v2}
    Suppose that Assumptions~\ref{ass: smmothness},~\ref{ass: bounded variance client-wise}, and~\ref{ass: bounded sto gradient cl} hold.
    Let $\alpha = 1 - 5100 \eta L$, $\beta = 1 - 50 \eta L$.
    $\eta < 1/ (5100 L)$, and $B^2 < 1/( 23625 \kappa)$,
    it holds for~\prettyref{alg: byzantine dp learning base stochastic v2} that

\begin{align*}
        \frac{1}{T}
        \sum_{t=0}^{T-1}
        \expect{\norm{\nabla \byzobj (\bfthe^{t})}^2} 
        &\le
        \frac{30 ( \expect{\byzobj (\bfthe^0)} - \byzobj^{\ast})}{\eta T}
        +
        \frac{25 d \sigma_{dp}^2 }{ \eta L}
        +
        \frac{2 \eta L}{T} 
        k_1 (k_1 + k_2)
        \frac{\sigma^2}{\abth{\calG}}
        +
        4 k_1^2 \eta L  \frac{\sigma^2}{\abth{\calG}}  \\
        &~~~+47250 k_1 \eta L
        \kappa 
        \pth{1 + \frac{1}{\abth{\calG}}}
        \sigma^2
        +
        23625 \kappa  G^2 
        .
\end{align*}
\end{theorem}
\begin{proof}
By~\prettyref{lmm: stochastic v2 descent}, we know that
\begin{align*}
    \expect{\byzobj (\bfthe^{t+1}) - \byzobj (\bfthe^{t}) \mid \calF^t}
    &\le
    - \eta
    \pth{\frac{3}{4} - 2 \eta L}
    \norm{\nabla \byzobj (\bfthe^{t})}^2
    +
    \frac{2
    \kappa  
    \eta (1 + \eta L)
    }{\abth{\calG}}
    \sum_{i \in \calG}
    \expect{\norm{
    \M_i^{t}
    -
    \overline{\M}^t
    }^2
    \mid
    \calF^t}
    \\
    &~~~
    +
    2 \eta (1 + \eta L)
    \expect{
    \norm{\nabla \byzobj(\bfthe^t) - \overline{\m}^t}^2
    \mid \calF^t
    }
    +
    \frac{\eta^2 L d}{2}
    \sigma_{dp}^2
    .
\end{align*}
Define Lyapunov function $V^t \triangleq \Expect [\byzobj(\bfthe^{t})] + \Expect [\|\nabla \byzobj(\bfthe^t) - \overline{\M}^t\|^2_2] / 16 L + \Expect [\|\nabla \byzobj(\bfthe^t) - \overline{\m}^t\|^2_2] / 16 L$, we have
\begin{small}
\begin{equation*}
\begin{aligned}
&V^{t+1} - V^{t}
=
\expect{\byzobj (\bfthe^{t+1})} - \expect{\byzobj (\bfthe^{t})}
+
\frac{1}{16L} 
\pth{\expect{\norm{\nabla \byzobj(\bfthe^{t+1}) - \overline{\M}^{t+1}}^2}
-
\expect{\norm{\nabla \byzobj(\bfthe^{t}) - \overline{\M}^{t}}^2}
} \\
&~~~+
\frac{1}{16L} 
\pth{\expect{\norm{\nabla \byzobj(\bfthe^{t+1}) - \overline{\m}^{t+1}}^2}
-
\expect{\norm{\nabla \byzobj(\bfthe^{t}) - \overline{\m}^{t}}^2}
} \\
&\le
    - \eta
    \pth{\frac{3}{4} - 2 \eta L}
    \expect{\norm{\nabla \byzobj (\bfthe^{t})}^2}
    +
    \frac{2
    \kappa  
    \eta (1 + \eta L)
    }{\abth{\calG}}
    \sum_{i \in \calG}
    \expect{\norm{
    \M_i^{t}
    -
    \overline{\M}^t
    }^2}
    \\
    &~~~
    +
    2 \eta (1 + \eta L)
    \expect{
    \norm{\nabla \byzobj(\bfthe^t) - \overline{\M}^t}^2}
    +
    \frac{\eta^2 L d}{2}
    \sigma_{dp}^2 
    +
    \frac{\beta^2
    +
    \eta L
    \pth{\beta + (1 - \beta) \alpha}
    \pth{\pth{\beta + (1 - \beta) \alpha}
    \pth{1 +\eta L} + \beta (1 - \beta)}}{16 L}
    d \sigma_{dp}^2
    \\
    &~~~~-
    \frac{1}{16L}
    \pth{
    \expect{\norm{\nabla \byzobj(\bfthe^{t}) - \overline{\M}^{t}}^2} +
    \expect{\norm{\nabla \byzobj(\bfthe^{t}) - \overline{\m}^{t}}^2}
    }
    +
    \frac{(1 - \beta)^2(1 - \alpha)^2}{16 L} \frac{\sigma^2}{\abth{\calG}} 
    \\
        &~~~+
        \frac{\beta}{16L}
        \pth{
        \beta  +
        \frac{(1 - \beta) \alpha }{k_2}
        +
        \eta L
        (\beta + (1 - \beta) \alpha)
        }
        \expect{\norm{\overline{\M}^{t} - \nabla \byzobj(\bfthe^{t})}^2 }\\
        &~~~+
        \frac{3 \eta }{16}
        \pth{\beta + (1 - \beta) \alpha}
        \pth{\pth{\beta + (1 - \beta) \alpha}
        \pth{1 +\eta L} + \beta (1 - \beta)}
        \expect{\norm{\overline{\M}^{t} - \nabla \byzobj(\bfthe^{t})}^2 }\\
        &~~~+
        \frac{(1 - \beta) \alpha}{16 L}
        \qth{\pth{(1 - \beta) \alpha
        +
        2 k_2 \beta}
        \pth{
         1
        +
        \eta L
        } + (1 - \beta) \beta}
       \expect{\norm{
        \overline{\m}^{t}
        -
        \nabla \byzobj (\bfthe^{t})
        }^2 }
        \\
        &~~~+
        \frac{3 \eta }{16}
        \pth{\beta + (1 - \beta) \alpha}
        \pth{\pth{\beta + (1 - \beta) \alpha}
        \pth{1 +\eta L} + \beta (1 - \beta)}
        \expect{\norm{\nabla \byzobj (\bfthe^{t})}^2} \\
        &~~~+
        \frac{(1 - \beta) \beta}{16 L}
        \pth{2 k_2 \beta + (1 - \beta) \alpha + \eta L (\beta + (1-\beta)\alpha)}
        \frac{\kappa}{\abth{\calG}}
        \sum_{i \in \calG}
        \expect{\norm{\m_i^{t} - \overline{\m}^{t}}^2 } \\
        &~~~+
        \frac{3 (1 - \beta)^2
        \eta }{16}
        \pth{\beta + (1 - \beta) \alpha}
        \pth{\pth{\beta + (1 - \beta) \alpha}
        \pth{1 +\eta L} + \beta (1 - \beta)}
        \frac{\kappa}{\abth{\calG}}
        \sum_{i \in \calG}
        \expect{\norm{\m_i^{t} - \overline{\m}^{t}}^2 }\\
    &~~~+
        \frac{\alpha^2 (1  + \eta L) }{16 L}
        \expect{\norm{\overline{\m}^{t} - \nabla \byzobj (\bfthe^{t})}^2 }
        +
        \frac{\alpha^2 
        \eta (1  + \eta L)}{16}
        d \sigma_{dp}^2 
        +
        \frac{(1 - \alpha)^2}{16 L} \frac{\sigma^2}{\abth{\calG}}\\
        &~~~+
        \frac{3 \alpha^2 
        \eta (1  + \eta L)}{16}
        \expect{\norm{\nabla \byzobj (\bfthe^{t})}^2} 
        +
        \frac{3 \alpha^2 \eta (1 + \eta L)}{16}
        \expect{\norm{\overline{\m}^{t} - \nabla \byzobj (\bfthe^{t})}^2 }
        +
        \frac{3 \alpha^2 
        \eta (1  + \eta L)}{16}
        \frac{(1 - \beta)^2 \kappa}{\abth{\calG}}
        \sum_{i \in \calG}
        \expect{\norm{\m_i^{t} - \overline{\m}^{t}}^2} .
\end{aligned}
\end{equation*}
\end{small}
Rearrange the terms, we have
\begin{align*}
    &V^{t+1} - V^{t}\\
    &\le
    - \eta
    \pth{\frac{3}{4} -  \frac{3}{16}\pth{\beta + (1 - \beta) \alpha}
        \pth{\pth{\beta + (1 - \beta) \alpha}
        \pth{1 +\eta L} + \beta (1 - \beta)} - \frac{3 \alpha^2 (1 + \eta L)}{16}- 2 \eta L}
    \expect{\norm{\nabla \byzobj (\bfthe^{t})}^2} \\
    &~~~+
        \frac{(3 \alpha^2 + 32)
        \eta (1  + \eta L)}{16}
        \frac{(1 - \beta)^2 \kappa}{\abth{\calG}}
        \sum_{i \in \calG}
        \expect{\norm{\m_i^{t} - \overline{\m}^{t}}^2 } \\
        &~~~+
        \frac{(1 - \beta) \beta}{16 L}
        \pth{2 k_2 \beta + (1 - \beta) \alpha + \eta L (\beta + (1-\beta)\alpha)}
        \frac{\kappa}{\abth{\calG}}
        \sum_{i \in \calG}
        \expect{\norm{\m_i^{t} - \overline{\m}^{t}}^2 } \\
        &~~~+
        \frac{3 (1 - \beta)^2
        \eta }{16}
        \pth{\beta + (1 - \beta) \alpha}
        \pth{\pth{\beta + (1 - \beta) \alpha}
        \pth{1 +\eta L} + \beta (1 - \beta)}
        \frac{\kappa}{\abth{\calG}}
        \sum_{i \in \calG}
        \expect{\norm{\m_i^{t} - \overline{\m}^{t}}^2 }  \\
        &~~~+
        \pth{2 \eta (1 + \eta L) - \frac{1}{16 L}
        +
        \frac{\beta}{16L}
        \pth{
        \beta  +
        (1 - \beta) \frac{\alpha}{k_2}
        +
        \eta L
        (\beta + (1 - \beta) \alpha)
        } 
        }
        \expect{\norm{\nabla \byzobj(\bfthe^t) - \overline{\M}^t}^2} \\
        &~~~+
        \frac{3 \eta }{16}
        \pth{\pth{\beta + (1 - \beta) \alpha}
        \pth{\pth{\beta + (1 - \beta) \alpha}
        \pth{1 +\eta L} + \beta (1 - \beta)} + \alpha^2  (1 + \eta L)}
        \expect{\norm{\nabla \byzobj(\bfthe^t) - \overline{\M}^t}^2} \\
        &~~~+
        \pth{
        - \frac{1}{16L}
        +
        \frac{\alpha^2 (1  + \eta L)}{16L}
        +
        \frac{(1 - \beta) \alpha}{16 L}
        \qth{\pth{(1 - \beta) \alpha
        +
        2 k_2 \beta}
        \pth{
         1
        +
        \eta L
        } + (1 - \beta) \beta}
        }
        \expect{\norm{\overline{\m}^{t} - \nabla \byzobj (\bfthe^{t})}^2 }
        \\
        &~~~
        +
        \pth{
        \frac{\eta^2 L}{2}
        +
        \frac{\beta^2
        +
        \eta L
        \pth{\beta + (1 - \beta) \alpha}
        \pth{\pth{\beta + (1 - \beta) \alpha}
        \pth{1 +\eta L} + \beta (1 - \beta)}}{16 L}
        +
        \frac{\alpha^2 
        \eta (1  + \eta L)}{16}
        }
        d \sigma_{dp}^2 \\
        &~~~+
        \frac{(1 - \alpha)^2 (1 + (1 - \beta)^2)}{16 L} \frac{\sigma^2}{\abth{\calG}}
        .
\end{align*}
Next, we bound the coefficients of the terms $\Expect [\|\nabla \byzobj (\bfthe^t) - \overline{\M}^t\|^2_2]$ and $\Expect[\|\nabla \byzobj (\bfthe^t) - \overline{\m}^t\|^2_2]$. 
Recall the definition of $\alpha = 1 - k_1 \eta L  = 1 - 5100 \eta L$ and $\beta = 1 - k_2 \eta L  = 1 - 50 \eta L$,
it holds that $\alpha \le \beta$.
\paragraph{Bounding $C_1$ for $\Expect [\|\nabla \byzobj (\bfthe^t) - \overline{\M}^t\|^2_2]$.}
\begin{align*}
    C_1 &\triangleq
    2 \eta (1 + \eta L) - \frac{1}{16 L}
    +
    \frac{\beta }{16L}
    \pth{
    \beta  +
    (1 - \beta) \frac{\alpha}{k_2}
    +
    \eta L
    (\beta + (1 - \beta) \alpha)
    } 
    \\&\qquad \qquad \qquad
    +
    \frac{3 \eta }{16}
    \pth{\pth{\beta + (1 - \beta) \alpha}
    \pth{\pth{\beta + (1 - \beta) \alpha}
    \pth{1 +\eta L} + \beta (1 - \beta)} + \alpha^2  (1 + \eta L)}\\
    &\le
    2 \eta (1 + \eta L) - \frac{k_2 \eta}{16}
    +
    \frac{\eta}{16}
    \pth{1 + (1 + k_2 \eta L) 
    }
    +
     \frac{3 \eta }{16}
    \pth{\pth{1 + k_2 \eta L}
    \pth{\pth{1 + k_2 \eta L}
    \pth{1 +\eta L} + k_2 \eta L} +  (1 + \eta L)}
    \\
    &\le
    -\frac{\eta}{16}
    \pth{
    k_2 - 32 (1 + \eta L)
    -
    \pth{2 + k_2 \eta L}
    -
    3 \pth{1 + k_2 \eta L}
    \pth{\pth{1 + k_2 \eta L}
    \pth{1 +\eta L} + k_2 \eta L +  (1 + \eta L)}
    } \\
    &\le
    -\frac{\eta}{16}
    \pth{
    50 
    - 33
    -
    2.5
    -
    12
    } 
    =
    -\frac{5\eta}{32}< 0
    ,
\end{align*}
where the third to last inequality holds because $\eta \le \frac{1}{132 L}$ and $k_2 = 50$.

\paragraph{Bounding $C_2$ for $\Expect[\|\nabla \byzobj (\bfthe^t) - \overline{\m}^t\|^2_2]$.}
\begin{align*}
    C_2 &\triangleq
    - \frac{1}{16L}
        +
        \frac{\alpha^2 (1  + \eta L)}{16L}
        +
        \frac{(1 - \beta) \alpha}{16 L}
        \qth{\pth{(1 - \beta) \alpha
        +
        2 k_2 \beta}
        \pth{
         1
        +
        \eta L
        } + (1 - \beta) \beta} \\
        &\le
        -
        \frac{1}{16 L}
        \pth{
        1 - \alpha
        - \alpha^2 \eta L
        -
        (1 - \beta) \alpha
        \pth{\pth{(1 - \beta) \alpha
        +
        2 k_2 \beta}
        \pth{
         1
        +
        \eta L
        } + (1 - \beta) \beta}
        } \\
        &\le
        -
        \frac{\eta}{16}
        \pth{
        k_1 - 1
        -
        k_2^2  
        \pth{\pth{\eta L 
        +
        2 }
        \pth{
         1
        +
        \eta L
        } + \eta L }
        } \\
        &\le
        -
        \frac{\eta}{16}
        \pth{
        5100 
        - \frac{3}{132}
        - 4 
        - 33^2 \pth{(\frac{1}{132}+2)\times(\frac{1}{132}+1) + \frac{1}{132}}
        }
        <
        -
        \frac{20\eta}{16}
        < 0,
\end{align*}
where the second to last inequality holds because $\eta \le \frac{1}{132 L}$ and $k_1 = 5100$.
Therefore, we have
\begin{small}
\begin{align*}
    &V^{t+1} - V^{t}\\
    &\le
    - \eta
    \pth{\frac{3}{4} -  \frac{3}{16}\pth{\beta + (1 - \beta) \alpha}
        \pth{\pth{\beta + (1 - \beta) \alpha}
        \pth{1 +\eta L} + \beta (1 - \beta)} - \frac{3 \alpha^2 (1 + \eta L)}{16}- 2 \eta L}
    \expect{\norm{\nabla \byzobj (\bfthe^{t})}^2} \\
    &~~~+
        \frac{(3 \alpha^2 + 32)
        \eta (1  + \eta L)}{16}
        \frac{(1 - \beta)^2 \kappa}{\abth{\calG}}
        \sum_{i \in \calG}
        \expect{\norm{\m_i^{t} - \overline{\m}^{t}}^2 } \\
        &~~~+
        \frac{(1 - \beta) \beta}{16 L}
        \pth{2 k_2 \beta + (1 - \beta) (\beta + \alpha) + \eta L (\beta + (1-\beta)\alpha)}
        \frac{\kappa}{\abth{\calG}}
        \sum_{i \in \calG}
        \expect{\norm{\m_i^{t} - \overline{\m}^{t}}^2 } \\
        &~~~+
        \frac{3 (1 - \beta)^2
        \eta }{16}
        \pth{\beta + (1 - \beta) \alpha}
        \pth{\pth{\beta + (1 - \beta) \alpha}
        \pth{1 +\eta L} + \beta (1 - \beta)}
        \frac{\kappa}{\abth{\calG}}
        \sum_{i \in \calG}
        \expect{\norm{\m_i^{t} - \overline{\m}^{t}}^2 }  \\
        &~~~+C_1
        \expect{\norm{\nabla \byzobj(\bfthe^t) - \overline{\M}^t}^2} 
        +C_2
        \expect{\norm{\overline{\m}^{t} - \nabla \byzobj (\bfthe^{t})}^2 }
        +
        \frac{(1 - \alpha)^2 (1 + (1 - \beta)^2)}{16 L} \frac{\sigma^2}{\abth{\calG}} 
        \\
        &~~~
        +
        \pth{
        \frac{\eta^2 L}{2}
        +
        \frac{\beta^2
        +
        \eta L
        \pth{\beta + (1 - \beta) \alpha}
        \pth{\pth{\beta + (1 - \beta) \alpha}
        \pth{1 +\eta L} + \beta (1 - \beta)}}{16 L}
        +
        \frac{\alpha^2 
        \eta (1  + \eta L)}{16}
        }
        d \sigma_{dp}^2 \\
        &\le
        - \eta
    \pth{\frac{3}{4} -  \frac{3}{16}\pth{\beta + (1 - \beta) \alpha}
        \pth{\pth{\beta + (1 - \beta) \alpha}
        \pth{1 +\eta L} + \beta (1 - \beta)} - \frac{3 \alpha^2 (1 + \eta L)}{16}- 2 \eta L}
    \expect{\norm{\nabla \byzobj (\bfthe^{t})}^2} \\
    &~~~+
        \frac{(3 \alpha^2 + 32)
        \eta (1  + \eta L)}{16}
        \frac{(1 - \beta)^2 \kappa}{\abth{\calG}}
        \sum_{i \in \calG}
        \expect{\norm{\m_i^{t} - \overline{\m}^{t}}^2 }
        +
        \frac{(1 - \alpha)^2 (1 + (1 - \beta)^2)}{16 L} \frac{\sigma^2}{\abth{\calG}} \\
        &~~~+
        \frac{(1 - \beta) \beta}{16 L}
        \pth{2 k_2 \beta + (1 - \beta) \alpha + \eta L (\beta + (1-\beta)\alpha)}
        \frac{\kappa}{\abth{\calG}}
        \sum_{i \in \calG}
        \expect{\norm{\m_i^{t} - \overline{\m}^{t}}^2 } \\
        &~~~+
        \frac{3 (1 - \beta)^2
        \eta }{16}
        \pth{\beta + (1 - \beta) \alpha}
        \pth{\pth{\beta + (1 - \beta) \alpha}
        \pth{1 +\eta L} + \beta (1 - \beta)}
        \frac{\kappa}{\abth{\calG}}
        \sum_{i \in \calG}
        \expect{\norm{\m_i^{t} - \overline{\m}^{t}}^2 }  \\
        &~~~
        +
        \pth{
        \frac{\eta^2 L}{2}
        +
        \frac{\beta^2
        +
        \eta L
        \pth{\beta + (1 - \beta) \alpha}
        \pth{\pth{\beta + (1 - \beta) \alpha}
        \pth{1 +\eta L} + \beta (1 - \beta)}}{16 L}
        +
        \frac{\alpha^2 
        \eta (1  + \eta L)}{16}
        }
        d \sigma_{dp}^2 
        .
\end{align*}
\end{small}
Furthermore, it holds for the coefficients of $\Expect [\|\nabla \byzobj (\bfthe^t) \|_2^2]$ that
\begin{align*}
    &\frac{3}{4} -  \frac{3}{16}\pth{\beta + (1 - \beta) \alpha}
    \pth{\pth{\beta + (1 - \beta) \alpha}
    \pth{1 +\eta L} + \beta (1 - \beta)} - \frac{3 \alpha^2 (1 + \eta L)}{16}- 2 \eta L \\
    &\ge
    \frac{3}{4} -  \frac{3}{16}\pth{1 + k_2 \eta L}
    \pth{\pth{1 + k_2 \eta L}
    \pth{1 +\eta L} + k_2 \eta L} - \frac{3 (1 + \eta L)}{16}- 2 \eta L 
    \ge 
    \frac{2}{25},
\end{align*}
where the last inequality follows from $\eta \le 1/132L$ and $k_2 = 50$.
\paragraph{Bounding $C_3$ for $\Expect\qth{\|\m_i^t - \overline{\m}^t\|_2^2}$.}
\begin{align*}
    &C_3
    \triangleq
    \frac{(3 \alpha^2 + 32) \eta (1  + \eta L) (1 - \beta)^2}{16} 
    +
    \frac{(1 - \beta) \beta}{16 L}
    \pth{2 k_2 \beta + (1 - \beta) (\beta + \alpha) + \eta L (\beta + (1-\beta)\alpha)} \\
    &~~~+
    \frac{3 (1 - \beta)^2
    \eta }{16}
    \pth{\beta + (1 - \beta) \alpha}
    \pth{\pth{\beta + (1 - \beta) \alpha}
    \pth{1 +\eta L} + \beta (1 - \beta)}\\
    &\le
    \frac{35 k_2^2 \eta^3 L^2 (1  + \eta L) }{16} 
    +
    \frac{k_2 \eta}{16}
    \pth{2 k_2 + 2 k_2 \eta L + \eta L (1 + k_2 \eta L)} 
    +
    \frac{3 k_2^2 \eta^3 L^2 }{16}
    \pth{1 + k_2 \eta L}
    \pth{\pth{1 + k_2 \eta L}
    \pth{1 +\eta L} +  k_2 \eta L} \\
    &\le
    \frac{35 \eta \pth{\frac{50}{132}}^2 (1 + \frac{1}{132})}{16} 
    +
    \frac{50 \eta}{16}
    \pth{2 \times 50 + \frac{2 \times 50}{132} + \frac{1}{132} \times (1 + \frac{50}{132})} \\
    &~~~+
    \frac{3 \pth{\frac{50}{132}}^2 \eta }{16}
    (1 + \frac{50}{132}) \times
    ((1 + \frac{50}{132}) \times (1 + \frac{1}{132}) + \frac{50}{132})
    \\
    &\le
    \eta \pth{0.316  + 314 + 0.06558}
    \le
    315 \eta.
\end{align*}
Moreover, we have
\begin{align*}
    &\frac{\eta^2 L}{2}
    +
    \frac{\beta^2
    +
    \eta L
    \pth{\beta + (1 - \beta) \alpha}
    \pth{\pth{\beta + (1 - \beta) \alpha}
    \pth{1 +\eta L} + \beta (1 - \beta)}}{16 L}
    +
    \frac{\alpha^2 
    \eta (1  + \eta L)}{16} \\
    &\le
    \frac{1}{L}
    \pth{
    \frac{\pth{\frac{1}{132}}^2}{2}
    +
    \frac{
    1
    +
    \frac{1}{132}
    \pth{1 + \frac{50}{132}}
    \times
    \pth{\pth{1 + \frac{50}{132}} \times
    \pth{1 +\frac{1}{132}} + \frac{50}{132}}
    }{16}
    +
    \frac{
    \frac{1}{132} \times (1  + \frac{1}{132})}{16} 
    } \\
    &\le
    \frac{3 \times 10^{-5} + 0.065 + 5 \times 10^{-4}}{L}
    \le
    \frac{1}{L}
    .
\end{align*}
Hence, the inequality simplifies to
\begin{align*}
        &V^{t+1} - V^{t}
        \le
        -\frac{2 \eta}{25}
        \expect{\norm{\nabla \byzobj (\bfthe^{t})}^2} 
        +
        \frac{315 \eta \kappa}{\abth{\calG}}
        \sum_{i \in \calG}
        \expect{\norm{\m_i^{t} - \overline{\m}^{t}}^2 } 
        +
        \frac{d \sigma_{dp}^2 }{L}
        +
        \frac{k_1^2 \eta^2 L}{8} \frac{\sigma^2}{\abth{\calG}} \\
        &\le
        -\frac{2 \eta}{25}
        \expect{\norm{\nabla \byzobj (\bfthe^{t})}^2} 
        +
        \frac{d \sigma_{dp}^2 }{L}
        +
        \frac{k_1^2 \eta^2 L}{8} \frac{\sigma^2}{\abth{\calG}}  \\
        &~~~+945 \kappa \eta 
        \pth{1 + \frac{1}{\abth{\calG}}}
        \pth{1 - \alpha^2}
        \sigma^2
        +
        945 \kappa \eta G^2 
        +
        945 \kappa \eta B^2 (1 - \beta)^2
        (1 - \alpha)
        \sum_{r=0}^t
        \alpha^{t - r}
        \expect{\norm{
        \nabla \byzobj (\bfthe^r)
        }^2}
        .
\end{align*}
By telescoping sum, it holds that
\begin{align*}
        &\frac{V^{T} - V^{0}}{T}
        \le
        - \frac{2\eta}{25}
        \frac{1}{T}
        \sum_{t=0}^{T-1}
        \expect{\norm{\nabla \byzobj (\bfthe^{t})}^2} 
        +
        \frac{d \sigma_{dp}^2 }{L}
        +
        \frac{k_1^2 \eta^2 L }{8} \frac{\sigma^2}{\abth{\calG}} \\
        &~~~+945 \kappa \eta
        \pth{1 + \frac{1}{\abth{\calG}}}
        \pth{1 - \alpha^2}
        \sigma^2
        +
        945 \kappa \eta G^2 
        +
        945 \kappa \eta B^2 
        (1 - \alpha)
        \frac{1}{T}
        \sum_{t=0}^{T-1}
        \sum_{r=0}^t
        \alpha^{t - r}
        \expect{\norm{
        \nabla \byzobj (\bfthe^r)
        }^2} \\
        &\le
        - \frac{2\eta}{25}
        \frac{1}{T}
        \sum_{t=0}^{T-1}
        \expect{\norm{\nabla \byzobj (\bfthe^{t})}^2} 
        +
        \frac{d \sigma_{dp}^2 }{L}
        +
        \frac{k_1^2 \eta^2 L }{8} \frac{\sigma^2}{\abth{\calG}} \\
        &~~~+945 \kappa \eta
        \pth{1 + \frac{1}{\abth{\calG}}}
        \pth{1 - \alpha^2}
        \sigma^2
        +
        945 \kappa \eta G^2 
        +
        945 \kappa \eta B^2  
        (1 - \alpha)
        \frac{1}{T}
        \sum_{t=0}^{T-1}
        \expect{\norm{
        \nabla \byzobj (\bfthe^t)
        }^2}
        \pth{\sum_{r=0}^{T-1}
        \alpha^{r}} \\
        &\le
        - \eta
        \pth{\frac{2}{25} - 945 \kappa B^2}
        \frac{1}{T}
        \sum_{t=0}^{T-1}
        \expect{\norm{\nabla \byzobj (\bfthe^{t})}^2} 
        +
        \frac{d \sigma_{dp}^2 }{L}
        +
        \frac{k_1^2 \eta^2 L }{8} \frac{\sigma^2}{\abth{\calG}}
        + 945 \kappa \eta
        \pth{1 + \frac{1}{\abth{\calG}}}
        \pth{1 - \alpha^2}
        \sigma^2
        +
        1638 \kappa \eta G^2 
        .
\end{align*}
For some $B^2 < 1/(23625 \kappa)$, it holds that
\begin{align*}
        \frac{V^{T} - V^{0}}{T}
        &\le
        - 
        \frac{\eta}{25}
        \frac{1}{T}
        \sum_{t=0}^{T-1}
        \expect{\norm{\nabla \byzobj (\bfthe^{t})}^2} 
        +
        \frac{d \sigma_{dp}^2}{L}
        +
        \frac{k_1^2 \eta^2 L }{8} \frac{\sigma^2}{\abth{\calG}} 
        +945 \kappa \eta
        \pth{1 + \frac{1}{\abth{\calG}}}
        \pth{1 - \alpha^2}
        \sigma^2
        +
        945 \kappa \eta G^2 
        .
\end{align*}
Rearrange the terms, we have
\begin{align*}
        \frac{1}{T}
        \sum_{t=0}^{T-1}
        \expect{\norm{\nabla \byzobj (\bfthe^{t})}^2} 
        &\le
        \frac{25 (V^{0} - V^{T})}{\eta T}
        +
        \frac{25 d \sigma_{dp}^2 }{\eta L}
        +
        4 k_1^2 \eta L  \frac{\sigma^2}{\abth{\calG}} 
        +47250 k_1 \eta L
        \kappa 
        \pth{1 + \frac{1}{\abth{\calG}}}
        \sigma^2
        +
        23625 \kappa  G^2 
        ,
\end{align*}
where the inequality holds because
\[
    (1 - \alpha^2) = (1 - \alpha) (1 + \alpha) \le 2 (1 - \alpha).
\]
For the term $-V^{T}$, it holds that
\begin{align*}
    - V^{T}
    &\le
    - \byzobj^{\ast}.
\end{align*}
For the term $V^{0}$, it holds that
\begin{align*}
    V^{0}
    &=
    \Expect [\byzobj(\bfthe^{0})] + \Expect [\|\nabla \byzobj(\bfthe^0) - \overline{\M}^0\|^2_2] / 16 L + \Expect [\|\nabla \byzobj(\bfthe^0) - \overline{\m}^0\|^2_2] / 16 L .
\end{align*}
By definition, we have
\begin{align*}
    \overline{\M}^0 
    &=
    \beta \tilde R^{-1} + (1 - \beta) \overline{\m}^0 
    =
    \beta \tilde R^{-1} + (1 - \beta) \alpha\overline{\m}^{-1} 
    +
    (1 - \beta) (1 - \alpha) \overline{\g}^{0}
    =
    (1 - \beta) (1 - \alpha) \overline{\g}^{0}
    ,
\end{align*}
where the last equality follows from the convention that $\tilde R^{-1} = \overline{\m}^{-1} = \bm{0}$.
It follows that
\begin{align*}
    \norm{\overline{\M}^0 - \nabla \byzobj (\bfthe^{0})}^2
    &=
    \norm{(1 - \beta) (1 - \alpha) \pth{\overline{\g}^{0} - \nabla \byzobj (\bfthe^{0})} - (1 - (1 - \beta) (1 - \alpha))\nabla \byzobj (\bfthe^{0})}^2 \\
    &\le
    (1 - \beta) (1 - \alpha)
    \norm{\overline{\g}^{0} - \nabla \byzobj (\bfthe^{0})}^2 + 
    (1 - (1 - \beta) (1 - \alpha))
    \norm{\nabla \byzobj (\bfthe^{0})}^2,
\end{align*}
where the last inequality holds because of Jensen's inequality.
Take expectation, we have
\begin{align*}
    \expect{\norm{\overline{\M}^0 - \nabla \byzobj (\bfthe^{0})}^2}
    &\le
    (1 - \beta) (1 - \alpha)
    \expect{\norm{\overline{\g}^{0} - \nabla \byzobj (\bfthe^{0})}^2} + 
    (1 - (1 - \beta) (1 - \alpha))
    \expect{\norm{\nabla \byzobj (\bfthe^{0})}^2} \\
    &\le
    k_1 k_2 \eta^2 L^2 \frac{\sigma^{2}}{\abth{\calG}}
    + 
    \expect{\norm{\nabla \byzobj (\bfthe^{0})}^2}
\end{align*}
Similarly, we have
\begin{align*}
    \expect{\norm{\overline{\m}^{0} - \nabla \byzobj (\bfthe^{0})}^2}
    &=
    \expect{\norm{\alpha \overline{\m}^{-1} + (1 - \alpha)\pth{\overline{\g}^0 - \nabla \byzobj (\bfthe^{0})} - \alpha \nabla \byzobj (\bfthe^{0})}^2} \\
    &\overset{(a)}{=}
    \expect{\norm{(1 - \alpha)\pth{\overline{\g}^0 - \nabla \byzobj (\bfthe^{0})} - \alpha \nabla \byzobj (\bfthe^{0})}^2} \\
    &\overset{(b)}{=}
    \expect{\norm{(1 - \alpha)\pth{\overline{\g}^0 - \nabla \byzobj (\bfthe^{0})}}^2} + \expect{\norm{\alpha \nabla \byzobj (\bfthe^{0})}^2} \\
    &\le
    (1 - \alpha)^2 \frac{\sigma^2}{\abth{\calG}}
    +
    \expect{\norm{\nabla \byzobj (\bfthe^{0})}^2}
    ,
\end{align*}
where equality $(a)$ follows from the convention that $\overline{\m}^{-1} = \bm{0}$, and equality $(b)$ follows from the independence in stochastic gradients.
Putting them together, we have
\begin{align*}
    V^{0}
    &\le
    \Expect [\byzobj(\bfthe^{0})] 
    +
    \eta^2 L 
    \frac{k_1 (k_1 + k_2)}{16}
    \frac{\sigma^2}{\abth{\calG}}
    + 
    \frac{1}{8 L}
    \Expect [\|\nabla \byzobj(\bfthe^0)\|^2_2] .
\end{align*}
We know, for a smooth objective, that
\[
\Expect [\|\nabla \byzobj(\bfthe^0)\|^2_2] \le 2 L (\byzobj (\bfthe^0) - \byzobj^{\ast}).
\]
It follows that
\begin{align*}
    V^{0}
    &\le
    \Expect [\byzobj(\bfthe^{0})] 
    +
    \eta^2 L 
    \frac{k_1 (k_1 + k_2)}{16}
    \frac{\sigma^2}{\abth{\calG}}
    + 
    \frac{1}{4}
    \Expect [\byzobj(\bfthe^{0}) - \byzobj^{\ast}] 
    .
\end{align*}
Combining everything together, we have
\begin{align*}
        \frac{1}{T}
        \sum_{t=0}^{T-1}
        \expect{\norm{\nabla \byzobj (\bfthe^{t})}^2} 
        &\le
        \frac{30 ( \expect{\byzobj (\bfthe^0)} - \byzobj^{\ast})}{\eta T}
        +
        \frac{25 d \sigma_{dp}^2 }{\eta L}
        +
        \frac{2 \eta L}{T} 
        k_1 (k_1 + k_2)
        \frac{\sigma^2}{\abth{\calG}}
        +
        4 k_1^2 \eta L  \frac{\sigma^2}{\abth{\calG}}  \\
        &~~~+47250 k_1 \eta L
        \kappa 
        \pth{1 + \frac{1}{\abth{\calG}}}
        \sigma^2
        +
        23625 \kappa  G^2 
        .
\end{align*}
\end{proof}

\subsubsection{Proof of~\prettyref{cor: non-convex convergence v2 main text}}
\label{app: nonconvex convergence learning rate}
\begin{corollary}
    \label{cor: non-convex convergence v2}
    Suppose that Assumptions~\ref{ass: smmothness},~\ref{ass: bounded variance client-wise},~\ref{ass: bounded similarity}, and~\ref{ass: bounded sto gradient cl} hold.
    When $m \ge (2 + \nu) f$, where $\nu > 0$ is an absolute constant,
    we have for~\prettyref{alg: byzantine dp learning base stochastic v2} with coordinate-wise trimmed mean as the robust aggregator that
    \begin{align}
        \notag
        \frac{1}{T}
        \sum_{t=0}^{T-1}
        \expect{\norm{\nabla \byzobj (\bfthe^{t})}^2} 
        &\lesssim
        \frac{\expect{\byzobj (\bfthe^0)} - \byzobj^{\ast}}{\eta T}
        +
        \eta L T 
        \pth{1 + \frac{f}{n}} 
        \frac{\log(1 / \delta) C^2 d }{\epsilon^2 n^2 }
        +
        \frac{\eta L}{T} 
        \frac{\sigma^2}{\abth{\calG}}
        +
        \eta L  \frac{\sigma^2}{\abth{\calG}}  \\
        \label{eq: raw non-convex convergence rate v2}
        &~~~+
        \eta L
        \kappa 
        \pth{1 + \frac{1}{\abth{\calG}}}
        \sigma^2
        +
        \kappa  G^2 
        .
    \end{align}
    When
    \begin{align*}
        \eta &= \min \Bigg\{
        \frac{1}{132 L}, \\
        &\frac{1}{\sqrt{T}}
        \sqrt{\frac{30 ( \expect{\byzobj (\bfthe^0)} - \byzobj^{\ast})}{
        \frac{800 L k_2^2  \log(1 / \delta) C^2 d }{\epsilon^2 n^2}
        \pth{1 + \frac{4 (2 + \nu)^2}{\nu^2} \frac{f}{n}} T + 2 L
        k_1 (k_1 + k_2)
        \frac{\sigma^2}{\abth{\calG}}
        + 4 k_1^2 L  \frac{\sigma^2}{\abth{\calG}}  
        + 94500 k_1 L
        \kappa 
        \sigma^2}}
        \Bigg\},
    \end{align*}
    it holds that
    \begin{small}
    \begin{align*}
        \frac{1}{T}
        \sum_{t=0}^{T-1}
        \expect{\norm{\nabla \byzobj (\bfthe^{t})}^2} 
        &\lesssim
        \frac{\sigma}{\sqrt{T}} 
        \sqrt{ L ( \expect{\byzobj (\bfthe^0)} - \byzobj^{\ast})
        \pth{
        \frac{1}{\abth{\calG}}
        + \kappa
        \pth{1 + 
        \frac{1}{\abth{\calG}}}}}
        + \frac{L ( \expect{\byzobj (\bfthe^0)} - \byzobj^{\ast})}{T} \\
        &~~~ +
        \sqrt{L (\expect{\byzobj (\bfthe^0)} - \byzobj^{\ast})
        \frac{ \log(1 / \delta) C^2 d }{\epsilon^2 n^2}
        \pth{1 + \frac{f}{n}}} +
        \kappa  G^2.
    \end{align*}
    \end{small}
\end{corollary}
    We put $\sigma^2_{dp}$ for coordinate-wise trimmed mean into the formula
    \begin{align*}
        \sigma_{dp}^2 
        &=
        \frac{8 T \log(1 / \delta) \Delta^2_{Byz}}{\epsilon^2}
        =
        \frac{32 (1 - \beta)^2 T \log(1 / \delta) C^2}{\epsilon^2 n^2}
        \pth{1 + \frac{4 (2 + \nu)^2}{\nu^2} \frac{f}{n}} \\
        &\le
        \frac{32 k_2^2 \eta^2 L^2 T \log(1 / \delta) C^2}{\epsilon^2 n^2}
        \pth{1 + \frac{4 (2 + \nu)^2}{\nu^2} \frac{f}{n}} 
        .
    \end{align*}
    Plugging this in, we have
    \begin{align*}
        \frac{1}{T}
        \sum_{t=0}^{T-1}
        \expect{\norm{\nabla \byzobj (\bfthe^{t})}^2} 
        &\le
        \frac{30 ( \expect{\byzobj (\bfthe^0)} - \byzobj^{\ast})}{\eta T}
        +
        \eta L T
        \frac{800 k_2^2  \log(1 / \delta) C^2 d }{\epsilon^2 n^2}
        \pth{1 + \frac{4 (2 + \nu)^2}{\nu^2} \frac{f}{n}} 
        +
        2 \eta L
        k_1 (k_1 + k_2)
        \frac{\sigma^2}{\abth{\calG}}\\
        &~~~+ 4 k_1^2 \eta L  \frac{\sigma^2}{\abth{\calG}}  
        + 94500 k_1 \eta L
        \kappa 
        \sigma^2
        +
        23625 \kappa  G^2 
        .
    \end{align*}
    When we have the optimal learning rate, \ie,
    \begin{align*}
        &\frac{30 ( \expect{\byzobj (\bfthe^0)} - \byzobj^{\ast})}{\eta T} \\
        &=
        \eta L T
        \frac{800 k_2^2  \log(1 / \delta) C^2 d }{\epsilon^2 n^2}
        \pth{1 + \frac{4 (2 + \nu)^2}{\nu^2} \frac{f}{n}} 
        + 2 \eta L
        k_1 (k_1 + k_2)
        \frac{\sigma^2}{\abth{\calG}}
        + 4 k_1^2 \eta L  \frac{\sigma^2}{\abth{\calG}}  
        + 94500 k_1 \eta L
        \kappa 
        \sigma^2,
      \end{align*} 
      we have
        \begin{align*}
        \eta &= \frac{1}{\sqrt{T}}
        \sqrt{\frac{30 ( \expect{\byzobj (\bfthe^0)} - \byzobj^{\ast})}{
        \frac{800 L k_2^2  \log(1 / \delta) C^2 d }{\epsilon^2 n^2}
        \pth{1 + \frac{4 (2 + \nu)^2}{\nu^2} \frac{f}{n}} T + 2 L
        k_1 (k_1 + k_2)
        \frac{\sigma^2}{\abth{\calG}}
        + 4 k_1^2 L  \frac{\sigma^2}{\abth{\calG}}  
        + 94500 k_1 L
        \kappa 
        \sigma^2}}.
        \end{align*}
    Together with the previous condition, we use
    \begin{align*}
        \eta &= \min \sth{
        \frac{1}{132 L}, \frac{1}{\sqrt{T}}
        \sqrt{\frac{30 ( \expect{\byzobj (\bfthe^0)} - \byzobj^{\ast})}{
        \frac{800 L k_2^2  \log(1 / \delta) C^2 d }{\epsilon^2 n^2}
        \pth{1 + \frac{4 (2 + \nu)^2}{\nu^2} \frac{f}{n}} T + 2 L
        k_1 (k_1 + k_2)
        \frac{\sigma^2}{\abth{\calG}}
        + 4 k_1^2 L  \frac{\sigma^2}{\abth{\calG}}  
        + 94500 k_1 L
        \kappa 
        \sigma^2}}
        }.
    \end{align*}
    It follows that
    \begin{align*}
        \frac{1}{\eta}
        &=
        \max \sth{
        132 L, 
        \sqrt{T}
        \sqrt{\frac{
        \frac{800 L k_2^2  \log(1 / \delta) C^2 d}{\epsilon^2 n^2}
        \pth{1 + \frac{4 (2 + \nu)^2}{\nu^2} \frac{f}{n}} T + 2 L
        k_1 (k_1 + k_2)
        \frac{\sigma^2}{\abth{\calG}}
        + 4 k_1^2 L  \frac{\sigma^2}{\abth{\calG}}  
        + 94500 k_1 L
        \kappa 
        \sigma^2}{30 ( \expect{\byzobj (\bfthe^0)} - \byzobj^{\ast})}}
        } \\
        &\le
        132 L + \sqrt{T}
        \sqrt{\frac{
        \frac{800 L k_2^2  \log(1 / \delta) C^2 d }{\epsilon^2 n^2}
        \pth{1 + \frac{4 (2 + \nu)^2}{\nu^2} \frac{f}{n}} T + 2 L
        k_1 (k_1 + k_2)
        \frac{\sigma^2}{\abth{\calG}}
        + 4 k_1^2 L  \frac{\sigma^2}{\abth{\calG}}  
        + 94500 k_1 L
        \kappa 
        \sigma^2}{30 ( \expect{\byzobj (\bfthe^0)} - \byzobj^{\ast})}}.
    \end{align*}
    Therefore, we have
    \begin{footnotesize}
    \begin{align*}
        &\frac{1}{T}
        \sum_{t=0}^{T-1}
        \expect{\norm{\nabla \byzobj (\bfthe^{t})}^2} 
        \le
        \frac{3960 L ( \expect{\byzobj (\bfthe^0)} - \byzobj^{\ast})}{T} \\
        &~~~+
        2 \sqrt{\frac{30 ( \expect{\byzobj (\bfthe^0)} - \byzobj^{\ast})}{T}}
        \sqrt{
        \frac{800 L k_2^2  \log(1 / \delta) C^2 d }{\epsilon^2 n^2}
        \pth{1 + \frac{4 (2 + \nu)^2}{\nu^2} \frac{f}{n}}  T + 2 L
        k_1 (k_1 + k_2)
        \frac{\sigma^2}{\abth{\calG}}
        + 4 k_1^2 L  \frac{\sigma^2}{\abth{\calG}}  
        + 94500 k_1 L
        \kappa 
        \sigma^2} \\
        &~~~ +
        23625 \kappa  G^2.
    \end{align*}
    \end{footnotesize}
    By $\sqrt{x + y} \le \sqrt{x} + \sqrt{y}$, it holds that
    \begin{footnotesize}
    \begin{align*}
        &\frac{1}{T}
        \sum_{t=0}^{T-1}
        \expect{\norm{\nabla \byzobj (\bfthe^{t})}^2} 
        \le
        \frac{3960 L ( \expect{\byzobj (\bfthe^0)} - \byzobj^{\ast})}{T} \\
        &~~~+
        2 \sqrt{\frac{30 ( \expect{\byzobj (\bfthe^0)} - \byzobj^{\ast})}{T}}
        \sqrt{
        \frac{800 L k_2^2  \log(1 / \delta) C^2 d }{\epsilon^2 m^2}
        \pth{1 + \frac{4 (2 + \nu)^2}{\nu^2} \frac{f}{n}}  T + 2 L
        k_1 (k_1 + k_2)
        \frac{\sigma^2}{\abth{\calG}}
        + 4 k_1^2 L  \frac{\sigma^2}{\abth{\calG}}  
        + 94500 k_1 L
        \kappa 
        \sigma^2} \\
        &~~~ +
        23625 \kappa  G^2 \\
        &\le
        \frac{3960 L ( \expect{\byzobj (\bfthe^0)} - \byzobj^{\ast})}{T} \\
        &~~~+
        2 \sqrt{\frac{30 ( \expect{\byzobj (\bfthe^0)} - \byzobj^{\ast})}{T}}
        \sqrt{
        \frac{800 L k_2^2  \log(1 / \delta) C^2 d }{\epsilon^2 n^2}
        \pth{1 + \frac{4 (2 + \nu)^2}{\nu^2} \frac{f}{n}}  T} \\
        &~~~ +
        2 \sqrt{\frac{30 ( \expect{\byzobj (\bfthe^0)} - \byzobj^{\ast})}{T}}
        \sqrt{2 L
        k_1 (k_1 + k_2)
        \frac{\sigma^2}{\abth{\calG}}
        + 4 k_1^2 L  \frac{\sigma^2}{\abth{\calG}}  
        + 94500 k_1 L
        \kappa 
        \sigma^2}
        +
        23625 \kappa  G^2 \\
        &=
        \frac{2 \sigma}{\sqrt{T}} 
        \sqrt{60 ( \expect{\byzobj (\bfthe^0)} - \byzobj^{\ast})
        \pth{L
        k_1 (k_1 + k_2)
        \frac{\sigma^2}{\abth{\calG}}
        + 2 k_1^2 L  \frac{\sigma^2}{\abth{\calG}}  
        + 47250 k_1 L
        \kappa }} 
        + \frac{3960 L ( \expect{\byzobj (\bfthe^0)} - \byzobj^{\ast})}{T} \\
        &~~~ +
        2 \sqrt{24000 L k_2^2 ( \expect{\byzobj (\bfthe^0)} - \byzobj^{\ast})
        \frac{ \log(1 / \delta) C^2 d }{\epsilon^2 n^2}
        \pth{1 + \frac{4 (2 + \nu)^2}{\nu^2} \frac{f}{n}} } +
        23625 \kappa  G^2 .
    \end{align*}
    \end{footnotesize}

\subsection{Supporting Results for the Optimality of {\tt CWTM}}
\label{app: additional convergence}
In this section, we present the remaining supporting results that establish the optimality of our results.

\subsection{Conditions of matching optimality}
\label{app: matching optimality condition}
{\bf\prettyref{lmm: dp lower bound non-adversarial cl}.}
From its proof, we have $$\nabla F_i(\theta)=\mu\theta-\frac{\mu G_0}{L \sqrt{d}}\sum_{j=1}^m\frac{x_i^{j}}{m},$$
where $\sum_{j=1}^m x_i^j/m \in\{\pm 1\}^d$ for $j\in[m]$ and $i\in[n]$. It follows that
$$C=\|\nabla F_i(\mathbf{0})\|_2=\frac{\mu G_0}{L\sqrt{d}}\|\sum_{j=1}^m\frac{x_i^{j}}{m}\|_2=\frac{\mu G_0}{L\sqrt{d}}\sqrt{d}=\frac{\mu G_0}{L}=\Theta(G),$$
where the last equality holds because $\mu=\Theta(L)$ and $G_0=G/2$.

{\bf Lemmas~\ref{lmm: non private lower bound cl} and~\ref{lmm: adversarial dp lower bound cl}.}
Let $\tau = f/(n-f)$ and $\Indc\in\mathbb{R}^d$ be an all-one vector. From the proofs, we have
$$\nabla F_i(\theta)=
\begin{cases}
    \mu(\theta-\frac{G_0 \Indc}{L \sqrt{\tau d}}),~&\text{if}~i\in[1,f];\\
    \mu(\theta+\frac{G_0 \Indc}{L \sqrt{\tau d}}),~&\text{if}~i\in[f+1,2f];\\
    \mu\theta,~&\text{otherwise}.
\end{cases}
$$
It follows that
$$\nabla F_i(\mathbf{0})=
\begin{cases}
    -\frac{\mu G_0 \Indc}{L \sqrt{\tau d}},~&\text{if}~i\in[1,f];\\
    +\frac{\mu G_0 \Indc}{L \sqrt{\tau d}},~&\text{if}~i\in[f+1,2f];\\
    \mathbf{0},~&\text{otherwise}.
\end{cases}
$$
We have 
$$
C = \max_{i\in[n]} \{\|\nabla F_i(\mathbf{0})\|_2\}
= \|\frac{\mu G_0 \Indc}{L \sqrt{\tau d}}\|_2
= \frac{\mu G_0}{L \sqrt{\tau}}
= \Theta(\frac{G}{\sqrt{\tau}}),
$$
where the last equality holds because $\mu=\Theta(L)$ and $G_0=G/\sqrt{2}$. In the regime of $f=\Theta(n)$, it holds that $C=\Theta(G)$. 

\subsection{Optimality of {\tt CWTM}}
\label{app: cwtm optimality}
The proofs and intermediate results below show that coordinate-wise trimmed mean ({\tt CWTM}) is order-optimal in terms of Byzantine-robustness for completeness.
The proofs follow the standard roadmap from~\cite{allouah2023fixing}.

\begin{lemma}
    \label{lmm:cwtm dimension scaling}
    Assume that $R: \reals^{d \times n} \rightarrow \reals^d$ is a coordinate-wise aggregation function, \ie, there exist $d$ real-valued functions $R_1, \ldots, R_{d}$ such that $\forall \x_1, \ldots, \x_{n} \in \reals^d$,
    $[R(\x_1, \ldots, \x_n)]_{r} = R_{r} ([\x_{1}]_{r}, \ldots, [\x_{n}]_{r})$.
    If, for each $r \in [d]$,
    $R_r$ is $(f, \kappa)$-robust then $R$ is $(f,\kappa)$-robust.
\end{lemma}
\begin{proof}[\bf Proof of~\prettyref{lmm:cwtm dimension scaling}]
    Consider $n$ arbitrary vectors in $\reals^{d}$ $\x_1, \ldots, \x_{n}$, and an arbitrary set $S \subseteq [n]$ such that $|S| = n - f$.
    Assume that, for each $r \in [d]$, $R_{r}$ is $(f,\kappa)$-robust.
    Since $R$ is assumed to be a coordinate-wise aggregation function, we have
    \begin{align*}
        \norm{R(\x_1, \ldots, \x_{n}) - \bar{\x}_{S}}^2
        &=
        \sum_{r=1}^{d}
        \abth{
        R_{r} ([\x_{1}]_{r}, \ldots, [\x_{n}]_{r})
        -
        [\bar{\x}_{S}]_{r}
        }^2 \\
        &\overset{(a)}{\le}
        \frac{\kappa}{|S|}
        \sum_{r=1}^{d}
        \sum_{i \in S}
        \abth{
        [\x_{i}]_{r}
        -
        [\bar{\x}_{S}]_{r}
        }^2 
        =
        \frac{\kappa}{|S|}
        \sum_{i \in S}
        \norm{
        \x_{i}
        -
        \bar{\x}_{S}
        }^2 
        ,
    \end{align*}
    where inequality $(a)$ follows from~\prettyref{def: robust agg}.
\end{proof}

\begin{lemma}
    \label{lmm:swapped sets}
    Consider $n$ real values $x_1, \ldots, x_{n}$ such that $x_1 \le \ldots \le x_n$.
    Let $S \subseteq [n]$ of size $|S| = n - f$, and $I \triangleq \{f+1, \ldots, n-f\}$.
    It holds that
    \begin{align*}
        \sum_{i \in I \setminus S}
        \abth{x_{i} - \bar{x}_{S}}^2
        &\le
        \sum_{i \in S \setminus I}
        \abth{x_{i} - \bar{x}_{S}}^2
        .
    \end{align*}
\end{lemma}
\begin{proof}[\bf Proof of~\prettyref{lmm:swapped sets}]
    Observer that $|I| = n - 2f$, $|S| = n -f$, $|I \setminus S| \le f$, and $|S \setminus I| \ge f$.
    Therefore, $|I \setminus S| \le |S \setminus I|$.
    It suffices to show that there exists an injection $\phi: I \setminus S \rightarrow S \setminus I$ such that
    \begin{align*}
        \abth{x_{i} - \bar{x}_{S}} 
        &\le
        \abth{x_{\phi(i)} - \bar{x}_{S}},
        ~~~ \forall i \in I \setminus S.
    \end{align*}
    Since $|I \setminus S| \le |S \setminus I|$, we have $\sum_{i \in I \setminus S} |x_{\phi(i)} - \bar{x}_{S}|^2 \le \sum_{i \in S \setminus I} |x_{\phi(i)} - \bar{x}_{S}|^2$.

    For east of notation, we denote $S^{\complement} \triangleq [n] \setminus S$.
    It follows that $|S^{\complement}| = f$ and $I \setminus S = I \cap S^{\complement}$.
    Let $I^{+} \triangleq \{ n-f+1, \ldots, n\}$ and $I^{-} \triangleq \{1, \ldots, f\}$ define the indices of values that are greater (or equal to) and smaller (or equal to) the values in $I$, respectively.
    Further let $|I \cap S^{\complement}| = q$.
    Since $S^{\complement} = [n] \setminus S$, it holds that
    \begin{align}
        \label{eq:i+s}
        |I^{+} \cap S|
        &=
        f - |I^{+} \cap S^{\complement}|
        \ge
        f - |S^{\complement} \setminus (I \cap S^{\complement})|
        =
        f - (f - q)=q.
    \end{align}
    Similarly, we have 
    \begin{align}
        \label{eq:i-s}
        |I^{-} \cap S| &\ge q.
    \end{align}
    Recall that $q = |I \cap S^{\complement}|$ where $I \setminus S = I \cap S^{\complement}$.
    Therefore, \eqref{eq:i+s} and \eqref{eq:i-s} show that there exists an injection from $I \setminus S$ to $(I^{-} \cap S) \times (I^{+} \cap S)$.
    Let $\psi$ denote this injection. For each $i \in I \setminus S$, $\psi (i)$ is a pair, denoted by $(\psi^{-} (i), \psi^{+} (i))$, in $(I^{-} \cap S) \times (I^{+} \cap S)$.
    Consider an arbitrary $i \in I \setminus S$.
    By the definitions of $I^{-}$ and $I^{+}$, we have
    \begin{align*}
        x_{\psi^{-}(i)}
        &\le
        x_{i} \le x_{\psi^{+}(i)}.
    \end{align*}
    It follows, for any real value $y$, that
    \begin{align*}
        \abth{x_i - y}
        \le
        \max \sth{
        \abth{x_{\psi^{+}(i)}  - y},
        \abth{x_{\psi^{-}(i)}  - y}
        }.
    \end{align*}
    The injection is thus defined as $\phi(i) = \arg\max_{j \in \sth{\psi^{+} (i), \psi^{-} (i)}} \abth{x_{j} - \bar{x}_{S}}$ for all $i \in I \setminus S$.
\end{proof}

\begin{proposition}
    \label{prop:cwtm order optimal}
    Suppose that $n \in \naturals^{+}$ and $f < n / 2$.
    {\tt CWTM} is $(f,\kappa)$-robust with 
    \[
        \kappa = \frac{6 f}{n - 2f} \pth{1 + \frac{f}{n-2f}}.
    \]
\end{proposition}
\begin{proof}[\bf Proof of~\prettyref{prop:cwtm order optimal}]
    Due to~\prettyref{lmm:cwtm dimension scaling}, it suffices to show {\tt CWTM} is $(f,\kappa)$-robust in the scalar domain.

    Without loss of generality, let $x_1, \ldots, x_{n} \in \reals$ such that $x_1 \le \ldots x_{n}$.
    We denote by $I \triangleq \{f+1, \ldots, n-f\}$, and let $S$ be an arbitrary subset of $[n]$ of size $n-f$.
    Recall~\prettyref{def: cwtm}, it holds that ${\tt TM} = \frac{1}{n-2f} \sum_{i \in I} x_i$.
    It follows that
    \begin{align*}
        \abth{{\tt CWTM} (x_1, \ldots, x_n) - \bar{x}_{S}}^2
        &=
        \abth{\frac{1}{n-2f} \sum_{i \in I} x_i - \bar{x}_S}^2
        =
        \abth{\frac{1}{n-2f} \sum_{i \in I} \pth{x_i - \bar{x}_S}}^2 \\
        &\overset{(a)}{=}
        \abth{\frac{1}{n-2f} \sum_{i \in I} \pth{x_i - \bar{x}_S}
        -
        \frac{1}{n-2f} \sum_{i \in S} \pth{x_i - \bar{x}_S}
        }^2 \\
        &=
        \abth{\frac{1}{n-2f} \sum_{i \in I \setminus S} \pth{x_i - \bar{x}_S}
        -
        \frac{1}{n-2f} \sum_{i \in S \setminus I} \pth{x_i - \bar{x}_S}
        }^2
        ,
    \end{align*}
    where equality $(a)$ follows from the fact that $\sum_{i \in S} (x_i - \bar{x}_S) = 0$.
    It follows that
    \begin{align*}
        \abth{{\tt CWTM} (x_1, \ldots, x_n) - \bar{x}_{S}}^2
        &\le
        \frac{|I \setminus S|+|S \setminus I|}{(n-2f)^2}
        \pth{
        \sum_{i \in I \setminus S}
        \abth{x_i - \bar{x}_{S}}^2
        +
        \sum_{i \in S \setminus I}
        \abth{x_i - \bar{x}_{S}}^2
        }.
    \end{align*}
    Note that $|I \setminus S| = |I \cup S| - |S| \le n - (n-f) = f$.
    Similarly, we have $|S \setminus I| = |I \cup S| - |I| \le n - (n - 2f) = 2f$,
    Therefore, it follows that $|I \setminus S| + |S \setminus I| \le 3f$, which yields 
    \begin{align*}
        \abth{{\tt CWTM} (x_1, \ldots, x_n) - \bar{x}_{S}}^2
        &\le
        \frac{3f}{(n-2f)^2}
        \pth{
        \sum_{i \in I \setminus S}
        \abth{x_i - \bar{x}_{S}}^2
        +
        \sum_{i \in S \setminus I}
        \abth{x_i - \bar{x}_{S}}^2
        } \\
        &\overset{(b)}{\le}
        \frac{6f}{(n-2f)^2}
        \sum_{i \in S \setminus I}
        \abth{x_i - \bar{x}_{S}}^2
        \le
        \frac{6f}{(n-2f)^2}
        \sum_{i \in S}
        \abth{x_i - \bar{x}_{S}}^2
        ,
    \end{align*}
    where inequality $(b)$ follows from~\prettyref{lmm:swapped sets}
    Since
    \begin{align*}
        \frac{f}{(n-2f)^2}
        &=
        \frac{f (n - f)}{(n-2f)^2}
        \frac{1}{n - f}
        =
        \pth{
        \frac{f}{n-2f}
        +
        \pth{\frac{f}{n-2f}}^2
        }
        \frac{1}{n - f} \\
        &\le
        \pth{
        \frac{f}{n-2f}
        +
        \pth{\frac{f}{n-2f}}^2
        }
        =
        \frac{f}{n-2f}
        \pth{
        1
        +
        \frac{f}{n-2f}
        }.
    \end{align*}
    
\end{proof}

\begin{claim}
    \label{clm:order optimal}
    When $n \ge (2 + \nu) f$, where $\nu > 0$, it holds that
    \begin{align*}
        \frac{f}{n}
        &\le
        \frac{f}{n-2f}
        \pth{1 + \frac{f}{n - 2f}}
        \le
        \frac{(\nu + 2) (\nu + 1)}{\nu^2}
        \frac{f}{n}.
    \end{align*}
    Equivalently, we have
    \[
        \frac{f}{n-2f}
        \pth{1 + \frac{f}{n - 2f}} = \Theta (\frac{f}{n}).
    \]
\end{claim}
\begin{proof}[\bf Proof of~\prettyref{clm:order optimal}]
    $n \ge (2 + \nu) f$, where $\nu > 0$, implies that $n -2f \ge \nu f$, which leads to
    \[
        \frac{f}{n - 2f} \le \frac{1}{\nu}.
    \]
    It also implies that
    \[
        n - 2f \ge n - \frac{2n}{\nu + 2} = \frac{\nu n}{\nu + 2}.
    \]

    \paragraph{Upper bound.}
    \begin{align*}
        \frac{f}{n - 2f}
        \pth{1 + \frac{f}{n - 2f}}
        &\le
        \frac{f}{n - 2f}
        \frac{\nu + 1}{\nu}
        \le
        \frac{\nu + 2}{\nu}
        \frac{\nu + 1}{\nu}
        \frac{f}{n}
        =
        \frac{(\nu + 2) (\nu + 1)}{\nu^2}
        \cdot
        \frac{f}{n}
    \end{align*}
    
    \paragraph{Lower bound.}
    \begin{align*}
        \frac{f}{n - 2f}
        \pth{1 + \frac{f}{n - 2f}}
        &\ge
        \frac{f}{n - 2f}
        \ge
        \frac{f}{n}
        .
    \end{align*}

    \paragraph{Putting them together.}
    \begin{align*}
        \frac{f}{n}
        &\le
        \frac{f}{n-2f}
        \pth{1 + \frac{f}{n - 2f}}
        \le
        \frac{(\nu + 2) (\nu + 1)}{\nu^2}
        \frac{f}{n}.
    \end{align*}
\end{proof}

\begin{corollary}
    \label{cor:cwtm optimal}
    Suppose that $n \in \naturals^{+}$ and $n \ge (2 + \nu) f$, where $\nu > 0$,
    {\tt CWTM} is $(f,\kappa)$-robust with 
    \[
        \kappa = \Theta \pth{\frac{f}{n}}.
    \]
\end{corollary}
\prettyref{cor:cwtm optimal} can be shown by combining~\prettyref{prop:cwtm order optimal} and~\prettyref{clm:order optimal}.

\newpage
\section{Experiment Details}
\label{app: experiments}

\subsection{Experimental setups}
\label{app: numerical setup}

\noindent{\bf Hardware and Software Setups.}
\begin{itemize}[leftmargin=*]
\item{\bf Hardware.}
The simulations are performed on a private cluster of multiple nodes, 
each node is equipped with 8 CPUs, 64 GB of RAM, and 1 NVIDIA A5000 GPU.

\item {\bf Software.}
We code our experiments based on PyTorch 2.8.0 and Python 3.9.1.
\end{itemize}

\noindent{\bf Neural Network and Hyper-parameter Specifications.}

Table~\ref{tbl: cnn structures} specifies details of the structures of the convolutional neural network and training.
The local learning rate $\eta$ 
is searched based on the best performance after $100$ global rounds, 
over two grids $\{0.75, 0.5, 0.1, 0.03, 0.01, 0.003, 0.001, 0.0003\}$.
The client-side momentum coefficient $\alpha = 0.1$, 
and the server-side momentum coefficient $\beta = 0.85$.
The DP $\delta$ is set to be $10^{-4}$ throughout.

\noindent\textbf{Datasets and Data Heterogeneity.}
The MNIST dataset \cite{lecun2010mnist} contains 10 classes of images. 
In total, there are 50,000 train images and 10,000 test images.
The Fashion-MNIST dataset \cite{xiao2017fashion} contains 10 classes of images. 
In total, there are 60,000 train images and 10,000 test images.
Each client holds two classes of images.

\begin{table}[!t]
\caption{
Neural network architecture, 
loss function, 
training steps,
and batch size specifications
}
\label{tbl: cnn structures}
\centering
\begin{tabular}{cc}
\toprule
{\bf Dataset}& 
{\bf MNIST} 
/ 
{\bf Fashion-MNIST}
\\
\toprule
Neural network &
CNN
\\
Model architecture$^*$ &
\begin{tabular}{p{.6\textwidth}}
\centering
{\bf C}(1,16,8,2,3)
-- {\bf T}
-- {\bf M}(2,1)
-- {\bf C}(16,32,4,2)
-- {\bf T}
-- {\bf M}(2,1)
-- {\bf L}(512,32)
-- {\bf T}
-- {\bf L}(10)
\end{tabular}
\\
\midrule
Loss function &
Cross-entropy loss
 \\
\addlinespace[1ex]
Number of global rounds $T$ &
800
\\
\midrule
Batch size &
20
\\
\bottomrule
\end{tabular}

\begin{tabular}{p{.95\textwidth}}
$^*$
\begin{footnotesize}    
{\bf C}
(\# in-channel, 
\# out-channel, kernel, stride, padding): a 2D convolution layer 
;
{\bf T}: tanh activation function;
{\bf M} 
(kernel, stride): a 2D max-pool layer;
{\bf L} 
(\# outputs) or 
{\bf L}
(\# inputs, 
\# outputs): a fully-connected linear layer;
{\bf D}: a dropout layer (probability 0.2).
\end{footnotesize}
\end{tabular}
\end{table}
\begin{table}[!t]
    \centering
    \caption{
    Test accuracy comparisons of different robust aggregators with and without server-side momentum on the MNIST and Fashion-MNIST datasets.
    Experiments run for a total of $800$ rounds.
    Each client holds $2$ classes of images out of a total of $10$ classes.
    $\beta = 0$ indicates that server-side momentum is turned {\em off}, 
    and $\beta > 0$ indicates that server-side momentum is turned {\em on}.
    All the results are obtained based on an average over the last $50$ rounds across $4$ different random seeds.
    }
    \label{tab:benefits of server momentum}
    \resizebox{\textwidth}{!}{
    \begin{tabular}{c|cccccc}
\toprule
{\bf Datasets} & 
{\bf Attack} & 
{\bf Settings} & 
{\tt CWTM} & 
{\tt CClip} & 
{\tt GM} & 
{\tt CWMED} \\
\midrule
\multirow{8}{*}{
\begin{tabular}{c}
 {\bf MNIST}
\end{tabular}
}
& \multirow{2}{*}{ALIE}
& $\beta = 0$ 
& $80.7 \pm 2.6$\% 
& $89.8 \pm 1.0$\% 
& $67.1 \pm 1.0$\% 
& $38.9 \pm 8.0$\% \\
& 
& $\beta > 0$ 
& $83.3 \pm 5.4$\% $(\ua 2.6\%)$
& $88.4 \pm 3.1$\% $ (\da -1.4\%)$
& $86.0 \pm 4.8$\% $(\ua 18.9\%)$
& $21.6 \pm 9.4$\% $(\da -17.3\%)$\\

\cline{2-7}
& \multirow{2}{*}{BitFlip}
& $\beta = 0$ 
& $67.5 \pm 2.3$\% 
& $80.7 \pm 2.0$\% 
& $77.4 \pm 1.7$\% 
& $68.9 \pm 3.9$\% \\
& 
& $\beta > 0$ 
& $87.1 \pm 3.7$\% $(\ua 19.6\%)$
& $89.4 \pm 4.1$\% $(\ua 8.7\%)$
& $89.0 \pm 1.7$\% $(\ua 11.6\%)$
& $84.8 \pm 1.9$\% $(\ua 15.9\%)$ \\

\cline{2-7}
& \multirow{2}{*}{Mimic}
& $\beta = 0$ 
& $70.1 \pm 3.2$\% 
& $87.5 \pm 1.1$\% 
& $40.5 \pm 6.3$\% 
& $19.3 \pm 1.0$\% \\
& 
& $\beta > 0$ 
& $85.9 \pm 3.6$\% $(\ua 15.8)$ 
& $90.0 \pm 4.0$\% $(\ua 2.5)$ 
& $79.2 \pm 4.2$\% $(\ua 38.7)$
& $59.6 \pm 6.8$\% $(\ua 40.3)$ \\

\cline{2-7}
& \multirow{2}{*}{IPM}
& $\beta = 0$ 
& $73.6 \pm 4.3$\% 
& $88.4 \pm 1.0$\% 
& $86.2 \pm 2.0$\% 
& $68.9 \pm 3.7$\% \\
& 
& $\beta > 0$ 
& $89.2 \pm 4.1$\% $(\ua 15.6\%)$
& $90.2 \pm 3.3$\% $(\ua 1.8\%)$
& $90.0 \pm 1.2$\% $(\ua 3.8\%)$
& $87.1 \pm 1.2$\% $(\ua 18.2\%)$\\
\midrule
\multirow{8}{*}{
\begin{tabular}{c}
 {\bf Fashion} \\
 {\bf MNIST}
\end{tabular}
}
& \multirow{2}{*}{ALIE}
& $\beta = 0$ 
& $68.6 \pm 2.0$\% 
& $71.7 \pm 1.0$\% 
& $62.8 \pm 2.5$\% 
& $48.0 \pm 3.6$\% \\
& 
& $\beta > 0$ 
& $71.1 \pm 3.7$\% $ (\ua 2.5\%)$ 
& $69.6 \pm 2.6$\% $(\da -2.1\%)$
& $64.6 \pm 3.4$\% $(\ua 1.8\%)$
& $44.1 \pm 9.4$\% $(\da -3.9\%)$ \\

\cline{2-7}
& \multirow{2}{*}{BitFlip}
& $\beta = 0$ 
& $63.5 \pm 4.6$\% 
& $65.8 \pm 3.4$\% 
& $64.0 \pm 4.1$\% 
& $59.2 \pm 5.2$\% \\
& 
& $\beta > 0$ 
& $73.9 \pm 4.8$\% $(\ua 10.4\%)$
& $73.5 \pm 3.0$\% $(\ua 7.7\%)$
& $72.7 \pm 4.8$\% $(\ua 8.7\%)$
& $69.4 \pm 4.0$\% $(\ua 10.2\%)$\\

\cline{2-7}
& \multirow{2}{*}{Mimic}
& $\beta = 0$ 
& $60.5 \pm 4.0$\% 
& $70.3 \pm 2.2$\% 
& $37.9 \pm 2.6$\% 
& $37.5 \pm 1.0$\% \\
& 
& $\beta > 0$ 
& $70.8 \pm 6.9$\% $(\ua 10.3\%)$
& $75.4 \pm 2.2$\% $(\ua 5.1\%)$
& $66.3 \pm 4.3$\% $(\ua 28.4\%)$
& $55.3 \pm 4.9$\% $(\ua 17.8\%)$\\

\cline{2-7}
& \multirow{2}{*}{IPM}
& $\beta = 0$ 
& $60.9 \pm 5.0$\% 
& $66.8 \pm 2.0$\% 
& $65.1 \pm 2.0$\% 
& $57.2 \pm 4.6$\% \\
& 
& $\beta > 0$ 
& $74.8 \pm 4.5$\% $(\ua 13.9\%)$
& $75.2 \pm 3.1$\% $(\ua 8.4\%)$
& $68.2 \pm 3.9$\% $(\ua 3.1\%)$
& $72.5 \pm 2.6$\% $(\ua 15.3\%)$ \\

\bottomrule
\end{tabular}
    }
\end{table}

\begin{figure}[!t]
    \begin{subfigure}{\linewidth}
    \centering
    \includegraphics[width=\linewidth]{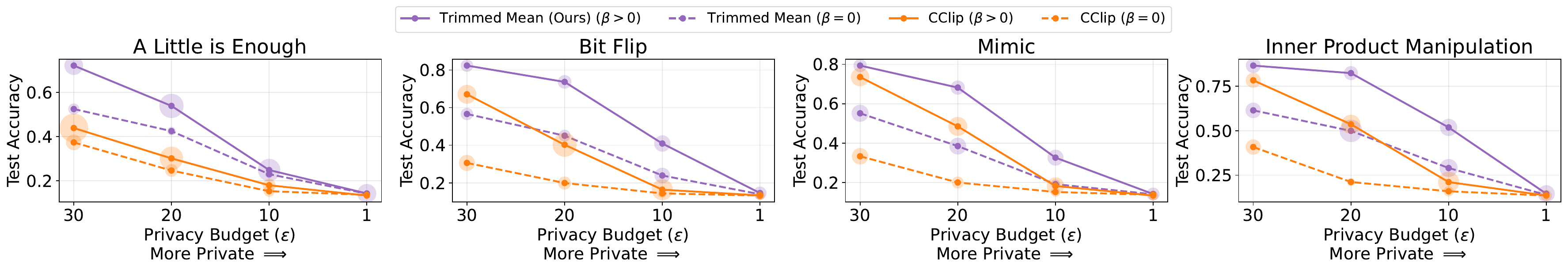}
    \caption{\small
    Test accuracy results on the MNIST dataset.}
    \label{fig:mnist beta}
    \end{subfigure}
    \begin{subfigure}{\linewidth}
     \centering
    \includegraphics[width=\linewidth]{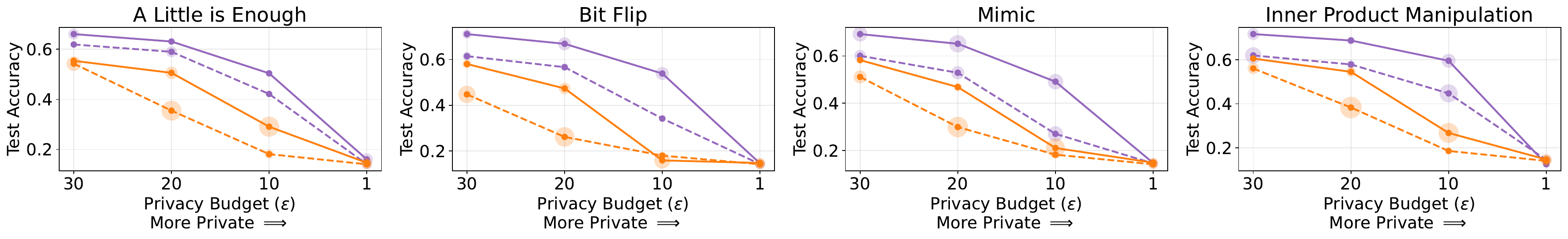}
    \caption{\small
    Test accuracy results on the Fashion-MNIST dataset.}
    \label{fig:fmnist beta}  
    \end{subfigure}
    \caption{\small
    Test accuracy results and comparisons on real-world datasets when the server-side momentum is turned {\em on} $(\beta > 0)$ and when it is turned {\em off} $(\beta = 0)$.
    The results are averaged over the last 50 rounds across $4$ different random seeds.
    All the experiments run for a total of $800$ rounds.
    The radius of the shaded circle denotes the standard deviation at a given privacy budget $\epsilon$.
    }
    \label{fig:beta and utility}
\end{figure}

\subsection{Byzantine Attack Details}
We consider four state-of-the-art gradient attacks, 
including A Little is Enough (ALIE) \cite{baruch2019little},
Bit-flipping (BitFlip) \cite{allenbyzantine},
Inner product manipulation (IPM) \cite{xie2020fall},
and Mimic \cite{karimireddy2021learning}.

Let $\bm{a}_t$ be the attack vector in round $t$ and $\gamma \ge 0$ a fixed constant.
The Byzantine clients send the PS a Byzantine vector $B_t = \bar{\bu}_t + \gamma \bm{a}_t$,
where $\bar{\bu}_t = \frac{1}{\abth{\calG}} \sum_{i \in \calG} \M_i^{t}$. 
We have the following attacks:
\begin{itemize}[leftmargin=*]
    \item {\bf ALIE.}
    $\bm{a}_t = -\bm{\sigma}_t$, where
    $\bm{\sigma}_t \triangleq [\sigma_t^{(1)}, \ldots, \sigma_t^{(d)}]$ and
    $\sigma_t^{(r)}$ is the coordinate-wise standard deviation of the honest
    updates at coordinate $r$. All Byzantine clients send
    $\bm{B}_t = \bar{\bu}_t - \gamma\,\bm{\sigma}_t$, with $\gamma = z_{\max}$
    computed in closed form from $(n, m)$ via the normal CDF~\cite{baruch2019little}.
    \item {\bf BitFlip.}
    $\bm{a}_t = - \bar{\bu}_{t}$ and $\gamma = 2$.
    All Byzantine clients send $\bm{B}_t = - \bar{\bu}_t$ in round $t$.
    \item {\bf Inner Product Manipulation.}
    $\bm{a}_t = - \bar{\bu}_{t}$.
    All Byzantine clients send $\bm{B}_t = (1 - \gamma) \bar{\bu}_t$ in round $t$.
\end{itemize}
The optimal $\gamma$ for \textbf{IPM} is determined by greedily searching over
a defined range of values~\cite{allouah2023fixing}; in each round we choose
$\gamma$ to maximize the $\ell_2$ distance between the robust aggregator's
output and the average of the honest clients' updates.
This strategy ensures the maximum degradation caused by the Byzantine clients in terms of $\ell_2$ distance.
A {\bf mimic} attack is defined as all Byzantine clients mimicking a specific honest client by simply sending their updates to the PS.
We follow the heuristic strategy in \cite{karimireddybyzantine} to determine the set of mocking honest clients.

\subsection{Additional results}
In the absence of user-level DP protections,~\prettyref{tab:benefits of server momentum} shows the test accuracies for different robust aggregators,
including {\tt CWTM}, 
{\tt CClip}, 
{\tt GM},
and {\tt CWMED},
on the MNIST and Fashion-MNIST datasets with server-side momentum enabled and disabled.
Detailed definitions of the robust aggregators can be found in~\prettyref{app: additional sensitivity}.
We can see that the server-side momentum boosts the performance in most cases.
Notably, the improvements are significant.
For example, {\tt GM} improves over $25\%$ under Mimic attack on the Fashion-MNIST dataset.
The improvement stems from the shared anchor $\beta_t \tilde R^{t-1}$, which pulls all clients' updates toward a common direction, thereby strengthening their concentration and improving the robust aggregator's effectiveness against Byzantine clients.
Occasionally, we can see a drop in performance.
We attribute this to the construction of {\tt CWMED}, which yields a suboptimal robust coefficient~\cite{allouah2023fixing}.
{\tt CClip} is also observed to have a mild accuracy drop.
It is interesting to explore how to improve the performance of {\tt CWMED} and {\tt CClip} in a provable way in the future.

\prettyref{fig:beta and utility} shows performance comparisons with server-side momentum enabled and disabled across different privacy budgets for {\tt CWTM} and {\tt CClip}.
Without server-side momentum, both robust aggregators experience a significant performance drop, with extreme cases greater than $30\%$, \eg, for ${\tt CClip}$ under bit-flip and mimic attacks on the MNIST dataset.
Turning off momentum amplifies DP noise and degrades the signal-to-noise ratio, drowning out the effective signals.
Furthermore, our analysis in~\prettyref{sec: convergence results} shows that disabling the server-side momentum will incur an unbounded error, which is proportional to the number of rounds $T$, in the convergence upper bound.
Consequently, the server-side momentum acts as a critical stabilizer, preventing client-level DP noise from causing catastrophic effects over extended training horizons.

\end{document}